\documentclass[sigconf,anonymous]{aamas}
\def\prismPaperMode{preprint} % Choose: anonymous, preprint, or camera-ready.
\def\prismAnonymousMode{anonymous}
\def\prismPreprintMode{preprint}
\def\prismCameraReadyMode{camera-ready}
\newif\ifprismAnonymousModeSelected
\newif\ifprismPreprintModeSelected
\ifx\prismPaperMode\prismAnonymousMode
  \prismAnonymousModeSelectedtrue
\else\ifx\prismPaperMode\prismPreprintMode
  \prismPreprintModeSelectedtrue
  \setkeys{aamas.cls}{anonymous=false,nonacm=true}
\else\ifx\prismPaperMode\prismCameraReadyMode
  \setkeys{aamas.cls}{anonymous=false,nonacm=false}
\else
  \PackageError{prism}{Unknown paper mode `\prismPaperMode'}{Choose anonymous, preprint, or camera-ready.}
\fi\fi\fi

\usepackage{xcolor}   % booktabs, amssymb, amsmath, pifont, graphicx: from the class

\ifprismPreprintModeSelected
  \setcopyright{none}
  \AtBeginDocument{%
    \fancypagestyle{standardpagestyle}{%
      \fancyhf{}%
      \fancyfoot[C]{\thepage}%
    }%
    \fancypagestyle{firstpagestyle}{%
      \fancyhf{}%
      \fancyfoot[C]{\thepage}%
    }%
    \pagestyle{standardpagestyle}%
  }
\else
  \makeatletter
  \gdef\@copyrightpermission{
    \begin{minipage}{0.2\columnwidth}
     \href{https://creativecommons.org/licenses/by/4.0/}{\includegraphics[width=0.90\textwidth]{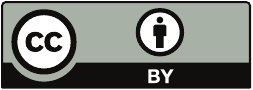}}
    \end{minipage}\hfill
    \begin{minipage}{0.8\columnwidth}
     \href{https://creativecommons.org/licenses/by/4.0/}{This work is licensed under a Creative Commons Attribution International 4.0 License.}
    \end{minipage}
    \vspace{5pt}
  }
  \makeatother

  \setcopyright{ifaamas}
  \acmConference[AAMAS 2027]{Proc.\@ of the 26th International Conference
  on Autonomous Agents and Multiagent Systems (AAMAS 2027)}{3--7 May 2027}{Hanoi, Vietnam}{M.~Baldoni, F.~Fang, W.~Yeoh, N.~Yorke-Smith (eds.)}
  \copyrightyear{2027}
  \acmYear{2027}
  \acmDOI{}
  \acmISBN{}

  \ifprismAnonymousModeSelected
    \acmSubmissionID{1800}   % replace with the OpenReview submission id
    \submissionType{Research Paper Track}
  \fi
\fi

\newcommand{\AuditRegimeStated}{3}

\newcommand{\nmark}{\textcolor{black!30}{--}}
\newcommand{\NC}{\mbox{N-C}}

\newcommand{\slideon}{\textsc{slide}}
\newcommand{\slideoff}{\textsc{block}}
\definecolor{better}{HTML}{2E7D32}
\definecolor{worse}{HTML}{C62828}
\newcommand{\goodchange}[1]{\,{\scriptsize\textcolor{better}{#1}}}
\newcommand{\badchange}[1]{\,{\scriptsize\textcolor{worse}{#1}}}

\newcommand{\benchname}{\textsc{VLN-Safe}}

\newcommand{\Tone}{\textsc{t1}}
\newcommand{\Ttwo}{\textsc{t2}}

\expandafter\def\csname pn@cma.mem.M1.false\endcsname{0.38}
\expandafter\def\csname pn@cma.mem.M1.prec\endcsname{87.7}
\expandafter\def\csname pn@cma.mem.M1.recall\endcsname{43.8}
\expandafter\def\csname pn@cma.mem.M8.false\endcsname{6.19}
\expandafter\def\csname pn@cma.mem.M8.prec\endcsname{35.1}
\expandafter\def\csname pn@cma.mem.M8.recall\endcsname{54.4}
\expandafter\def\csname pn@cma.mem.episodes\endcsname{154}
\expandafter\def\csname pn@cma.r2r.auc.ent\endcsname{.523}
\expandafter\def\csname pn@cma.r2r.auc.geom\endcsname{.826}
\expandafter\def\csname pn@cma.r2r.auc.hist\endcsname{.974}
\expandafter\def\csname pn@cma.r2r.auc.marg\endcsname{.519}
\expandafter\def\csname pn@cma.r2r.auc.stop\endcsname{.301}
\expandafter\def\csname pn@cma.r2r.avail10\endcsname{2.5}
\expandafter\def\csname pn@cma.r2r.avail50\endcsname{23.8}
\expandafter\def\csname pn@cma.r2r.block.cleansuccess\endcsname{9.3}
\expandafter\def\csname pn@cma.r2r.block.cpds\endcsname{34.9}
\expandafter\def\csname pn@cma.r2r.block.cpdsmed\endcsname{5}
\expandafter\def\csname pn@cma.r2r.block.dirtysuccess\endcsname{2.8}
\expandafter\def\csname pn@cma.r2r.block.dsr\endcsname{23.3}
\expandafter\def\csname pn@cma.r2r.block.dsrhi\endcsname{29.3}
\expandafter\def\csname pn@cma.r2r.block.dsrlo\endcsname{18.2}
\expandafter\def\csname pn@cma.r2r.block.nc\endcsname{62.4}
\expandafter\def\csname pn@cma.r2r.block.ncfail\endcsname{70.3}
\expandafter\def\csname pn@cma.r2r.block.ncsucc\endcsname{5.1}
\expandafter\def\csname pn@cma.r2r.block.proxydsr\endcsname{20.2}
\expandafter\def\csname pn@cma.r2r.block.shareinsucc\endcsname{0.3}
\expandafter\def\csname pn@cma.r2r.block.sr\endcsname{12.1}
\expandafter\def\csname pn@cma.r2r.block.stall-mem1.abortep\endcsname{54}
\expandafter\def\csname pn@cma.r2r.block.stall-mem1.aborts\endcsname{1.6}
\expandafter\def\csname pn@cma.r2r.block.stall-mem1.clean_success.d\endcsname{+3.4}
\expandafter\def\csname pn@cma.r2r.block.stall-mem1.clean_success.hi\endcsname{+4.3}
\expandafter\def\csname pn@cma.r2r.block.stall-mem1.clean_success.lo\endcsname{+2.4}
\expandafter\def\csname pn@cma.r2r.block.stall-mem1.clean_success.off\endcsname{9.3}
\expandafter\def\csname pn@cma.r2r.block.stall-mem1.clean_success.on\endcsname{12.7}
\expandafter\def\csname pn@cma.r2r.block.stall-mem1.clean_success.rel\endcsname{+36}
\expandafter\def\csname pn@cma.r2r.block.stall-mem1.collision_count.d\endcsname{$-$293.5}
\expandafter\def\csname pn@cma.r2r.block.stall-mem1.collision_count.hi\endcsname{$-$283.7}
\expandafter\def\csname pn@cma.r2r.block.stall-mem1.collision_count.lo\endcsname{$-$303.3}
\expandafter\def\csname pn@cma.r2r.block.stall-mem1.collision_count.off\endcsname{299.8}
\expandafter\def\csname pn@cma.r2r.block.stall-mem1.collision_count.on\endcsname{6.3}
\expandafter\def\csname pn@cma.r2r.block.stall-mem1.collision_count.rel\endcsname{$-$98}
\expandafter\def\csname pn@cma.r2r.block.stall-mem1.collision_event_count.d\endcsname{+2.14}
\expandafter\def\csname pn@cma.r2r.block.stall-mem1.collision_event_count.hi\endcsname{+2.92}
\expandafter\def\csname pn@cma.r2r.block.stall-mem1.collision_event_count.lo\endcsname{+1.32}
\expandafter\def\csname pn@cma.r2r.block.stall-mem1.collision_event_count.off\endcsname{3.63}
\expandafter\def\csname pn@cma.r2r.block.stall-mem1.collision_event_count.on\endcsname{5.78}
\expandafter\def\csname pn@cma.r2r.block.stall-mem1.collision_event_count.rel\endcsname{+59}
\expandafter\def\csname pn@cma.r2r.block.stall-mem1.dirty_success_flag.d\endcsname{+9.2}
\expandafter\def\csname pn@cma.r2r.block.stall-mem1.dirty_success_flag.hi\endcsname{+10.7}
\expandafter\def\csname pn@cma.r2r.block.stall-mem1.dirty_success_flag.lo\endcsname{+7.7}
\expandafter\def\csname pn@cma.r2r.block.stall-mem1.dirty_success_flag.off\endcsname{2.8}
\expandafter\def\csname pn@cma.r2r.block.stall-mem1.dirty_success_flag.on\endcsname{12.0}
\expandafter\def\csname pn@cma.r2r.block.stall-mem1.dirty_success_flag.rel\endcsname{+325}
\expandafter\def\csname pn@cma.r2r.block.stall-mem1.gained\endcsname{251}
\expandafter\def\csname pn@cma.r2r.block.stall-mem1.lost\endcsname{20}
\expandafter\def\csname pn@cma.r2r.block.stall-mem1.navigation_error_m.d\endcsname{$-$0.3}
\expandafter\def\csname pn@cma.r2r.block.stall-mem1.navigation_error_m.hi\endcsname{$-$0.2}
\expandafter\def\csname pn@cma.r2r.block.stall-mem1.navigation_error_m.lo\endcsname{$-$0.5}
\expandafter\def\csname pn@cma.r2r.block.stall-mem1.navigation_error_m.off\endcsname{7.8}
\expandafter\def\csname pn@cma.r2r.block.stall-mem1.navigation_error_m.on\endcsname{7.4}
\expandafter\def\csname pn@cma.r2r.block.stall-mem1.navigation_error_m.rel\endcsname{$-$4}
\expandafter\def\csname pn@cma.r2r.block.stall-mem1.nc_rate.d\endcsname{$-$58.1}
\expandafter\def\csname pn@cma.r2r.block.stall-mem1.nc_rate.hi\endcsname{$-$56.2}
\expandafter\def\csname pn@cma.r2r.block.stall-mem1.nc_rate.lo\endcsname{$-$60.1}
\expandafter\def\csname pn@cma.r2r.block.stall-mem1.nc_rate.off\endcsname{62.4}
\expandafter\def\csname pn@cma.r2r.block.stall-mem1.nc_rate.on\endcsname{4.3}
\expandafter\def\csname pn@cma.r2r.block.stall-mem1.nc_rate.rel\endcsname{$-$93}
\expandafter\def\csname pn@cma.r2r.block.stall-mem1.ndtw.d\endcsname{+3.0}
\expandafter\def\csname pn@cma.r2r.block.stall-mem1.ndtw.hi\endcsname{+3.9}
\expandafter\def\csname pn@cma.r2r.block.stall-mem1.ndtw.lo\endcsname{+2.1}
\expandafter\def\csname pn@cma.r2r.block.stall-mem1.ndtw.off\endcsname{46.1}
\expandafter\def\csname pn@cma.r2r.block.stall-mem1.ndtw.on\endcsname{49.1}
\expandafter\def\csname pn@cma.r2r.block.stall-mem1.ndtw.rel\endcsname{+7}
\expandafter\def\csname pn@cma.r2r.block.stall-mem1.owned\endcsname{11.7}
\expandafter\def\csname pn@cma.r2r.block.stall-mem1.resid.maneuver\endcsname{51}
\expandafter\def\csname pn@cma.r2r.block.stall-mem1.resid.pass\endcsname{46}
\expandafter\def\csname pn@cma.r2r.block.stall-mem1.spl.d\endcsname{+10.9}
\expandafter\def\csname pn@cma.r2r.block.stall-mem1.spl.hi\endcsname{+12.4}
\expandafter\def\csname pn@cma.r2r.block.stall-mem1.spl.lo\endcsname{+9.5}
\expandafter\def\csname pn@cma.r2r.block.stall-mem1.spl.off\endcsname{11.5}
\expandafter\def\csname pn@cma.r2r.block.stall-mem1.spl.on\endcsname{22.4}
\expandafter\def\csname pn@cma.r2r.block.stall-mem1.spl.rel\endcsname{+94}
\expandafter\def\csname pn@cma.r2r.block.stall-mem1.stalls\endcsname{4.7}
\expandafter\def\csname pn@cma.r2r.block.stall-mem1.step_count.d\endcsname{$-$209.6}
\expandafter\def\csname pn@cma.r2r.block.stall-mem1.step_count.hi\endcsname{$-$199.8}
\expandafter\def\csname pn@cma.r2r.block.stall-mem1.step_count.lo\endcsname{$-$218.9}
\expandafter\def\csname pn@cma.r2r.block.stall-mem1.step_count.off\endcsname{341.2}
\expandafter\def\csname pn@cma.r2r.block.stall-mem1.step_count.on\endcsname{131.7}
\expandafter\def\csname pn@cma.r2r.block.stall-mem1.step_count.rel\endcsname{$-$61}
\expandafter\def\csname pn@cma.r2r.block.stall-mem1.stuck_event_count.d\endcsname{$-$0.9}
\expandafter\def\csname pn@cma.r2r.block.stall-mem1.stuck_event_count.hi\endcsname{$-$0.8}
\expandafter\def\csname pn@cma.r2r.block.stall-mem1.stuck_event_count.lo\endcsname{$-$0.9}
\expandafter\def\csname pn@cma.r2r.block.stall-mem1.stuck_event_count.off\endcsname{0.9}
\expandafter\def\csname pn@cma.r2r.block.stall-mem1.stuck_event_count.on\endcsname{0.0}
\expandafter\def\csname pn@cma.r2r.block.stall-mem1.stuck_event_count.rel\endcsname{$-$100}
\expandafter\def\csname pn@cma.r2r.block.stall-mem1.stuck_flag.d\endcsname{$-$68.1}
\expandafter\def\csname pn@cma.r2r.block.stall-mem1.stuck_flag.hi\endcsname{$-$65.9}
\expandafter\def\csname pn@cma.r2r.block.stall-mem1.stuck_flag.lo\endcsname{$-$70.2}
\expandafter\def\csname pn@cma.r2r.block.stall-mem1.stuck_flag.off\endcsname{68.1}
\expandafter\def\csname pn@cma.r2r.block.stall-mem1.stuck_flag.on\endcsname{0.0}
\expandafter\def\csname pn@cma.r2r.block.stall-mem1.stuck_flag.rel\endcsname{$-$100}
\expandafter\def\csname pn@cma.r2r.block.stall-mem1.success.d\endcsname{+12.6}
\expandafter\def\csname pn@cma.r2r.block.stall-mem1.success.hi\endcsname{+14.3}
\expandafter\def\csname pn@cma.r2r.block.stall-mem1.success.lo\endcsname{+10.9}
\expandafter\def\csname pn@cma.r2r.block.stall-mem1.success.off\endcsname{12.1}
\expandafter\def\csname pn@cma.r2r.block.stall-mem1.success.on\endcsname{24.7}
\expandafter\def\csname pn@cma.r2r.block.stall-mem1.success.rel\endcsname{+104}
\expandafter\def\csname pn@cma.r2r.block.stall-mem1.yields\endcsname{0.16}
\expandafter\def\csname pn@cma.r2r.block.stuck\endcsname{68.1}
\expandafter\def\csname pn@cma.r2r.block.tir\endcsname{35.9}
\expandafter\def\csname pn@cma.r2r.ddsr_pp\endcsname{+17.6}
\expandafter\def\csname pn@cma.r2r.dsr_pp\endcsname{+18.1}
\expandafter\def\csname pn@cma.r2r.hist.flagged\endcsname{547{,}444}
\expandafter\def\csname pn@cma.r2r.hist.recall\endcsname{99.3}
\expandafter\def\csname pn@cma.r2r.hist.tp\endcsname{547{,}444}
\expandafter\def\csname pn@cma.r2r.kmiss\endcsname{83.9}
\expandafter\def\csname pn@cma.r2r.slide.cleansuccess\endcsname{17.8}
\expandafter\def\csname pn@cma.r2r.slide.cpds\endcsname{4.0}
\expandafter\def\csname pn@cma.r2r.slide.cpdsmed\endcsname{2}
\expandafter\def\csname pn@cma.r2r.slide.dirtysuccess\endcsname{12.3}
\expandafter\def\csname pn@cma.r2r.slide.dsr\endcsname{40.9}
\expandafter\def\csname pn@cma.r2r.slide.dsrhi\endcsname{45.0}
\expandafter\def\csname pn@cma.r2r.slide.dsrlo\endcsname{36.9}
\expandafter\def\csname pn@cma.r2r.slide.nc\endcsname{8.0}
\expandafter\def\csname pn@cma.r2r.slide.ncfail\endcsname{10.4}
\expandafter\def\csname pn@cma.r2r.slide.ncsucc\endcsname{2.4}
\expandafter\def\csname pn@cma.r2r.slide.proxydsr\endcsname{1.4}
\expandafter\def\csname pn@cma.r2r.slide.shareinsucc\endcsname{1.8}
\expandafter\def\csname pn@cma.r2r.slide.sr\endcsname{30.2}
\expandafter\def\csname pn@cma.r2r.slide.stuck\endcsname{6.3}
\expandafter\def\csname pn@cma.r2r.slide.tir\endcsname{12.1}
\expandafter\def\csname pn@cma.r2r.swap10\endcsname{2.5}
\expandafter\def\csname pn@cma.r2r.swap50\endcsname{23.8}
\expandafter\def\csname pn@etpnav.r2r.auc.ent\endcsname{.537}
\expandafter\def\csname pn@etpnav.r2r.auc.entnorm\endcsname{.564}
\expandafter\def\csname pn@etpnav.r2r.auc.geom\endcsname{.759}
\expandafter\def\csname pn@etpnav.r2r.auc.hist\endcsname{.916}
\expandafter\def\csname pn@etpnav.r2r.auc.marg\endcsname{.492}
\expandafter\def\csname pn@etpnav.r2r.auc.stop\endcsname{.524}
\expandafter\def\csname pn@etpnav.r2r.avail10\endcsname{3.9}
\expandafter\def\csname pn@etpnav.r2r.avail50\endcsname{31.2}
\expandafter\def\csname pn@etpnav.r2r.block.cleansuccess\endcsname{28.3}
\expandafter\def\csname pn@etpnav.r2r.block.cpds\endcsname{17.2}
\expandafter\def\csname pn@etpnav.r2r.block.cpdsmed\endcsname{9}
\expandafter\def\csname pn@etpnav.r2r.block.dirtysuccess\endcsname{13.2}
\expandafter\def\csname pn@etpnav.r2r.block.dsr\endcsname{31.8}
\expandafter\def\csname pn@etpnav.r2r.block.dsrhi\endcsname{35.2}
\expandafter\def\csname pn@etpnav.r2r.block.dsrlo\endcsname{28.6}
\expandafter\def\csname pn@etpnav.r2r.block.nc\endcsname{21.0}
\expandafter\def\csname pn@etpnav.r2r.block.ncfail\endcsname{32.0}
\expandafter\def\csname pn@etpnav.r2r.block.ncsucc\endcsname{5.5}
\expandafter\def\csname pn@etpnav.r2r.block.proxydsr\endcsname{28.3}
\expandafter\def\csname pn@etpnav.r2r.block.shareinsucc\endcsname{10.8}
\expandafter\def\csname pn@etpnav.r2r.block.sr\endcsname{41.5}
\expandafter\def\csname pn@etpnav.r2r.block.stall-mem1.abortep\endcsname{37}
\expandafter\def\csname pn@etpnav.r2r.block.stall-mem1.aborts\endcsname{0.7}
\expandafter\def\csname pn@etpnav.r2r.block.stall-mem1.caphit.off\endcsname{41.2}
\expandafter\def\csname pn@etpnav.r2r.block.stall-mem1.caphit.on\endcsname{26.8}
\expandafter\def\csname pn@etpnav.r2r.block.stall-mem1.clean_success.d\endcsname{+9.6}
\expandafter\def\csname pn@etpnav.r2r.block.stall-mem1.clean_success.hi\endcsname{+11.0}
\expandafter\def\csname pn@etpnav.r2r.block.stall-mem1.clean_success.lo\endcsname{+8.2}
\expandafter\def\csname pn@etpnav.r2r.block.stall-mem1.clean_success.off\endcsname{28.3}
\expandafter\def\csname pn@etpnav.r2r.block.stall-mem1.clean_success.on\endcsname{37.9}
\expandafter\def\csname pn@etpnav.r2r.block.stall-mem1.clean_success.rel\endcsname{+34}
\expandafter\def\csname pn@etpnav.r2r.block.stall-mem1.collision_count.d\endcsname{$-$18.1}
\expandafter\def\csname pn@etpnav.r2r.block.stall-mem1.collision_count.hi\endcsname{$-$17.0}
\expandafter\def\csname pn@etpnav.r2r.block.stall-mem1.collision_count.lo\endcsname{$-$19.3}
\expandafter\def\csname pn@etpnav.r2r.block.stall-mem1.collision_count.off\endcsname{21.1}
\expandafter\def\csname pn@etpnav.r2r.block.stall-mem1.collision_count.on\endcsname{3.0}
\expandafter\def\csname pn@etpnav.r2r.block.stall-mem1.collision_count.rel\endcsname{$-$86}
\expandafter\def\csname pn@etpnav.r2r.block.stall-mem1.collision_event_count.d\endcsname{+0.36}
\expandafter\def\csname pn@etpnav.r2r.block.stall-mem1.collision_event_count.hi\endcsname{+0.54}
\expandafter\def\csname pn@etpnav.r2r.block.stall-mem1.collision_event_count.lo\endcsname{+0.18}
\expandafter\def\csname pn@etpnav.r2r.block.stall-mem1.collision_event_count.off\endcsname{2.15}
\expandafter\def\csname pn@etpnav.r2r.block.stall-mem1.collision_event_count.on\endcsname{2.51}
\expandafter\def\csname pn@etpnav.r2r.block.stall-mem1.collision_event_count.rel\endcsname{+17}
\expandafter\def\csname pn@etpnav.r2r.block.stall-mem1.dirty_success_flag.d\endcsname{+1.1}
\expandafter\def\csname pn@etpnav.r2r.block.stall-mem1.dirty_success_flag.hi\endcsname{+3.0}
\expandafter\def\csname pn@etpnav.r2r.block.stall-mem1.dirty_success_flag.lo\endcsname{$-$0.8}
\expandafter\def\csname pn@etpnav.r2r.block.stall-mem1.dirty_success_flag.off\endcsname{13.2}
\expandafter\def\csname pn@etpnav.r2r.block.stall-mem1.dirty_success_flag.on\endcsname{14.3}
\expandafter\def\csname pn@etpnav.r2r.block.stall-mem1.dirty_success_flag.rel\endcsname{+8}
\expandafter\def\csname pn@etpnav.r2r.block.stall-mem1.gained\endcsname{233}
\expandafter\def\csname pn@etpnav.r2r.block.stall-mem1.lost\endcsname{37}
\expandafter\def\csname pn@etpnav.r2r.block.stall-mem1.navigation_error_m.d\endcsname{$-$0.6}
\expandafter\def\csname pn@etpnav.r2r.block.stall-mem1.navigation_error_m.hi\endcsname{$-$0.5}
\expandafter\def\csname pn@etpnav.r2r.block.stall-mem1.navigation_error_m.lo\endcsname{$-$0.7}
\expandafter\def\csname pn@etpnav.r2r.block.stall-mem1.navigation_error_m.off\endcsname{5.6}
\expandafter\def\csname pn@etpnav.r2r.block.stall-mem1.navigation_error_m.on\endcsname{5.0}
\expandafter\def\csname pn@etpnav.r2r.block.stall-mem1.navigation_error_m.rel\endcsname{$-$10}
\expandafter\def\csname pn@etpnav.r2r.block.stall-mem1.nc_rate.d\endcsname{$-$18.3}
\expandafter\def\csname pn@etpnav.r2r.block.stall-mem1.nc_rate.hi\endcsname{$-$17.1}
\expandafter\def\csname pn@etpnav.r2r.block.stall-mem1.nc_rate.lo\endcsname{$-$19.6}
\expandafter\def\csname pn@etpnav.r2r.block.stall-mem1.nc_rate.off\endcsname{21.0}
\expandafter\def\csname pn@etpnav.r2r.block.stall-mem1.nc_rate.on\endcsname{2.6}
\expandafter\def\csname pn@etpnav.r2r.block.stall-mem1.nc_rate.rel\endcsname{$-$87}
\expandafter\def\csname pn@etpnav.r2r.block.stall-mem1.ndtw.d\endcsname{+4.1}
\expandafter\def\csname pn@etpnav.r2r.block.stall-mem1.ndtw.hi\endcsname{+4.8}
\expandafter\def\csname pn@etpnav.r2r.block.stall-mem1.ndtw.lo\endcsname{+3.3}
\expandafter\def\csname pn@etpnav.r2r.block.stall-mem1.ndtw.off\endcsname{57.9}
\expandafter\def\csname pn@etpnav.r2r.block.stall-mem1.ndtw.on\endcsname{62.0}
\expandafter\def\csname pn@etpnav.r2r.block.stall-mem1.ndtw.rel\endcsname{+7}
\expandafter\def\csname pn@etpnav.r2r.block.stall-mem1.overcap\endcsname{0}
\expandafter\def\csname pn@etpnav.r2r.block.stall-mem1.owned\endcsname{9.8}
\expandafter\def\csname pn@etpnav.r2r.block.stall-mem1.resid.maneuver\endcsname{55}
\expandafter\def\csname pn@etpnav.r2r.block.stall-mem1.resid.pass\endcsname{43}
\expandafter\def\csname pn@etpnav.r2r.block.stall-mem1.spl.d\endcsname{+9.3}
\expandafter\def\csname pn@etpnav.r2r.block.stall-mem1.spl.hi\endcsname{+10.7}
\expandafter\def\csname pn@etpnav.r2r.block.stall-mem1.spl.lo\endcsname{+7.8}
\expandafter\def\csname pn@etpnav.r2r.block.stall-mem1.spl.off\endcsname{36.4}
\expandafter\def\csname pn@etpnav.r2r.block.stall-mem1.spl.on\endcsname{45.7}
\expandafter\def\csname pn@etpnav.r2r.block.stall-mem1.spl.rel\endcsname{+26}
\expandafter\def\csname pn@etpnav.r2r.block.stall-mem1.stalls\endcsname{2.0}
\expandafter\def\csname pn@etpnav.r2r.block.stall-mem1.step_count.d\endcsname{$-$0.7}
\expandafter\def\csname pn@etpnav.r2r.block.stall-mem1.step_count.hi\endcsname{+0.8}
\expandafter\def\csname pn@etpnav.r2r.block.stall-mem1.step_count.lo\endcsname{$-$2.2}
\expandafter\def\csname pn@etpnav.r2r.block.stall-mem1.step_count.off\endcsname{88.6}
\expandafter\def\csname pn@etpnav.r2r.block.stall-mem1.step_count.on\endcsname{88.0}
\expandafter\def\csname pn@etpnav.r2r.block.stall-mem1.step_count.rel\endcsname{$-$1}
\expandafter\def\csname pn@etpnav.r2r.block.stall-mem1.stuck_event_count.d\endcsname{$-$1.2}
\expandafter\def\csname pn@etpnav.r2r.block.stall-mem1.stuck_event_count.hi\endcsname{$-$1.0}
\expandafter\def\csname pn@etpnav.r2r.block.stall-mem1.stuck_event_count.lo\endcsname{$-$1.3}
\expandafter\def\csname pn@etpnav.r2r.block.stall-mem1.stuck_event_count.off\endcsname{1.2}
\expandafter\def\csname pn@etpnav.r2r.block.stall-mem1.stuck_event_count.on\endcsname{0.0}
\expandafter\def\csname pn@etpnav.r2r.block.stall-mem1.stuck_event_count.rel\endcsname{$-$100}
\expandafter\def\csname pn@etpnav.r2r.block.stall-mem1.stuck_flag.d\endcsname{$-$43.8}
\expandafter\def\csname pn@etpnav.r2r.block.stall-mem1.stuck_flag.hi\endcsname{$-$41.5}
\expandafter\def\csname pn@etpnav.r2r.block.stall-mem1.stuck_flag.lo\endcsname{$-$45.9}
\expandafter\def\csname pn@etpnav.r2r.block.stall-mem1.stuck_flag.off\endcsname{43.8}
\expandafter\def\csname pn@etpnav.r2r.block.stall-mem1.stuck_flag.on\endcsname{0.0}
\expandafter\def\csname pn@etpnav.r2r.block.stall-mem1.stuck_flag.rel\endcsname{$-$100}
\expandafter\def\csname pn@etpnav.r2r.block.stall-mem1.success.d\endcsname{+10.7}
\expandafter\def\csname pn@etpnav.r2r.block.stall-mem1.success.hi\endcsname{+12.4}
\expandafter\def\csname pn@etpnav.r2r.block.stall-mem1.success.lo\endcsname{+9.0}
\expandafter\def\csname pn@etpnav.r2r.block.stall-mem1.success.off\endcsname{41.5}
\expandafter\def\csname pn@etpnav.r2r.block.stall-mem1.success.on\endcsname{52.2}
\expandafter\def\csname pn@etpnav.r2r.block.stall-mem1.success.rel\endcsname{+26}
\expandafter\def\csname pn@etpnav.r2r.block.stall-mem1.yields\endcsname{0.08}
\expandafter\def\csname pn@etpnav.r2r.block.stuck\endcsname{43.8}
\expandafter\def\csname pn@etpnav.r2r.block.tir\endcsname{24.3}
\expandafter\def\csname pn@etpnav.r2r.ddsr_pp\endcsname{+8.4}
\expandafter\def\csname pn@etpnav.r2r.dsr_pp\endcsname{+15.4}
\expandafter\def\csname pn@etpnav.r2r.hist.flagged\endcsname{35{,}301}
\expandafter\def\csname pn@etpnav.r2r.hist.recall\endcsname{91.1}
\expandafter\def\csname pn@etpnav.r2r.hist.tp\endcsname{35{,}301}
\expandafter\def\csname pn@etpnav.r2r.kmiss\endcsname{85.0}
\expandafter\def\csname pn@etpnav.r2r.slide.cleansuccess\endcsname{34.0}
\expandafter\def\csname pn@etpnav.r2r.slide.cpds\endcsname{6.2}
\expandafter\def\csname pn@etpnav.r2r.slide.cpdsmed\endcsname{4}
\expandafter\def\csname pn@etpnav.r2r.slide.dirtysuccess\endcsname{22.8}
\expandafter\def\csname pn@etpnav.r2r.slide.dsr\endcsname{40.2}
\expandafter\def\csname pn@etpnav.r2r.slide.dsrhi\endcsname{43.2}
\expandafter\def\csname pn@etpnav.r2r.slide.dsrlo\endcsname{37.2}
\expandafter\def\csname pn@etpnav.r2r.slide.nc\endcsname{4.9}
\expandafter\def\csname pn@etpnav.r2r.slide.ncfail\endcsname{7.0}
\expandafter\def\csname pn@etpnav.r2r.slide.ncsucc\endcsname{3.3}
\expandafter\def\csname pn@etpnav.r2r.slide.proxydsr\endcsname{3.3}
\expandafter\def\csname pn@etpnav.r2r.slide.shareinsucc\endcsname{31.6}
\expandafter\def\csname pn@etpnav.r2r.slide.sr\endcsname{56.9}
\expandafter\def\csname pn@etpnav.r2r.slide.stuck\endcsname{4.2}
\expandafter\def\csname pn@etpnav.r2r.slide.tir\endcsname{17.4}
\expandafter\def\csname pn@etpnav.r2r.swap10\endcsname{0.7}
\expandafter\def\csname pn@etpnav.r2r.swap50\endcsname{11.2}
\expandafter\def\csname pn@etpnav.rxr.auc.ent\endcsname{.521}
\expandafter\def\csname pn@etpnav.rxr.auc.entnorm\endcsname{.560}
\expandafter\def\csname pn@etpnav.rxr.auc.geom\endcsname{.826}
\expandafter\def\csname pn@etpnav.rxr.auc.hist\endcsname{.814}
\expandafter\def\csname pn@etpnav.rxr.auc.marg\endcsname{.505}
\expandafter\def\csname pn@etpnav.rxr.auc.stop\endcsname{.523}
\expandafter\def\csname pn@etpnav.rxr.avail10\endcsname{11.9}
\expandafter\def\csname pn@etpnav.rxr.avail50\endcsname{53.2}
\expandafter\def\csname pn@etpnav.rxr.block.cleansuccess\endcsname{8.7}
\expandafter\def\csname pn@etpnav.rxr.block.cpds\endcsname{53.2}
\expandafter\def\csname pn@etpnav.rxr.block.cpdsmed\endcsname{52}
\expandafter\def\csname pn@etpnav.rxr.block.dirtysuccess\endcsname{14.7}
\expandafter\def\csname pn@etpnav.rxr.block.dsr\endcsname{62.9}
\expandafter\def\csname pn@etpnav.rxr.block.dsrhi\endcsname{68.3}
\expandafter\def\csname pn@etpnav.rxr.block.dsrlo\endcsname{57.2}
\expandafter\def\csname pn@etpnav.rxr.block.nc\endcsname{57.3}
\expandafter\def\csname pn@etpnav.rxr.block.ncfail\endcsname{66.9}
\expandafter\def\csname pn@etpnav.rxr.block.ncsucc\endcsname{26.1}
\expandafter\def\csname pn@etpnav.rxr.block.proxydsr\endcsname{55.2}
\expandafter\def\csname pn@etpnav.rxr.block.shareinsucc\endcsname{9.5}
\expandafter\def\csname pn@etpnav.rxr.block.sr\endcsname{23.4}
\expandafter\def\csname pn@etpnav.rxr.block.stuck\endcsname{81.0}
\expandafter\def\csname pn@etpnav.rxr.block.tir\endcsname{47.9}
\expandafter\def\csname pn@etpnav.rxr.ddsr_pp\endcsname{+15.4}
\expandafter\def\csname pn@etpnav.rxr.dsr_pp\endcsname{+34.6}
\expandafter\def\csname pn@etpnav.rxr.hist.flagged\endcsname{96{,}769}
\expandafter\def\csname pn@etpnav.rxr.hist.recall\endcsname{95.7}
\expandafter\def\csname pn@etpnav.rxr.hist.tp\endcsname{96{,}769}
\expandafter\def\csname pn@etpnav.rxr.kmiss\endcsname{86.0}
\expandafter\def\csname pn@etpnav.rxr.slide.cleansuccess\endcsname{12.6}
\expandafter\def\csname pn@etpnav.rxr.slide.cpds\endcsname{6.6}
\expandafter\def\csname pn@etpnav.rxr.slide.cpdsmed\endcsname{4}
\expandafter\def\csname pn@etpnav.rxr.slide.dirtysuccess\endcsname{45.4}
\expandafter\def\csname pn@etpnav.rxr.slide.dsr\endcsname{78.3}
\expandafter\def\csname pn@etpnav.rxr.slide.dsrhi\endcsname{81.2}
\expandafter\def\csname pn@etpnav.rxr.slide.dsrlo\endcsname{75.1}
\expandafter\def\csname pn@etpnav.rxr.slide.nc\endcsname{10.0}
\expandafter\def\csname pn@etpnav.rxr.slide.ncfail\endcsname{14.5}
\expandafter\def\csname pn@etpnav.rxr.slide.ncsucc\endcsname{6.8}
\expandafter\def\csname pn@etpnav.rxr.slide.proxydsr\endcsname{2.7}
\expandafter\def\csname pn@etpnav.rxr.slide.shareinsucc\endcsname{24.8}
\expandafter\def\csname pn@etpnav.rxr.slide.sr\endcsname{58.0}
\expandafter\def\csname pn@etpnav.rxr.slide.stuck\endcsname{8.1}
\expandafter\def\csname pn@etpnav.rxr.slide.tir\endcsname{16.4}
\expandafter\def\csname pn@etpnav.rxr.swap10\endcsname{2.6}
\expandafter\def\csname pn@etpnav.rxr.swap50\endcsname{21.0}
\expandafter\def\csname pn@etpr1.mem.M1.false\endcsname{0.10}
\expandafter\def\csname pn@etpr1.mem.M1.prec\endcsname{92.5}
\expandafter\def\csname pn@etpr1.mem.M1.recall\endcsname{36.8}
\expandafter\def\csname pn@etpr1.mem.M8.false\endcsname{13.51}
\expandafter\def\csname pn@etpr1.mem.M8.prec\endcsname{12.5}
\expandafter\def\csname pn@etpr1.mem.M8.recall\endcsname{59.2}
\expandafter\def\csname pn@etpr1.mem.episodes\endcsname{154}
\expandafter\def\csname pn@etpr1.r2r.auc.ent\endcsname{.646}
\expandafter\def\csname pn@etpr1.r2r.auc.entnorm\endcsname{.673}
\expandafter\def\csname pn@etpr1.r2r.auc.geom\endcsname{.811}
\expandafter\def\csname pn@etpr1.r2r.auc.hist\endcsname{.891}
\expandafter\def\csname pn@etpr1.r2r.auc.marg\endcsname{.637}
\expandafter\def\csname pn@etpr1.r2r.auc.stop\endcsname{.416}
\expandafter\def\csname pn@etpr1.r2r.avail10\endcsname{24.1}
\expandafter\def\csname pn@etpr1.r2r.avail50\endcsname{64.7}
\expandafter\def\csname pn@etpr1.r2r.block.cleansuccess\endcsname{29.1}
\expandafter\def\csname pn@etpr1.r2r.block.cpds\endcsname{16.4}
\expandafter\def\csname pn@etpr1.r2r.block.cpdsmed\endcsname{9}
\expandafter\def\csname pn@etpr1.r2r.block.default-exempt.abortep\endcsname{95}
\expandafter\def\csname pn@etpr1.r2r.block.default-exempt.aborts\endcsname{5.9}
\expandafter\def\csname pn@etpr1.r2r.block.default-exempt.caphit.off\endcsname{38.4}
\expandafter\def\csname pn@etpr1.r2r.block.default-exempt.caphit.on\endcsname{43.4}
\expandafter\def\csname pn@etpr1.r2r.block.default-exempt.clean_success.d\endcsname{+15.9}
\expandafter\def\csname pn@etpr1.r2r.block.default-exempt.clean_success.hi\endcsname{+18.1}
\expandafter\def\csname pn@etpr1.r2r.block.default-exempt.clean_success.lo\endcsname{+13.8}
\expandafter\def\csname pn@etpr1.r2r.block.default-exempt.clean_success.off\endcsname{29.1}
\expandafter\def\csname pn@etpr1.r2r.block.default-exempt.clean_success.on\endcsname{45.0}
\expandafter\def\csname pn@etpr1.r2r.block.default-exempt.clean_success.rel\endcsname{+54}
\expandafter\def\csname pn@etpr1.r2r.block.default-exempt.collision_count.d\endcsname{$-$3.6}
\expandafter\def\csname pn@etpr1.r2r.block.default-exempt.collision_count.hi\endcsname{$-$2.5}
\expandafter\def\csname pn@etpr1.r2r.block.default-exempt.collision_count.lo\endcsname{$-$4.8}
\expandafter\def\csname pn@etpr1.r2r.block.default-exempt.collision_count.off\endcsname{18.1}
\expandafter\def\csname pn@etpr1.r2r.block.default-exempt.collision_count.on\endcsname{14.5}
\expandafter\def\csname pn@etpr1.r2r.block.default-exempt.collision_count.rel\endcsname{$-$20}
\expandafter\def\csname pn@etpr1.r2r.block.default-exempt.collision_event_count.d\endcsname{$-$0.36}
\expandafter\def\csname pn@etpr1.r2r.block.default-exempt.collision_event_count.hi\endcsname{$-$0.15}
\expandafter\def\csname pn@etpr1.r2r.block.default-exempt.collision_event_count.lo\endcsname{$-$0.55}
\expandafter\def\csname pn@etpr1.r2r.block.default-exempt.collision_event_count.off\endcsname{2.60}
\expandafter\def\csname pn@etpr1.r2r.block.default-exempt.collision_event_count.on\endcsname{2.24}
\expandafter\def\csname pn@etpr1.r2r.block.default-exempt.collision_event_count.rel\endcsname{$-$14}
\expandafter\def\csname pn@etpr1.r2r.block.default-exempt.dirty_success_flag.d\endcsname{$-$12.0}
\expandafter\def\csname pn@etpr1.r2r.block.default-exempt.dirty_success_flag.hi\endcsname{$-$9.9}
\expandafter\def\csname pn@etpr1.r2r.block.default-exempt.dirty_success_flag.lo\endcsname{$-$14.0}
\expandafter\def\csname pn@etpr1.r2r.block.default-exempt.dirty_success_flag.off\endcsname{21.4}
\expandafter\def\csname pn@etpr1.r2r.block.default-exempt.dirty_success_flag.on\endcsname{9.4}
\expandafter\def\csname pn@etpr1.r2r.block.default-exempt.dirty_success_flag.rel\endcsname{$-$56}
\expandafter\def\csname pn@etpr1.r2r.block.default-exempt.gained\endcsname{227}
\expandafter\def\csname pn@etpr1.r2r.block.default-exempt.lost\endcsname{156}
\expandafter\def\csname pn@etpr1.r2r.block.default-exempt.navigation_error_m.d\endcsname{$-$0.3}
\expandafter\def\csname pn@etpr1.r2r.block.default-exempt.navigation_error_m.hi\endcsname{$-$0.2}
\expandafter\def\csname pn@etpr1.r2r.block.default-exempt.navigation_error_m.lo\endcsname{$-$0.4}
\expandafter\def\csname pn@etpr1.r2r.block.default-exempt.navigation_error_m.off\endcsname{4.8}
\expandafter\def\csname pn@etpr1.r2r.block.default-exempt.navigation_error_m.on\endcsname{4.5}
\expandafter\def\csname pn@etpr1.r2r.block.default-exempt.navigation_error_m.rel\endcsname{$-$6}
\expandafter\def\csname pn@etpr1.r2r.block.default-exempt.nc_rate.d\endcsname{$-$6.2}
\expandafter\def\csname pn@etpr1.r2r.block.default-exempt.nc_rate.hi\endcsname{$-$5.2}
\expandafter\def\csname pn@etpr1.r2r.block.default-exempt.nc_rate.lo\endcsname{$-$7.1}
\expandafter\def\csname pn@etpr1.r2r.block.default-exempt.nc_rate.off\endcsname{13.9}
\expandafter\def\csname pn@etpr1.r2r.block.default-exempt.nc_rate.on\endcsname{7.7}
\expandafter\def\csname pn@etpr1.r2r.block.default-exempt.nc_rate.rel\endcsname{$-$44}
\expandafter\def\csname pn@etpr1.r2r.block.default-exempt.ndtw.d\endcsname{+0.2}
\expandafter\def\csname pn@etpr1.r2r.block.default-exempt.ndtw.hi\endcsname{+1.2}
\expandafter\def\csname pn@etpr1.r2r.block.default-exempt.ndtw.lo\endcsname{$-$0.7}
\expandafter\def\csname pn@etpr1.r2r.block.default-exempt.ndtw.off\endcsname{60.9}
\expandafter\def\csname pn@etpr1.r2r.block.default-exempt.ndtw.on\endcsname{61.2}
\expandafter\def\csname pn@etpr1.r2r.block.default-exempt.ndtw.rel\endcsname{+0}
\expandafter\def\csname pn@etpr1.r2r.block.default-exempt.overcap\endcsname{0}
\expandafter\def\csname pn@etpr1.r2r.block.default-exempt.owned\endcsname{17.6}
\expandafter\def\csname pn@etpr1.r2r.block.default-exempt.resid.maneuver\endcsname{2}
\expandafter\def\csname pn@etpr1.r2r.block.default-exempt.resid.pass\endcsname{80}
\expandafter\def\csname pn@etpr1.r2r.block.default-exempt.spl.d\endcsname{+1.3}
\expandafter\def\csname pn@etpr1.r2r.block.default-exempt.spl.hi\endcsname{+3.0}
\expandafter\def\csname pn@etpr1.r2r.block.default-exempt.spl.lo\endcsname{$-$0.5}
\expandafter\def\csname pn@etpr1.r2r.block.default-exempt.spl.off\endcsname{43.3}
\expandafter\def\csname pn@etpr1.r2r.block.default-exempt.spl.on\endcsname{44.6}
\expandafter\def\csname pn@etpr1.r2r.block.default-exempt.spl.rel\endcsname{+3}
\expandafter\def\csname pn@etpr1.r2r.block.default-exempt.stalls\endcsname{0.0}
\expandafter\def\csname pn@etpr1.r2r.block.default-exempt.step_count.d\endcsname{+60.8}
\expandafter\def\csname pn@etpr1.r2r.block.default-exempt.step_count.hi\endcsname{+65.6}
\expandafter\def\csname pn@etpr1.r2r.block.default-exempt.step_count.lo\endcsname{+55.7}
\expandafter\def\csname pn@etpr1.r2r.block.default-exempt.step_count.off\endcsname{111.6}
\expandafter\def\csname pn@etpr1.r2r.block.default-exempt.step_count.on\endcsname{172.3}
\expandafter\def\csname pn@etpr1.r2r.block.default-exempt.step_count.rel\endcsname{+54}
\expandafter\def\csname pn@etpr1.r2r.block.default-exempt.stuck_event_count.d\endcsname{$-$0.3}
\expandafter\def\csname pn@etpr1.r2r.block.default-exempt.stuck_event_count.hi\endcsname{$-$0.2}
\expandafter\def\csname pn@etpr1.r2r.block.default-exempt.stuck_event_count.lo\endcsname{$-$0.4}
\expandafter\def\csname pn@etpr1.r2r.block.default-exempt.stuck_event_count.off\endcsname{1.2}
\expandafter\def\csname pn@etpr1.r2r.block.default-exempt.stuck_event_count.on\endcsname{0.9}
\expandafter\def\csname pn@etpr1.r2r.block.default-exempt.stuck_event_count.rel\endcsname{$-$27}
\expandafter\def\csname pn@etpr1.r2r.block.default-exempt.stuck_flag.d\endcsname{$-$14.0}
\expandafter\def\csname pn@etpr1.r2r.block.default-exempt.stuck_flag.hi\endcsname{$-$11.6}
\expandafter\def\csname pn@etpr1.r2r.block.default-exempt.stuck_flag.lo\endcsname{$-$16.4}
\expandafter\def\csname pn@etpr1.r2r.block.default-exempt.stuck_flag.off\endcsname{44.4}
\expandafter\def\csname pn@etpr1.r2r.block.default-exempt.stuck_flag.on\endcsname{30.5}
\expandafter\def\csname pn@etpr1.r2r.block.default-exempt.stuck_flag.rel\endcsname{$-$31}
\expandafter\def\csname pn@etpr1.r2r.block.default-exempt.success.d\endcsname{+3.9}
\expandafter\def\csname pn@etpr1.r2r.block.default-exempt.success.hi\endcsname{+5.9}
\expandafter\def\csname pn@etpr1.r2r.block.default-exempt.success.lo\endcsname{+1.8}
\expandafter\def\csname pn@etpr1.r2r.block.default-exempt.success.off\endcsname{50.5}
\expandafter\def\csname pn@etpr1.r2r.block.default-exempt.success.on\endcsname{54.4}
\expandafter\def\csname pn@etpr1.r2r.block.default-exempt.success.rel\endcsname{+8}
\expandafter\def\csname pn@etpr1.r2r.block.default-exempt.yields\endcsname{4.64}
\expandafter\def\csname pn@etpr1.r2r.block.default.abortep\endcsname{95}
\expandafter\def\csname pn@etpr1.r2r.block.default.aborts\endcsname{5.2}
\expandafter\def\csname pn@etpr1.r2r.block.default.caphit.off\endcsname{38.4}
\expandafter\def\csname pn@etpr1.r2r.block.default.caphit.on\endcsname{43.4}
\expandafter\def\csname pn@etpr1.r2r.block.default.clean_success.d\endcsname{+15.0}
\expandafter\def\csname pn@etpr1.r2r.block.default.clean_success.hi\endcsname{+17.1}
\expandafter\def\csname pn@etpr1.r2r.block.default.clean_success.lo\endcsname{+12.9}
\expandafter\def\csname pn@etpr1.r2r.block.default.clean_success.off\endcsname{29.1}
\expandafter\def\csname pn@etpr1.r2r.block.default.clean_success.on\endcsname{44.1}
\expandafter\def\csname pn@etpr1.r2r.block.default.clean_success.rel\endcsname{+51}
\expandafter\def\csname pn@etpr1.r2r.block.default.collision_count.d\endcsname{$-$8.7}
\expandafter\def\csname pn@etpr1.r2r.block.default.collision_count.hi\endcsname{$-$7.6}
\expandafter\def\csname pn@etpr1.r2r.block.default.collision_count.lo\endcsname{$-$9.7}
\expandafter\def\csname pn@etpr1.r2r.block.default.collision_count.off\endcsname{18.1}
\expandafter\def\csname pn@etpr1.r2r.block.default.collision_count.on\endcsname{9.5}
\expandafter\def\csname pn@etpr1.r2r.block.default.collision_count.rel\endcsname{$-$48}
\expandafter\def\csname pn@etpr1.r2r.block.default.collision_event_count.d\endcsname{$-$1.24}
\expandafter\def\csname pn@etpr1.r2r.block.default.collision_event_count.hi\endcsname{$-$1.08}
\expandafter\def\csname pn@etpr1.r2r.block.default.collision_event_count.lo\endcsname{$-$1.39}
\expandafter\def\csname pn@etpr1.r2r.block.default.collision_event_count.off\endcsname{2.60}
\expandafter\def\csname pn@etpr1.r2r.block.default.collision_event_count.on\endcsname{1.36}
\expandafter\def\csname pn@etpr1.r2r.block.default.collision_event_count.rel\endcsname{$-$48}
\expandafter\def\csname pn@etpr1.r2r.block.default.dirty_success_flag.d\endcsname{$-$15.5}
\expandafter\def\csname pn@etpr1.r2r.block.default.dirty_success_flag.hi\endcsname{$-$13.5}
\expandafter\def\csname pn@etpr1.r2r.block.default.dirty_success_flag.lo\endcsname{$-$17.4}
\expandafter\def\csname pn@etpr1.r2r.block.default.dirty_success_flag.off\endcsname{21.4}
\expandafter\def\csname pn@etpr1.r2r.block.default.dirty_success_flag.on\endcsname{5.9}
\expandafter\def\csname pn@etpr1.r2r.block.default.dirty_success_flag.rel\endcsname{$-$73}
\expandafter\def\csname pn@etpr1.r2r.block.default.gained\endcsname{187}
\expandafter\def\csname pn@etpr1.r2r.block.default.lost\endcsname{197}
\expandafter\def\csname pn@etpr1.r2r.block.default.navigation_error_m.d\endcsname{$-$0.1}
\expandafter\def\csname pn@etpr1.r2r.block.default.navigation_error_m.hi\endcsname{+0.1}
\expandafter\def\csname pn@etpr1.r2r.block.default.navigation_error_m.lo\endcsname{$-$0.2}
\expandafter\def\csname pn@etpr1.r2r.block.default.navigation_error_m.off\endcsname{4.8}
\expandafter\def\csname pn@etpr1.r2r.block.default.navigation_error_m.on\endcsname{4.8}
\expandafter\def\csname pn@etpr1.r2r.block.default.navigation_error_m.rel\endcsname{$-$1}
\expandafter\def\csname pn@etpr1.r2r.block.default.nc_rate.d\endcsname{$-$6.7}
\expandafter\def\csname pn@etpr1.r2r.block.default.nc_rate.hi\endcsname{$-$5.7}
\expandafter\def\csname pn@etpr1.r2r.block.default.nc_rate.lo\endcsname{$-$7.6}
\expandafter\def\csname pn@etpr1.r2r.block.default.nc_rate.off\endcsname{13.9}
\expandafter\def\csname pn@etpr1.r2r.block.default.nc_rate.on\endcsname{7.2}
\expandafter\def\csname pn@etpr1.r2r.block.default.nc_rate.rel\endcsname{$-$48}
\expandafter\def\csname pn@etpr1.r2r.block.default.ndtw.d\endcsname{+1.5}
\expandafter\def\csname pn@etpr1.r2r.block.default.ndtw.hi\endcsname{+2.4}
\expandafter\def\csname pn@etpr1.r2r.block.default.ndtw.lo\endcsname{+0.6}
\expandafter\def\csname pn@etpr1.r2r.block.default.ndtw.off\endcsname{60.9}
\expandafter\def\csname pn@etpr1.r2r.block.default.ndtw.on\endcsname{62.4}
\expandafter\def\csname pn@etpr1.r2r.block.default.ndtw.rel\endcsname{+2}
\expandafter\def\csname pn@etpr1.r2r.block.default.overcap\endcsname{0}
\expandafter\def\csname pn@etpr1.r2r.block.default.owned\endcsname{19.9}
\expandafter\def\csname pn@etpr1.r2r.block.default.resid.maneuver\endcsname{3}
\expandafter\def\csname pn@etpr1.r2r.block.default.resid.pass\endcsname{90}
\expandafter\def\csname pn@etpr1.r2r.block.default.spl.d\endcsname{$-$0.7}
\expandafter\def\csname pn@etpr1.r2r.block.default.spl.hi\endcsname{+1.0}
\expandafter\def\csname pn@etpr1.r2r.block.default.spl.lo\endcsname{$-$2.6}
\expandafter\def\csname pn@etpr1.r2r.block.default.spl.off\endcsname{43.3}
\expandafter\def\csname pn@etpr1.r2r.block.default.spl.on\endcsname{42.6}
\expandafter\def\csname pn@etpr1.r2r.block.default.spl.rel\endcsname{$-$2}
\expandafter\def\csname pn@etpr1.r2r.block.default.stalls\endcsname{0.0}
\expandafter\def\csname pn@etpr1.r2r.block.default.step_count.d\endcsname{+23.8}
\expandafter\def\csname pn@etpr1.r2r.block.default.step_count.hi\endcsname{+27.1}
\expandafter\def\csname pn@etpr1.r2r.block.default.step_count.lo\endcsname{+20.6}
\expandafter\def\csname pn@etpr1.r2r.block.default.step_count.off\endcsname{111.6}
\expandafter\def\csname pn@etpr1.r2r.block.default.step_count.on\endcsname{135.4}
\expandafter\def\csname pn@etpr1.r2r.block.default.step_count.rel\endcsname{+21}
\expandafter\def\csname pn@etpr1.r2r.block.default.stuck_event_count.d\endcsname{$-$0.6}
\expandafter\def\csname pn@etpr1.r2r.block.default.stuck_event_count.hi\endcsname{$-$0.5}
\expandafter\def\csname pn@etpr1.r2r.block.default.stuck_event_count.lo\endcsname{$-$0.7}
\expandafter\def\csname pn@etpr1.r2r.block.default.stuck_event_count.off\endcsname{1.2}
\expandafter\def\csname pn@etpr1.r2r.block.default.stuck_event_count.on\endcsname{0.5}
\expandafter\def\csname pn@etpr1.r2r.block.default.stuck_event_count.rel\endcsname{$-$54}
\expandafter\def\csname pn@etpr1.r2r.block.default.stuck_flag.d\endcsname{$-$20.4}
\expandafter\def\csname pn@etpr1.r2r.block.default.stuck_flag.hi\endcsname{$-$18.0}
\expandafter\def\csname pn@etpr1.r2r.block.default.stuck_flag.lo\endcsname{$-$22.7}
\expandafter\def\csname pn@etpr1.r2r.block.default.stuck_flag.off\endcsname{44.4}
\expandafter\def\csname pn@etpr1.r2r.block.default.stuck_flag.on\endcsname{24.0}
\expandafter\def\csname pn@etpr1.r2r.block.default.stuck_flag.rel\endcsname{$-$46}
\expandafter\def\csname pn@etpr1.r2r.block.default.success.d\endcsname{$-$0.5}
\expandafter\def\csname pn@etpr1.r2r.block.default.success.hi\endcsname{+1.5}
\expandafter\def\csname pn@etpr1.r2r.block.default.success.lo\endcsname{$-$2.6}
\expandafter\def\csname pn@etpr1.r2r.block.default.success.off\endcsname{50.5}
\expandafter\def\csname pn@etpr1.r2r.block.default.success.on\endcsname{50.0}
\expandafter\def\csname pn@etpr1.r2r.block.default.success.rel\endcsname{$-$1}
\expandafter\def\csname pn@etpr1.r2r.block.default.yields\endcsname{0.91}
\expandafter\def\csname pn@etpr1.r2r.block.dirtysuccess\endcsname{21.4}
\expandafter\def\csname pn@etpr1.r2r.block.dsr\endcsname{42.3}
\expandafter\def\csname pn@etpr1.r2r.block.dsrhi\endcsname{45.5}
\expandafter\def\csname pn@etpr1.r2r.block.dsrlo\endcsname{39.2}
\expandafter\def\csname pn@etpr1.r2r.block.nc\endcsname{13.9}
\expandafter\def\csname pn@etpr1.r2r.block.ncfail\endcsname{21.8}
\expandafter\def\csname pn@etpr1.r2r.block.ncsucc\endcsname{6.2}
\expandafter\def\csname pn@etpr1.r2r.block.proxydsr\endcsname{38.4}
\expandafter\def\csname pn@etpr1.r2r.block.shareinsucc\endcsname{19.3}
\expandafter\def\csname pn@etpr1.r2r.block.sr\endcsname{50.5}
\expandafter\def\csname pn@etpr1.r2r.block.stall-mem1.abortep\endcsname{39}
\expandafter\def\csname pn@etpr1.r2r.block.stall-mem1.aborts\endcsname{0.9}
\expandafter\def\csname pn@etpr1.r2r.block.stall-mem1.caphit.off\endcsname{38.4}
\expandafter\def\csname pn@etpr1.r2r.block.stall-mem1.caphit.on\endcsname{28.6}
\expandafter\def\csname pn@etpr1.r2r.block.stall-mem1.clean_success.d\endcsname{+10.1}
\expandafter\def\csname pn@etpr1.r2r.block.stall-mem1.clean_success.hi\endcsname{+11.4}
\expandafter\def\csname pn@etpr1.r2r.block.stall-mem1.clean_success.lo\endcsname{+8.7}
\expandafter\def\csname pn@etpr1.r2r.block.stall-mem1.clean_success.off\endcsname{29.1}
\expandafter\def\csname pn@etpr1.r2r.block.stall-mem1.clean_success.on\endcsname{39.2}
\expandafter\def\csname pn@etpr1.r2r.block.stall-mem1.clean_success.rel\endcsname{+35}
\expandafter\def\csname pn@etpr1.r2r.block.stall-mem1.collision_count.d\endcsname{$-$15.0}
\expandafter\def\csname pn@etpr1.r2r.block.stall-mem1.collision_count.hi\endcsname{$-$14.0}
\expandafter\def\csname pn@etpr1.r2r.block.stall-mem1.collision_count.lo\endcsname{$-$15.9}
\expandafter\def\csname pn@etpr1.r2r.block.stall-mem1.collision_count.off\endcsname{18.1}
\expandafter\def\csname pn@etpr1.r2r.block.stall-mem1.collision_count.on\endcsname{3.2}
\expandafter\def\csname pn@etpr1.r2r.block.stall-mem1.collision_count.rel\endcsname{$-$83}
\expandafter\def\csname pn@etpr1.r2r.block.stall-mem1.collision_event_count.d\endcsname{+0.05}
\expandafter\def\csname pn@etpr1.r2r.block.stall-mem1.collision_event_count.hi\endcsname{+0.22}
\expandafter\def\csname pn@etpr1.r2r.block.stall-mem1.collision_event_count.lo\endcsname{$-$0.11}
\expandafter\def\csname pn@etpr1.r2r.block.stall-mem1.collision_event_count.off\endcsname{2.60}
\expandafter\def\csname pn@etpr1.r2r.block.stall-mem1.collision_event_count.on\endcsname{2.65}
\expandafter\def\csname pn@etpr1.r2r.block.stall-mem1.collision_event_count.rel\endcsname{+2}
\expandafter\def\csname pn@etpr1.r2r.block.stall-mem1.dirty_success_flag.d\endcsname{$-$3.2}
\expandafter\def\csname pn@etpr1.r2r.block.stall-mem1.dirty_success_flag.hi\endcsname{$-$1.3}
\expandafter\def\csname pn@etpr1.r2r.block.stall-mem1.dirty_success_flag.lo\endcsname{$-$5.1}
\expandafter\def\csname pn@etpr1.r2r.block.stall-mem1.dirty_success_flag.off\endcsname{21.4}
\expandafter\def\csname pn@etpr1.r2r.block.stall-mem1.dirty_success_flag.on\endcsname{18.2}
\expandafter\def\csname pn@etpr1.r2r.block.stall-mem1.dirty_success_flag.rel\endcsname{$-$15}
\expandafter\def\csname pn@etpr1.r2r.block.stall-mem1.gained\endcsname{173}
\expandafter\def\csname pn@etpr1.r2r.block.stall-mem1.lost\endcsname{46}
\expandafter\def\csname pn@etpr1.r2r.block.stall-mem1.navigation_error_m.d\endcsname{$-$0.4}
\expandafter\def\csname pn@etpr1.r2r.block.stall-mem1.navigation_error_m.hi\endcsname{$-$0.3}
\expandafter\def\csname pn@etpr1.r2r.block.stall-mem1.navigation_error_m.lo\endcsname{$-$0.5}
\expandafter\def\csname pn@etpr1.r2r.block.stall-mem1.navigation_error_m.off\endcsname{4.8}
\expandafter\def\csname pn@etpr1.r2r.block.stall-mem1.navigation_error_m.on\endcsname{4.4}
\expandafter\def\csname pn@etpr1.r2r.block.stall-mem1.navigation_error_m.rel\endcsname{$-$9}
\expandafter\def\csname pn@etpr1.r2r.block.stall-mem1.nc_rate.d\endcsname{$-$11.4}
\expandafter\def\csname pn@etpr1.r2r.block.stall-mem1.nc_rate.hi\endcsname{$-$10.7}
\expandafter\def\csname pn@etpr1.r2r.block.stall-mem1.nc_rate.lo\endcsname{$-$12.3}
\expandafter\def\csname pn@etpr1.r2r.block.stall-mem1.nc_rate.off\endcsname{13.9}
\expandafter\def\csname pn@etpr1.r2r.block.stall-mem1.nc_rate.on\endcsname{2.5}
\expandafter\def\csname pn@etpr1.r2r.block.stall-mem1.nc_rate.rel\endcsname{$-$82}
\expandafter\def\csname pn@etpr1.r2r.block.stall-mem1.ndtw.d\endcsname{+3.4}
\expandafter\def\csname pn@etpr1.r2r.block.stall-mem1.ndtw.hi\endcsname{+4.1}
\expandafter\def\csname pn@etpr1.r2r.block.stall-mem1.ndtw.lo\endcsname{+2.7}
\expandafter\def\csname pn@etpr1.r2r.block.stall-mem1.ndtw.off\endcsname{60.9}
\expandafter\def\csname pn@etpr1.r2r.block.stall-mem1.ndtw.on\endcsname{64.3}
\expandafter\def\csname pn@etpr1.r2r.block.stall-mem1.ndtw.rel\endcsname{+6}
\expandafter\def\csname pn@etpr1.r2r.block.stall-mem1.overcap\endcsname{0}
\expandafter\def\csname pn@etpr1.r2r.block.stall-mem1.owned\endcsname{8.4}
\expandafter\def\csname pn@etpr1.r2r.block.stall-mem1.resid.maneuver\endcsname{49}
\expandafter\def\csname pn@etpr1.r2r.block.stall-mem1.resid.pass\endcsname{47}
\expandafter\def\csname pn@etpr1.r2r.block.stall-mem1.spl.d\endcsname{+6.3}
\expandafter\def\csname pn@etpr1.r2r.block.stall-mem1.spl.hi\endcsname{+7.5}
\expandafter\def\csname pn@etpr1.r2r.block.stall-mem1.spl.lo\endcsname{+5.1}
\expandafter\def\csname pn@etpr1.r2r.block.stall-mem1.spl.off\endcsname{43.3}
\expandafter\def\csname pn@etpr1.r2r.block.stall-mem1.spl.on\endcsname{49.7}
\expandafter\def\csname pn@etpr1.r2r.block.stall-mem1.spl.rel\endcsname{+15}
\expandafter\def\csname pn@etpr1.r2r.block.stall-mem1.stalls\endcsname{1.9}
\expandafter\def\csname pn@etpr1.r2r.block.stall-mem1.step_count.d\endcsname{$-$9.3}
\expandafter\def\csname pn@etpr1.r2r.block.stall-mem1.step_count.hi\endcsname{$-$7.0}
\expandafter\def\csname pn@etpr1.r2r.block.stall-mem1.step_count.lo\endcsname{$-$11.8}
\expandafter\def\csname pn@etpr1.r2r.block.stall-mem1.step_count.off\endcsname{111.6}
\expandafter\def\csname pn@etpr1.r2r.block.stall-mem1.step_count.on\endcsname{102.2}
\expandafter\def\csname pn@etpr1.r2r.block.stall-mem1.step_count.rel\endcsname{$-$8}
\expandafter\def\csname pn@etpr1.r2r.block.stall-mem1.stuck_event_count.d\endcsname{$-$1.2}
\expandafter\def\csname pn@etpr1.r2r.block.stall-mem1.stuck_event_count.hi\endcsname{$-$1.1}
\expandafter\def\csname pn@etpr1.r2r.block.stall-mem1.stuck_event_count.lo\endcsname{$-$1.3}
\expandafter\def\csname pn@etpr1.r2r.block.stall-mem1.stuck_event_count.off\endcsname{1.2}
\expandafter\def\csname pn@etpr1.r2r.block.stall-mem1.stuck_event_count.on\endcsname{0.0}
\expandafter\def\csname pn@etpr1.r2r.block.stall-mem1.stuck_event_count.rel\endcsname{$-$100}
\expandafter\def\csname pn@etpr1.r2r.block.stall-mem1.stuck_flag.d\endcsname{$-$44.4}
\expandafter\def\csname pn@etpr1.r2r.block.stall-mem1.stuck_flag.hi\endcsname{$-$42.1}
\expandafter\def\csname pn@etpr1.r2r.block.stall-mem1.stuck_flag.lo\endcsname{$-$46.7}
\expandafter\def\csname pn@etpr1.r2r.block.stall-mem1.stuck_flag.off\endcsname{44.4}
\expandafter\def\csname pn@etpr1.r2r.block.stall-mem1.stuck_flag.on\endcsname{0.0}
\expandafter\def\csname pn@etpr1.r2r.block.stall-mem1.stuck_flag.rel\endcsname{$-$100}
\expandafter\def\csname pn@etpr1.r2r.block.stall-mem1.success.d\endcsname{+6.9}
\expandafter\def\csname pn@etpr1.r2r.block.stall-mem1.success.hi\endcsname{+8.4}
\expandafter\def\csname pn@etpr1.r2r.block.stall-mem1.success.lo\endcsname{+5.3}
\expandafter\def\csname pn@etpr1.r2r.block.stall-mem1.success.off\endcsname{50.5}
\expandafter\def\csname pn@etpr1.r2r.block.stall-mem1.success.on\endcsname{57.4}
\expandafter\def\csname pn@etpr1.r2r.block.stall-mem1.success.rel\endcsname{+14}
\expandafter\def\csname pn@etpr1.r2r.block.stall-mem1.yields\endcsname{0.10}
\expandafter\def\csname pn@etpr1.r2r.block.stall.abortep\endcsname{96}
\expandafter\def\csname pn@etpr1.r2r.block.stall.aborts\endcsname{5.4}
\expandafter\def\csname pn@etpr1.r2r.block.stall.caphit.off\endcsname{38.4}
\expandafter\def\csname pn@etpr1.r2r.block.stall.caphit.on\endcsname{38.8}
\expandafter\def\csname pn@etpr1.r2r.block.stall.clean_success.d\endcsname{+15.0}
\expandafter\def\csname pn@etpr1.r2r.block.stall.clean_success.hi\endcsname{+17.1}
\expandafter\def\csname pn@etpr1.r2r.block.stall.clean_success.lo\endcsname{+12.9}
\expandafter\def\csname pn@etpr1.r2r.block.stall.clean_success.off\endcsname{29.1}
\expandafter\def\csname pn@etpr1.r2r.block.stall.clean_success.on\endcsname{44.1}
\expandafter\def\csname pn@etpr1.r2r.block.stall.clean_success.rel\endcsname{+51}
\expandafter\def\csname pn@etpr1.r2r.block.stall.collision_count.d\endcsname{$-$16.0}
\expandafter\def\csname pn@etpr1.r2r.block.stall.collision_count.hi\endcsname{$-$15.0}
\expandafter\def\csname pn@etpr1.r2r.block.stall.collision_count.lo\endcsname{$-$17.0}
\expandafter\def\csname pn@etpr1.r2r.block.stall.collision_count.off\endcsname{18.1}
\expandafter\def\csname pn@etpr1.r2r.block.stall.collision_count.on\endcsname{2.1}
\expandafter\def\csname pn@etpr1.r2r.block.stall.collision_count.rel\endcsname{$-$88}
\expandafter\def\csname pn@etpr1.r2r.block.stall.collision_event_count.d\endcsname{$-$0.79}
\expandafter\def\csname pn@etpr1.r2r.block.stall.collision_event_count.hi\endcsname{$-$0.61}
\expandafter\def\csname pn@etpr1.r2r.block.stall.collision_event_count.lo\endcsname{$-$0.97}
\expandafter\def\csname pn@etpr1.r2r.block.stall.collision_event_count.off\endcsname{2.60}
\expandafter\def\csname pn@etpr1.r2r.block.stall.collision_event_count.on\endcsname{1.81}
\expandafter\def\csname pn@etpr1.r2r.block.stall.collision_event_count.rel\endcsname{$-$30}
\expandafter\def\csname pn@etpr1.r2r.block.stall.dirty_success_flag.d\endcsname{$-$10.3}
\expandafter\def\csname pn@etpr1.r2r.block.stall.dirty_success_flag.hi\endcsname{$-$8.2}
\expandafter\def\csname pn@etpr1.r2r.block.stall.dirty_success_flag.lo\endcsname{$-$12.4}
\expandafter\def\csname pn@etpr1.r2r.block.stall.dirty_success_flag.off\endcsname{21.4}
\expandafter\def\csname pn@etpr1.r2r.block.stall.dirty_success_flag.on\endcsname{11.1}
\expandafter\def\csname pn@etpr1.r2r.block.stall.dirty_success_flag.rel\endcsname{$-$48}
\expandafter\def\csname pn@etpr1.r2r.block.stall.gained\endcsname{235}
\expandafter\def\csname pn@etpr1.r2r.block.stall.lost\endcsname{149}
\expandafter\def\csname pn@etpr1.r2r.block.stall.navigation_error_m.d\endcsname{$-$0.4}
\expandafter\def\csname pn@etpr1.r2r.block.stall.navigation_error_m.hi\endcsname{$-$0.2}
\expandafter\def\csname pn@etpr1.r2r.block.stall.navigation_error_m.lo\endcsname{$-$0.5}
\expandafter\def\csname pn@etpr1.r2r.block.stall.navigation_error_m.off\endcsname{4.8}
\expandafter\def\csname pn@etpr1.r2r.block.stall.navigation_error_m.on\endcsname{4.4}
\expandafter\def\csname pn@etpr1.r2r.block.stall.navigation_error_m.rel\endcsname{$-$8}
\expandafter\def\csname pn@etpr1.r2r.block.stall.nc_rate.d\endcsname{$-$12.5}
\expandafter\def\csname pn@etpr1.r2r.block.stall.nc_rate.hi\endcsname{$-$11.6}
\expandafter\def\csname pn@etpr1.r2r.block.stall.nc_rate.lo\endcsname{$-$13.4}
\expandafter\def\csname pn@etpr1.r2r.block.stall.nc_rate.off\endcsname{13.9}
\expandafter\def\csname pn@etpr1.r2r.block.stall.nc_rate.on\endcsname{1.4}
\expandafter\def\csname pn@etpr1.r2r.block.stall.nc_rate.rel\endcsname{$-$90}
\expandafter\def\csname pn@etpr1.r2r.block.stall.ndtw.d\endcsname{+3.7}
\expandafter\def\csname pn@etpr1.r2r.block.stall.ndtw.hi\endcsname{+4.7}
\expandafter\def\csname pn@etpr1.r2r.block.stall.ndtw.lo\endcsname{+2.8}
\expandafter\def\csname pn@etpr1.r2r.block.stall.ndtw.off\endcsname{60.9}
\expandafter\def\csname pn@etpr1.r2r.block.stall.ndtw.on\endcsname{64.6}
\expandafter\def\csname pn@etpr1.r2r.block.stall.ndtw.rel\endcsname{+6}
\expandafter\def\csname pn@etpr1.r2r.block.stall.overcap\endcsname{0}
\expandafter\def\csname pn@etpr1.r2r.block.stall.owned\endcsname{23.8}
\expandafter\def\csname pn@etpr1.r2r.block.stall.resid.maneuver\endcsname{54}
\expandafter\def\csname pn@etpr1.r2r.block.stall.resid.pass\endcsname{41}
\expandafter\def\csname pn@etpr1.r2r.block.stall.spl.d\endcsname{+3.7}
\expandafter\def\csname pn@etpr1.r2r.block.stall.spl.hi\endcsname{+5.6}
\expandafter\def\csname pn@etpr1.r2r.block.stall.spl.lo\endcsname{+2.0}
\expandafter\def\csname pn@etpr1.r2r.block.stall.spl.off\endcsname{43.3}
\expandafter\def\csname pn@etpr1.r2r.block.stall.spl.on\endcsname{47.1}
\expandafter\def\csname pn@etpr1.r2r.block.stall.spl.rel\endcsname{+9}
\expandafter\def\csname pn@etpr1.r2r.block.stall.stalls\endcsname{1.3}
\expandafter\def\csname pn@etpr1.r2r.block.stall.step_count.d\endcsname{+22.8}
\expandafter\def\csname pn@etpr1.r2r.block.stall.step_count.hi\endcsname{+26.3}
\expandafter\def\csname pn@etpr1.r2r.block.stall.step_count.lo\endcsname{+19.6}
\expandafter\def\csname pn@etpr1.r2r.block.stall.step_count.off\endcsname{111.6}
\expandafter\def\csname pn@etpr1.r2r.block.stall.step_count.on\endcsname{134.4}
\expandafter\def\csname pn@etpr1.r2r.block.stall.step_count.rel\endcsname{+20}
\expandafter\def\csname pn@etpr1.r2r.block.stall.stuck_event_count.d\endcsname{$-$1.2}
\expandafter\def\csname pn@etpr1.r2r.block.stall.stuck_event_count.hi\endcsname{$-$1.1}
\expandafter\def\csname pn@etpr1.r2r.block.stall.stuck_event_count.lo\endcsname{$-$1.3}
\expandafter\def\csname pn@etpr1.r2r.block.stall.stuck_event_count.off\endcsname{1.2}
\expandafter\def\csname pn@etpr1.r2r.block.stall.stuck_event_count.on\endcsname{0.0}
\expandafter\def\csname pn@etpr1.r2r.block.stall.stuck_event_count.rel\endcsname{$-$100}
\expandafter\def\csname pn@etpr1.r2r.block.stall.stuck_flag.d\endcsname{$-$44.4}
\expandafter\def\csname pn@etpr1.r2r.block.stall.stuck_flag.hi\endcsname{$-$42.1}
\expandafter\def\csname pn@etpr1.r2r.block.stall.stuck_flag.lo\endcsname{$-$46.7}
\expandafter\def\csname pn@etpr1.r2r.block.stall.stuck_flag.off\endcsname{44.4}
\expandafter\def\csname pn@etpr1.r2r.block.stall.stuck_flag.on\endcsname{0.0}
\expandafter\def\csname pn@etpr1.r2r.block.stall.stuck_flag.rel\endcsname{$-$100}
\expandafter\def\csname pn@etpr1.r2r.block.stall.success.d\endcsname{+4.7}
\expandafter\def\csname pn@etpr1.r2r.block.stall.success.hi\endcsname{+6.7}
\expandafter\def\csname pn@etpr1.r2r.block.stall.success.lo\endcsname{+2.7}
\expandafter\def\csname pn@etpr1.r2r.block.stall.success.off\endcsname{50.5}
\expandafter\def\csname pn@etpr1.r2r.block.stall.success.on\endcsname{55.2}
\expandafter\def\csname pn@etpr1.r2r.block.stall.success.rel\endcsname{+9}
\expandafter\def\csname pn@etpr1.r2r.block.stall.yields\endcsname{0.34}
\expandafter\def\csname pn@etpr1.r2r.block.stuck\endcsname{44.4}
\expandafter\def\csname pn@etpr1.r2r.block.tir\endcsname{22.8}
\expandafter\def\csname pn@etpr1.r2r.block.uncapped-exempt.abortep\endcsname{95}
\expandafter\def\csname pn@etpr1.r2r.block.uncapped-exempt.aborts\endcsname{6.9}
\expandafter\def\csname pn@etpr1.r2r.block.uncapped-exempt.caphit.off\endcsname{38.4}
\expandafter\def\csname pn@etpr1.r2r.block.uncapped-exempt.caphit.on\endcsname{43.4}
\expandafter\def\csname pn@etpr1.r2r.block.uncapped-exempt.clean_success.d\endcsname{+16.1}
\expandafter\def\csname pn@etpr1.r2r.block.uncapped-exempt.clean_success.hi\endcsname{+18.3}
\expandafter\def\csname pn@etpr1.r2r.block.uncapped-exempt.clean_success.lo\endcsname{+14.0}
\expandafter\def\csname pn@etpr1.r2r.block.uncapped-exempt.clean_success.off\endcsname{29.1}
\expandafter\def\csname pn@etpr1.r2r.block.uncapped-exempt.clean_success.on\endcsname{45.2}
\expandafter\def\csname pn@etpr1.r2r.block.uncapped-exempt.clean_success.rel\endcsname{+55}
\expandafter\def\csname pn@etpr1.r2r.block.uncapped-exempt.collision_count.d\endcsname{$-$4.8}
\expandafter\def\csname pn@etpr1.r2r.block.uncapped-exempt.collision_count.hi\endcsname{$-$3.6}
\expandafter\def\csname pn@etpr1.r2r.block.uncapped-exempt.collision_count.lo\endcsname{$-$5.9}
\expandafter\def\csname pn@etpr1.r2r.block.uncapped-exempt.collision_count.off\endcsname{18.1}
\expandafter\def\csname pn@etpr1.r2r.block.uncapped-exempt.collision_count.on\endcsname{13.4}
\expandafter\def\csname pn@etpr1.r2r.block.uncapped-exempt.collision_count.rel\endcsname{$-$26}
\expandafter\def\csname pn@etpr1.r2r.block.uncapped-exempt.collision_event_count.d\endcsname{$-$0.60}
\expandafter\def\csname pn@etpr1.r2r.block.uncapped-exempt.collision_event_count.hi\endcsname{$-$0.39}
\expandafter\def\csname pn@etpr1.r2r.block.uncapped-exempt.collision_event_count.lo\endcsname{$-$0.79}
\expandafter\def\csname pn@etpr1.r2r.block.uncapped-exempt.collision_event_count.off\endcsname{2.60}
\expandafter\def\csname pn@etpr1.r2r.block.uncapped-exempt.collision_event_count.on\endcsname{2.00}
\expandafter\def\csname pn@etpr1.r2r.block.uncapped-exempt.collision_event_count.rel\endcsname{$-$23}
\expandafter\def\csname pn@etpr1.r2r.block.uncapped-exempt.dirty_success_flag.d\endcsname{$-$12.6}
\expandafter\def\csname pn@etpr1.r2r.block.uncapped-exempt.dirty_success_flag.hi\endcsname{$-$10.4}
\expandafter\def\csname pn@etpr1.r2r.block.uncapped-exempt.dirty_success_flag.lo\endcsname{$-$14.7}
\expandafter\def\csname pn@etpr1.r2r.block.uncapped-exempt.dirty_success_flag.off\endcsname{21.4}
\expandafter\def\csname pn@etpr1.r2r.block.uncapped-exempt.dirty_success_flag.on\endcsname{8.8}
\expandafter\def\csname pn@etpr1.r2r.block.uncapped-exempt.dirty_success_flag.rel\endcsname{$-$59}
\expandafter\def\csname pn@etpr1.r2r.block.uncapped-exempt.gained\endcsname{230}
\expandafter\def\csname pn@etpr1.r2r.block.uncapped-exempt.lost\endcsname{166}
\expandafter\def\csname pn@etpr1.r2r.block.uncapped-exempt.navigation_error_m.d\endcsname{$-$0.3}
\expandafter\def\csname pn@etpr1.r2r.block.uncapped-exempt.navigation_error_m.hi\endcsname{$-$0.2}
\expandafter\def\csname pn@etpr1.r2r.block.uncapped-exempt.navigation_error_m.lo\endcsname{$-$0.4}
\expandafter\def\csname pn@etpr1.r2r.block.uncapped-exempt.navigation_error_m.off\endcsname{4.8}
\expandafter\def\csname pn@etpr1.r2r.block.uncapped-exempt.navigation_error_m.on\endcsname{4.5}
\expandafter\def\csname pn@etpr1.r2r.block.uncapped-exempt.navigation_error_m.rel\endcsname{$-$7}
\expandafter\def\csname pn@etpr1.r2r.block.uncapped-exempt.nc_rate.d\endcsname{$-$6.6}
\expandafter\def\csname pn@etpr1.r2r.block.uncapped-exempt.nc_rate.hi\endcsname{$-$5.6}
\expandafter\def\csname pn@etpr1.r2r.block.uncapped-exempt.nc_rate.lo\endcsname{$-$7.5}
\expandafter\def\csname pn@etpr1.r2r.block.uncapped-exempt.nc_rate.off\endcsname{13.9}
\expandafter\def\csname pn@etpr1.r2r.block.uncapped-exempt.nc_rate.on\endcsname{7.4}
\expandafter\def\csname pn@etpr1.r2r.block.uncapped-exempt.nc_rate.rel\endcsname{$-$47}
\expandafter\def\csname pn@etpr1.r2r.block.uncapped-exempt.ndtw.d\endcsname{+0.4}
\expandafter\def\csname pn@etpr1.r2r.block.uncapped-exempt.ndtw.hi\endcsname{+1.4}
\expandafter\def\csname pn@etpr1.r2r.block.uncapped-exempt.ndtw.lo\endcsname{$-$0.5}
\expandafter\def\csname pn@etpr1.r2r.block.uncapped-exempt.ndtw.off\endcsname{60.9}
\expandafter\def\csname pn@etpr1.r2r.block.uncapped-exempt.ndtw.on\endcsname{61.4}
\expandafter\def\csname pn@etpr1.r2r.block.uncapped-exempt.ndtw.rel\endcsname{+1}
\expandafter\def\csname pn@etpr1.r2r.block.uncapped-exempt.overcap\endcsname{0}
\expandafter\def\csname pn@etpr1.r2r.block.uncapped-exempt.owned\endcsname{20.0}
\expandafter\def\csname pn@etpr1.r2r.block.uncapped-exempt.resid.maneuver\endcsname{3}
\expandafter\def\csname pn@etpr1.r2r.block.uncapped-exempt.resid.pass\endcsname{87}
\expandafter\def\csname pn@etpr1.r2r.block.uncapped-exempt.spl.d\endcsname{+1.2}
\expandafter\def\csname pn@etpr1.r2r.block.uncapped-exempt.spl.hi\endcsname{+3.0}
\expandafter\def\csname pn@etpr1.r2r.block.uncapped-exempt.spl.lo\endcsname{$-$0.6}
\expandafter\def\csname pn@etpr1.r2r.block.uncapped-exempt.spl.off\endcsname{43.3}
\expandafter\def\csname pn@etpr1.r2r.block.uncapped-exempt.spl.on\endcsname{44.6}
\expandafter\def\csname pn@etpr1.r2r.block.uncapped-exempt.spl.rel\endcsname{+3}
\expandafter\def\csname pn@etpr1.r2r.block.uncapped-exempt.stalls\endcsname{0.0}
\expandafter\def\csname pn@etpr1.r2r.block.uncapped-exempt.step_count.d\endcsname{+65.7}
\expandafter\def\csname pn@etpr1.r2r.block.uncapped-exempt.step_count.hi\endcsname{+71.1}
\expandafter\def\csname pn@etpr1.r2r.block.uncapped-exempt.step_count.lo\endcsname{+60.4}
\expandafter\def\csname pn@etpr1.r2r.block.uncapped-exempt.step_count.off\endcsname{111.6}
\expandafter\def\csname pn@etpr1.r2r.block.uncapped-exempt.step_count.on\endcsname{177.2}
\expandafter\def\csname pn@etpr1.r2r.block.uncapped-exempt.step_count.rel\endcsname{+59}
\expandafter\def\csname pn@etpr1.r2r.block.uncapped-exempt.stuck_event_count.d\endcsname{$-$0.4}
\expandafter\def\csname pn@etpr1.r2r.block.uncapped-exempt.stuck_event_count.hi\endcsname{$-$0.3}
\expandafter\def\csname pn@etpr1.r2r.block.uncapped-exempt.stuck_event_count.lo\endcsname{$-$0.5}
\expandafter\def\csname pn@etpr1.r2r.block.uncapped-exempt.stuck_event_count.off\endcsname{1.2}
\expandafter\def\csname pn@etpr1.r2r.block.uncapped-exempt.stuck_event_count.on\endcsname{0.8}
\expandafter\def\csname pn@etpr1.r2r.block.uncapped-exempt.stuck_event_count.rel\endcsname{$-$35}
\expandafter\def\csname pn@etpr1.r2r.block.uncapped-exempt.stuck_flag.d\endcsname{$-$16.6}
\expandafter\def\csname pn@etpr1.r2r.block.uncapped-exempt.stuck_flag.hi\endcsname{$-$14.2}
\expandafter\def\csname pn@etpr1.r2r.block.uncapped-exempt.stuck_flag.lo\endcsname{$-$19.0}
\expandafter\def\csname pn@etpr1.r2r.block.uncapped-exempt.stuck_flag.off\endcsname{44.4}
\expandafter\def\csname pn@etpr1.r2r.block.uncapped-exempt.stuck_flag.on\endcsname{27.8}
\expandafter\def\csname pn@etpr1.r2r.block.uncapped-exempt.stuck_flag.rel\endcsname{$-$37}
\expandafter\def\csname pn@etpr1.r2r.block.uncapped-exempt.success.d\endcsname{+3.5}
\expandafter\def\csname pn@etpr1.r2r.block.uncapped-exempt.success.hi\endcsname{+5.5}
\expandafter\def\csname pn@etpr1.r2r.block.uncapped-exempt.success.lo\endcsname{+1.5}
\expandafter\def\csname pn@etpr1.r2r.block.uncapped-exempt.success.off\endcsname{50.5}
\expandafter\def\csname pn@etpr1.r2r.block.uncapped-exempt.success.on\endcsname{54.0}
\expandafter\def\csname pn@etpr1.r2r.block.uncapped-exempt.success.rel\endcsname{+7}
\expandafter\def\csname pn@etpr1.r2r.block.uncapped-exempt.yields\endcsname{1.36}
\expandafter\def\csname pn@etpr1.r2r.block.uncapped.abortep\endcsname{95}
\expandafter\def\csname pn@etpr1.r2r.block.uncapped.aborts\endcsname{5.2}
\expandafter\def\csname pn@etpr1.r2r.block.uncapped.caphit.off\endcsname{38.4}
\expandafter\def\csname pn@etpr1.r2r.block.uncapped.caphit.on\endcsname{43.4}
\expandafter\def\csname pn@etpr1.r2r.block.uncapped.clean_success.d\endcsname{+15.0}
\expandafter\def\csname pn@etpr1.r2r.block.uncapped.clean_success.hi\endcsname{+17.3}
\expandafter\def\csname pn@etpr1.r2r.block.uncapped.clean_success.lo\endcsname{+12.7}
\expandafter\def\csname pn@etpr1.r2r.block.uncapped.clean_success.off\endcsname{29.1}
\expandafter\def\csname pn@etpr1.r2r.block.uncapped.clean_success.on\endcsname{44.2}
\expandafter\def\csname pn@etpr1.r2r.block.uncapped.clean_success.rel\endcsname{+51}
\expandafter\def\csname pn@etpr1.r2r.block.uncapped.collision_count.d\endcsname{$-$8.7}
\expandafter\def\csname pn@etpr1.r2r.block.uncapped.collision_count.hi\endcsname{$-$7.6}
\expandafter\def\csname pn@etpr1.r2r.block.uncapped.collision_count.lo\endcsname{$-$9.9}
\expandafter\def\csname pn@etpr1.r2r.block.uncapped.collision_count.off\endcsname{18.1}
\expandafter\def\csname pn@etpr1.r2r.block.uncapped.collision_count.on\endcsname{9.4}
\expandafter\def\csname pn@etpr1.r2r.block.uncapped.collision_count.rel\endcsname{$-$48}
\expandafter\def\csname pn@etpr1.r2r.block.uncapped.collision_event_count.d\endcsname{$-$1.26}
\expandafter\def\csname pn@etpr1.r2r.block.uncapped.collision_event_count.hi\endcsname{$-$1.11}
\expandafter\def\csname pn@etpr1.r2r.block.uncapped.collision_event_count.lo\endcsname{$-$1.42}
\expandafter\def\csname pn@etpr1.r2r.block.uncapped.collision_event_count.off\endcsname{2.60}
\expandafter\def\csname pn@etpr1.r2r.block.uncapped.collision_event_count.on\endcsname{1.34}
\expandafter\def\csname pn@etpr1.r2r.block.uncapped.collision_event_count.rel\endcsname{$-$48}
\expandafter\def\csname pn@etpr1.r2r.block.uncapped.dirty_success_flag.d\endcsname{$-$15.7}
\expandafter\def\csname pn@etpr1.r2r.block.uncapped.dirty_success_flag.hi\endcsname{$-$13.6}
\expandafter\def\csname pn@etpr1.r2r.block.uncapped.dirty_success_flag.lo\endcsname{$-$17.7}
\expandafter\def\csname pn@etpr1.r2r.block.uncapped.dirty_success_flag.off\endcsname{21.4}
\expandafter\def\csname pn@etpr1.r2r.block.uncapped.dirty_success_flag.on\endcsname{5.7}
\expandafter\def\csname pn@etpr1.r2r.block.uncapped.dirty_success_flag.rel\endcsname{$-$73}
\expandafter\def\csname pn@etpr1.r2r.block.uncapped.gained\endcsname{186}
\expandafter\def\csname pn@etpr1.r2r.block.uncapped.lost\endcsname{198}
\expandafter\def\csname pn@etpr1.r2r.block.uncapped.navigation_error_m.d\endcsname{$-$0.1}
\expandafter\def\csname pn@etpr1.r2r.block.uncapped.navigation_error_m.hi\endcsname{+0.1}
\expandafter\def\csname pn@etpr1.r2r.block.uncapped.navigation_error_m.lo\endcsname{$-$0.2}
\expandafter\def\csname pn@etpr1.r2r.block.uncapped.navigation_error_m.off\endcsname{4.8}
\expandafter\def\csname pn@etpr1.r2r.block.uncapped.navigation_error_m.on\endcsname{4.7}
\expandafter\def\csname pn@etpr1.r2r.block.uncapped.navigation_error_m.rel\endcsname{$-$1}
\expandafter\def\csname pn@etpr1.r2r.block.uncapped.nc_rate.d\endcsname{$-$6.7}
\expandafter\def\csname pn@etpr1.r2r.block.uncapped.nc_rate.hi\endcsname{$-$5.8}
\expandafter\def\csname pn@etpr1.r2r.block.uncapped.nc_rate.lo\endcsname{$-$7.6}
\expandafter\def\csname pn@etpr1.r2r.block.uncapped.nc_rate.off\endcsname{13.9}
\expandafter\def\csname pn@etpr1.r2r.block.uncapped.nc_rate.on\endcsname{7.2}
\expandafter\def\csname pn@etpr1.r2r.block.uncapped.nc_rate.rel\endcsname{$-$48}
\expandafter\def\csname pn@etpr1.r2r.block.uncapped.ndtw.d\endcsname{+1.6}
\expandafter\def\csname pn@etpr1.r2r.block.uncapped.ndtw.hi\endcsname{+2.5}
\expandafter\def\csname pn@etpr1.r2r.block.uncapped.ndtw.lo\endcsname{+0.7}
\expandafter\def\csname pn@etpr1.r2r.block.uncapped.ndtw.off\endcsname{60.9}
\expandafter\def\csname pn@etpr1.r2r.block.uncapped.ndtw.on\endcsname{62.5}
\expandafter\def\csname pn@etpr1.r2r.block.uncapped.ndtw.rel\endcsname{+3}
\expandafter\def\csname pn@etpr1.r2r.block.uncapped.overcap\endcsname{0}
\expandafter\def\csname pn@etpr1.r2r.block.uncapped.owned\endcsname{20.0}
\expandafter\def\csname pn@etpr1.r2r.block.uncapped.resid.maneuver\endcsname{3}
\expandafter\def\csname pn@etpr1.r2r.block.uncapped.resid.pass\endcsname{91}
\expandafter\def\csname pn@etpr1.r2r.block.uncapped.spl.d\endcsname{$-$0.8}
\expandafter\def\csname pn@etpr1.r2r.block.uncapped.spl.hi\endcsname{+0.9}
\expandafter\def\csname pn@etpr1.r2r.block.uncapped.spl.lo\endcsname{$-$2.5}
\expandafter\def\csname pn@etpr1.r2r.block.uncapped.spl.off\endcsname{43.3}
\expandafter\def\csname pn@etpr1.r2r.block.uncapped.spl.on\endcsname{42.6}
\expandafter\def\csname pn@etpr1.r2r.block.uncapped.spl.rel\endcsname{$-$2}
\expandafter\def\csname pn@etpr1.r2r.block.uncapped.stalls\endcsname{0.0}
\expandafter\def\csname pn@etpr1.r2r.block.uncapped.step_count.d\endcsname{+23.5}
\expandafter\def\csname pn@etpr1.r2r.block.uncapped.step_count.hi\endcsname{+26.6}
\expandafter\def\csname pn@etpr1.r2r.block.uncapped.step_count.lo\endcsname{+20.3}
\expandafter\def\csname pn@etpr1.r2r.block.uncapped.step_count.off\endcsname{111.6}
\expandafter\def\csname pn@etpr1.r2r.block.uncapped.step_count.on\endcsname{135.0}
\expandafter\def\csname pn@etpr1.r2r.block.uncapped.step_count.rel\endcsname{+21}
\expandafter\def\csname pn@etpr1.r2r.block.uncapped.stuck_event_count.d\endcsname{$-$0.6}
\expandafter\def\csname pn@etpr1.r2r.block.uncapped.stuck_event_count.hi\endcsname{$-$0.5}
\expandafter\def\csname pn@etpr1.r2r.block.uncapped.stuck_event_count.lo\endcsname{$-$0.7}
\expandafter\def\csname pn@etpr1.r2r.block.uncapped.stuck_event_count.off\endcsname{1.2}
\expandafter\def\csname pn@etpr1.r2r.block.uncapped.stuck_event_count.on\endcsname{0.5}
\expandafter\def\csname pn@etpr1.r2r.block.uncapped.stuck_event_count.rel\endcsname{$-$54}
\expandafter\def\csname pn@etpr1.r2r.block.uncapped.stuck_flag.d\endcsname{$-$20.6}
\expandafter\def\csname pn@etpr1.r2r.block.uncapped.stuck_flag.hi\endcsname{$-$18.2}
\expandafter\def\csname pn@etpr1.r2r.block.uncapped.stuck_flag.lo\endcsname{$-$23.0}
\expandafter\def\csname pn@etpr1.r2r.block.uncapped.stuck_flag.off\endcsname{44.4}
\expandafter\def\csname pn@etpr1.r2r.block.uncapped.stuck_flag.on\endcsname{23.8}
\expandafter\def\csname pn@etpr1.r2r.block.uncapped.stuck_flag.rel\endcsname{$-$46}
\expandafter\def\csname pn@etpr1.r2r.block.uncapped.success.d\endcsname{$-$0.7}
\expandafter\def\csname pn@etpr1.r2r.block.uncapped.success.hi\endcsname{+1.4}
\expandafter\def\csname pn@etpr1.r2r.block.uncapped.success.lo\endcsname{$-$2.8}
\expandafter\def\csname pn@etpr1.r2r.block.uncapped.success.off\endcsname{50.5}
\expandafter\def\csname pn@etpr1.r2r.block.uncapped.success.on\endcsname{49.9}
\expandafter\def\csname pn@etpr1.r2r.block.uncapped.success.rel\endcsname{$-$1}
\expandafter\def\csname pn@etpr1.r2r.block.uncapped.yields\endcsname{0.65}
\expandafter\def\csname pn@etpr1.r2r.block_tryout.collision_count\endcsname{6.9}
\expandafter\def\csname pn@etpr1.r2r.block_tryout.collision_event_count\endcsname{6.7}
\expandafter\def\csname pn@etpr1.r2r.block_tryout.combovsguard.clean_success.d\endcsname{+1.3}
\expandafter\def\csname pn@etpr1.r2r.block_tryout.combovsguard.clean_success.hi\endcsname{+2.5}
\expandafter\def\csname pn@etpr1.r2r.block_tryout.combovsguard.clean_success.lo\endcsname{+0.0}
\expandafter\def\csname pn@etpr1.r2r.block_tryout.combovsguard.clean_success.off\endcsname{39.2}
\expandafter\def\csname pn@etpr1.r2r.block_tryout.combovsguard.clean_success.on\endcsname{40.5}
\expandafter\def\csname pn@etpr1.r2r.block_tryout.combovsguard.clean_success.rel\endcsname{+3}
\expandafter\def\csname pn@etpr1.r2r.block_tryout.combovsguard.collision_count.d\endcsname{$-$0.2}
\expandafter\def\csname pn@etpr1.r2r.block_tryout.combovsguard.collision_count.hi\endcsname{+0.0}
\expandafter\def\csname pn@etpr1.r2r.block_tryout.combovsguard.collision_count.lo\endcsname{$-$0.4}
\expandafter\def\csname pn@etpr1.r2r.block_tryout.combovsguard.collision_count.off\endcsname{3.2}
\expandafter\def\csname pn@etpr1.r2r.block_tryout.combovsguard.collision_count.on\endcsname{3.0}
\expandafter\def\csname pn@etpr1.r2r.block_tryout.combovsguard.collision_count.rel\endcsname{$-$6}
\expandafter\def\csname pn@etpr1.r2r.block_tryout.combovsguard.collision_event_count.d\endcsname{+0.32}
\expandafter\def\csname pn@etpr1.r2r.block_tryout.combovsguard.collision_event_count.hi\endcsname{+0.53}
\expandafter\def\csname pn@etpr1.r2r.block_tryout.combovsguard.collision_event_count.lo\endcsname{+0.10}
\expandafter\def\csname pn@etpr1.r2r.block_tryout.combovsguard.collision_event_count.off\endcsname{2.65}
\expandafter\def\csname pn@etpr1.r2r.block_tryout.combovsguard.collision_event_count.on\endcsname{2.97}
\expandafter\def\csname pn@etpr1.r2r.block_tryout.combovsguard.collision_event_count.rel\endcsname{+12}
\expandafter\def\csname pn@etpr1.r2r.block_tryout.combovsguard.dirty_success_flag.d\endcsname{+3.9}
\expandafter\def\csname pn@etpr1.r2r.block_tryout.combovsguard.dirty_success_flag.hi\endcsname{+5.5}
\expandafter\def\csname pn@etpr1.r2r.block_tryout.combovsguard.dirty_success_flag.lo\endcsname{+2.2}
\expandafter\def\csname pn@etpr1.r2r.block_tryout.combovsguard.dirty_success_flag.off\endcsname{18.2}
\expandafter\def\csname pn@etpr1.r2r.block_tryout.combovsguard.dirty_success_flag.on\endcsname{22.1}
\expandafter\def\csname pn@etpr1.r2r.block_tryout.combovsguard.dirty_success_flag.rel\endcsname{+21}
\expandafter\def\csname pn@etpr1.r2r.block_tryout.combovsguard.gained\endcsname{149}
\expandafter\def\csname pn@etpr1.r2r.block_tryout.combovsguard.lost\endcsname{55}
\expandafter\def\csname pn@etpr1.r2r.block_tryout.combovsguard.navigation_error_m.d\endcsname{$-$0.3}
\expandafter\def\csname pn@etpr1.r2r.block_tryout.combovsguard.navigation_error_m.hi\endcsname{$-$0.2}
\expandafter\def\csname pn@etpr1.r2r.block_tryout.combovsguard.navigation_error_m.lo\endcsname{$-$0.4}
\expandafter\def\csname pn@etpr1.r2r.block_tryout.combovsguard.navigation_error_m.off\endcsname{4.4}
\expandafter\def\csname pn@etpr1.r2r.block_tryout.combovsguard.navigation_error_m.on\endcsname{4.1}
\expandafter\def\csname pn@etpr1.r2r.block_tryout.combovsguard.navigation_error_m.rel\endcsname{$-$7}
\expandafter\def\csname pn@etpr1.r2r.block_tryout.combovsguard.nc_rate.d\endcsname{$-$0.9}
\expandafter\def\csname pn@etpr1.r2r.block_tryout.combovsguard.nc_rate.hi\endcsname{$-$0.8}
\expandafter\def\csname pn@etpr1.r2r.block_tryout.combovsguard.nc_rate.lo\endcsname{$-$1.0}
\expandafter\def\csname pn@etpr1.r2r.block_tryout.combovsguard.nc_rate.off\endcsname{2.5}
\expandafter\def\csname pn@etpr1.r2r.block_tryout.combovsguard.nc_rate.on\endcsname{1.6}
\expandafter\def\csname pn@etpr1.r2r.block_tryout.combovsguard.nc_rate.rel\endcsname{$-$36}
\expandafter\def\csname pn@etpr1.r2r.block_tryout.combovsguard.ndtw.d\endcsname{+0.7}
\expandafter\def\csname pn@etpr1.r2r.block_tryout.combovsguard.ndtw.hi\endcsname{+1.4}
\expandafter\def\csname pn@etpr1.r2r.block_tryout.combovsguard.ndtw.lo\endcsname{+0.1}
\expandafter\def\csname pn@etpr1.r2r.block_tryout.combovsguard.ndtw.off\endcsname{64.3}
\expandafter\def\csname pn@etpr1.r2r.block_tryout.combovsguard.ndtw.on\endcsname{65.1}
\expandafter\def\csname pn@etpr1.r2r.block_tryout.combovsguard.ndtw.rel\endcsname{+1}
\expandafter\def\csname pn@etpr1.r2r.block_tryout.combovsguard.spl.d\endcsname{+3.2}
\expandafter\def\csname pn@etpr1.r2r.block_tryout.combovsguard.spl.hi\endcsname{+4.5}
\expandafter\def\csname pn@etpr1.r2r.block_tryout.combovsguard.spl.lo\endcsname{+2.1}
\expandafter\def\csname pn@etpr1.r2r.block_tryout.combovsguard.spl.off\endcsname{49.7}
\expandafter\def\csname pn@etpr1.r2r.block_tryout.combovsguard.spl.on\endcsname{52.9}
\expandafter\def\csname pn@etpr1.r2r.block_tryout.combovsguard.spl.rel\endcsname{+7}
\expandafter\def\csname pn@etpr1.r2r.block_tryout.combovsguard.step_count.d\endcsname{+22.3}
\expandafter\def\csname pn@etpr1.r2r.block_tryout.combovsguard.step_count.hi\endcsname{+25.7}
\expandafter\def\csname pn@etpr1.r2r.block_tryout.combovsguard.step_count.lo\endcsname{+19.0}
\expandafter\def\csname pn@etpr1.r2r.block_tryout.combovsguard.step_count.off\endcsname{102.2}
\expandafter\def\csname pn@etpr1.r2r.block_tryout.combovsguard.step_count.on\endcsname{124.5}
\expandafter\def\csname pn@etpr1.r2r.block_tryout.combovsguard.step_count.rel\endcsname{+22}
\expandafter\def\csname pn@etpr1.r2r.block_tryout.combovsguard.stuck_event_count.d\endcsname{+0.0}
\expandafter\def\csname pn@etpr1.r2r.block_tryout.combovsguard.stuck_event_count.hi\endcsname{+0.0}
\expandafter\def\csname pn@etpr1.r2r.block_tryout.combovsguard.stuck_event_count.lo\endcsname{+0.0}
\expandafter\def\csname pn@etpr1.r2r.block_tryout.combovsguard.stuck_event_count.off\endcsname{0.0}
\expandafter\def\csname pn@etpr1.r2r.block_tryout.combovsguard.stuck_event_count.on\endcsname{0.0}
\expandafter\def\csname pn@etpr1.r2r.block_tryout.combovsguard.stuck_flag.d\endcsname{+0.0}
\expandafter\def\csname pn@etpr1.r2r.block_tryout.combovsguard.stuck_flag.hi\endcsname{+0.0}
\expandafter\def\csname pn@etpr1.r2r.block_tryout.combovsguard.stuck_flag.lo\endcsname{+0.0}
\expandafter\def\csname pn@etpr1.r2r.block_tryout.combovsguard.stuck_flag.off\endcsname{0.0}
\expandafter\def\csname pn@etpr1.r2r.block_tryout.combovsguard.stuck_flag.on\endcsname{0.0}
\expandafter\def\csname pn@etpr1.r2r.block_tryout.combovsguard.success.d\endcsname{+5.1}
\expandafter\def\csname pn@etpr1.r2r.block_tryout.combovsguard.success.hi\endcsname{+6.6}
\expandafter\def\csname pn@etpr1.r2r.block_tryout.combovsguard.success.lo\endcsname{+3.6}
\expandafter\def\csname pn@etpr1.r2r.block_tryout.combovsguard.success.off\endcsname{57.4}
\expandafter\def\csname pn@etpr1.r2r.block_tryout.combovsguard.success.on\endcsname{62.5}
\expandafter\def\csname pn@etpr1.r2r.block_tryout.combovsguard.success.rel\endcsname{+9}
\expandafter\def\csname pn@etpr1.r2r.block_tryout.navigation_error_m\endcsname{4.1}
\expandafter\def\csname pn@etpr1.r2r.block_tryout.ndtw\endcsname{64.3}
\expandafter\def\csname pn@etpr1.r2r.block_tryout.spl\endcsname{53.3}
\expandafter\def\csname pn@etpr1.r2r.block_tryout.stall-mem1.abortep\endcsname{46}
\expandafter\def\csname pn@etpr1.r2r.block_tryout.stall-mem1.aborts\endcsname{3.3}
\expandafter\def\csname pn@etpr1.r2r.block_tryout.stall-mem1.caphit.off\endcsname{15.4}
\expandafter\def\csname pn@etpr1.r2r.block_tryout.stall-mem1.caphit.on\endcsname{16.4}
\expandafter\def\csname pn@etpr1.r2r.block_tryout.stall-mem1.clean_success.d\endcsname{+11.3}
\expandafter\def\csname pn@etpr1.r2r.block_tryout.stall-mem1.clean_success.hi\endcsname{+12.8}
\expandafter\def\csname pn@etpr1.r2r.block_tryout.stall-mem1.clean_success.lo\endcsname{+9.8}
\expandafter\def\csname pn@etpr1.r2r.block_tryout.stall-mem1.clean_success.off\endcsname{29.1}
\expandafter\def\csname pn@etpr1.r2r.block_tryout.stall-mem1.clean_success.on\endcsname{40.5}
\expandafter\def\csname pn@etpr1.r2r.block_tryout.stall-mem1.clean_success.rel\endcsname{+39}
\expandafter\def\csname pn@etpr1.r2r.block_tryout.stall-mem1.collision_count.d\endcsname{$-$3.9}
\expandafter\def\csname pn@etpr1.r2r.block_tryout.stall-mem1.collision_count.hi\endcsname{$-$3.5}
\expandafter\def\csname pn@etpr1.r2r.block_tryout.stall-mem1.collision_count.lo\endcsname{$-$4.3}
\expandafter\def\csname pn@etpr1.r2r.block_tryout.stall-mem1.collision_count.off\endcsname{6.9}
\expandafter\def\csname pn@etpr1.r2r.block_tryout.stall-mem1.collision_count.on\endcsname{3.0}
\expandafter\def\csname pn@etpr1.r2r.block_tryout.stall-mem1.collision_count.rel\endcsname{$-$56}
\expandafter\def\csname pn@etpr1.r2r.block_tryout.stall-mem1.collision_event_count.d\endcsname{$-$3.70}
\expandafter\def\csname pn@etpr1.r2r.block_tryout.stall-mem1.collision_event_count.hi\endcsname{$-$3.31}
\expandafter\def\csname pn@etpr1.r2r.block_tryout.stall-mem1.collision_event_count.lo\endcsname{$-$4.09}
\expandafter\def\csname pn@etpr1.r2r.block_tryout.stall-mem1.collision_event_count.off\endcsname{6.67}
\expandafter\def\csname pn@etpr1.r2r.block_tryout.stall-mem1.collision_event_count.on\endcsname{2.97}
\expandafter\def\csname pn@etpr1.r2r.block_tryout.stall-mem1.collision_event_count.rel\endcsname{$-$55}
\expandafter\def\csname pn@etpr1.r2r.block_tryout.stall-mem1.dirty_success_flag.d\endcsname{$-$12.2}
\expandafter\def\csname pn@etpr1.r2r.block_tryout.stall-mem1.dirty_success_flag.hi\endcsname{$-$10.5}
\expandafter\def\csname pn@etpr1.r2r.block_tryout.stall-mem1.dirty_success_flag.lo\endcsname{$-$13.8}
\expandafter\def\csname pn@etpr1.r2r.block_tryout.stall-mem1.dirty_success_flag.off\endcsname{34.3}
\expandafter\def\csname pn@etpr1.r2r.block_tryout.stall-mem1.dirty_success_flag.on\endcsname{22.1}
\expandafter\def\csname pn@etpr1.r2r.block_tryout.stall-mem1.dirty_success_flag.rel\endcsname{$-$36}
\expandafter\def\csname pn@etpr1.r2r.block_tryout.stall-mem1.gained\endcsname{38}
\expandafter\def\csname pn@etpr1.r2r.block_tryout.stall-mem1.lost\endcsname{54}
\expandafter\def\csname pn@etpr1.r2r.block_tryout.stall-mem1.navigation_error_m.d\endcsname{+0.0}
\expandafter\def\csname pn@etpr1.r2r.block_tryout.stall-mem1.navigation_error_m.hi\endcsname{+0.1}
\expandafter\def\csname pn@etpr1.r2r.block_tryout.stall-mem1.navigation_error_m.lo\endcsname{$-$0.1}
\expandafter\def\csname pn@etpr1.r2r.block_tryout.stall-mem1.navigation_error_m.off\endcsname{4.1}
\expandafter\def\csname pn@etpr1.r2r.block_tryout.stall-mem1.navigation_error_m.on\endcsname{4.1}
\expandafter\def\csname pn@etpr1.r2r.block_tryout.stall-mem1.navigation_error_m.rel\endcsname{+0}
\expandafter\def\csname pn@etpr1.r2r.block_tryout.stall-mem1.nc_rate.d\endcsname{$-$2.1}
\expandafter\def\csname pn@etpr1.r2r.block_tryout.stall-mem1.nc_rate.hi\endcsname{$-$1.9}
\expandafter\def\csname pn@etpr1.r2r.block_tryout.stall-mem1.nc_rate.lo\endcsname{$-$2.2}
\expandafter\def\csname pn@etpr1.r2r.block_tryout.stall-mem1.nc_rate.off\endcsname{3.6}
\expandafter\def\csname pn@etpr1.r2r.block_tryout.stall-mem1.nc_rate.on\endcsname{1.6}
\expandafter\def\csname pn@etpr1.r2r.block_tryout.stall-mem1.nc_rate.rel\endcsname{$-$57}
\expandafter\def\csname pn@etpr1.r2r.block_tryout.stall-mem1.ndtw.d\endcsname{+0.8}
\expandafter\def\csname pn@etpr1.r2r.block_tryout.stall-mem1.ndtw.hi\endcsname{+1.3}
\expandafter\def\csname pn@etpr1.r2r.block_tryout.stall-mem1.ndtw.lo\endcsname{+0.3}
\expandafter\def\csname pn@etpr1.r2r.block_tryout.stall-mem1.ndtw.off\endcsname{64.3}
\expandafter\def\csname pn@etpr1.r2r.block_tryout.stall-mem1.ndtw.on\endcsname{65.1}
\expandafter\def\csname pn@etpr1.r2r.block_tryout.stall-mem1.ndtw.rel\endcsname{+1}
\expandafter\def\csname pn@etpr1.r2r.block_tryout.stall-mem1.overcap\endcsname{0}
\expandafter\def\csname pn@etpr1.r2r.block_tryout.stall-mem1.owned\endcsname{3.3}
\expandafter\def\csname pn@etpr1.r2r.block_tryout.stall-mem1.resid.maneuver\endcsname{0}
\expandafter\def\csname pn@etpr1.r2r.block_tryout.stall-mem1.resid.pass\endcsname{99}
\expandafter\def\csname pn@etpr1.r2r.block_tryout.stall-mem1.spl.d\endcsname{$-$0.3}
\expandafter\def\csname pn@etpr1.r2r.block_tryout.stall-mem1.spl.hi\endcsname{+0.5}
\expandafter\def\csname pn@etpr1.r2r.block_tryout.stall-mem1.spl.lo\endcsname{$-$1.2}
\expandafter\def\csname pn@etpr1.r2r.block_tryout.stall-mem1.spl.off\endcsname{53.3}
\expandafter\def\csname pn@etpr1.r2r.block_tryout.stall-mem1.spl.on\endcsname{52.9}
\expandafter\def\csname pn@etpr1.r2r.block_tryout.stall-mem1.spl.rel\endcsname{$-$1}
\expandafter\def\csname pn@etpr1.r2r.block_tryout.stall-mem1.stalls\endcsname{0.8}
\expandafter\def\csname pn@etpr1.r2r.block_tryout.stall-mem1.step_count.d\endcsname{$-$3.6}
\expandafter\def\csname pn@etpr1.r2r.block_tryout.stall-mem1.step_count.hi\endcsname{$-$1.0}
\expandafter\def\csname pn@etpr1.r2r.block_tryout.stall-mem1.step_count.lo\endcsname{$-$6.3}
\expandafter\def\csname pn@etpr1.r2r.block_tryout.stall-mem1.step_count.off\endcsname{128.1}
\expandafter\def\csname pn@etpr1.r2r.block_tryout.stall-mem1.step_count.on\endcsname{124.5}
\expandafter\def\csname pn@etpr1.r2r.block_tryout.stall-mem1.step_count.rel\endcsname{$-$3}
\expandafter\def\csname pn@etpr1.r2r.block_tryout.stall-mem1.stuck_event_count.d\endcsname{+0.0}
\expandafter\def\csname pn@etpr1.r2r.block_tryout.stall-mem1.stuck_event_count.hi\endcsname{+0.0}
\expandafter\def\csname pn@etpr1.r2r.block_tryout.stall-mem1.stuck_event_count.lo\endcsname{+0.0}
\expandafter\def\csname pn@etpr1.r2r.block_tryout.stall-mem1.stuck_event_count.off\endcsname{0.0}
\expandafter\def\csname pn@etpr1.r2r.block_tryout.stall-mem1.stuck_event_count.on\endcsname{0.0}
\expandafter\def\csname pn@etpr1.r2r.block_tryout.stall-mem1.stuck_flag.d\endcsname{+0.0}
\expandafter\def\csname pn@etpr1.r2r.block_tryout.stall-mem1.stuck_flag.hi\endcsname{+0.0}
\expandafter\def\csname pn@etpr1.r2r.block_tryout.stall-mem1.stuck_flag.lo\endcsname{+0.0}
\expandafter\def\csname pn@etpr1.r2r.block_tryout.stall-mem1.stuck_flag.off\endcsname{0.0}
\expandafter\def\csname pn@etpr1.r2r.block_tryout.stall-mem1.stuck_flag.on\endcsname{0.0}
\expandafter\def\csname pn@etpr1.r2r.block_tryout.stall-mem1.success.d\endcsname{$-$0.9}
\expandafter\def\csname pn@etpr1.r2r.block_tryout.stall-mem1.success.hi\endcsname{+0.2}
\expandafter\def\csname pn@etpr1.r2r.block_tryout.stall-mem1.success.lo\endcsname{$-$1.9}
\expandafter\def\csname pn@etpr1.r2r.block_tryout.stall-mem1.success.off\endcsname{63.4}
\expandafter\def\csname pn@etpr1.r2r.block_tryout.stall-mem1.success.on\endcsname{62.5}
\expandafter\def\csname pn@etpr1.r2r.block_tryout.stall-mem1.success.rel\endcsname{$-$1}
\expandafter\def\csname pn@etpr1.r2r.block_tryout.stall-mem1.yields\endcsname{0.05}
\expandafter\def\csname pn@etpr1.r2r.block_tryout.step_count\endcsname{128.1}
\expandafter\def\csname pn@etpr1.r2r.block_tryout.stuck_flag\endcsname{0.0}
\expandafter\def\csname pn@etpr1.r2r.block_tryout.success\endcsname{63.4}
\expandafter\def\csname pn@etpr1.r2r.block_tryout.vsguard.clean_success.d\endcsname{+10.1}
\expandafter\def\csname pn@etpr1.r2r.block_tryout.vsguard.clean_success.hi\endcsname{+11.4}
\expandafter\def\csname pn@etpr1.r2r.block_tryout.vsguard.clean_success.lo\endcsname{+8.7}
\expandafter\def\csname pn@etpr1.r2r.block_tryout.vsguard.clean_success.off\endcsname{29.1}
\expandafter\def\csname pn@etpr1.r2r.block_tryout.vsguard.clean_success.on\endcsname{39.2}
\expandafter\def\csname pn@etpr1.r2r.block_tryout.vsguard.clean_success.rel\endcsname{+35}
\expandafter\def\csname pn@etpr1.r2r.block_tryout.vsguard.collision_count.d\endcsname{$-$3.7}
\expandafter\def\csname pn@etpr1.r2r.block_tryout.vsguard.collision_count.hi\endcsname{$-$3.2}
\expandafter\def\csname pn@etpr1.r2r.block_tryout.vsguard.collision_count.lo\endcsname{$-$4.2}
\expandafter\def\csname pn@etpr1.r2r.block_tryout.vsguard.collision_count.off\endcsname{6.9}
\expandafter\def\csname pn@etpr1.r2r.block_tryout.vsguard.collision_count.on\endcsname{3.2}
\expandafter\def\csname pn@etpr1.r2r.block_tryout.vsguard.collision_count.rel\endcsname{$-$54}
\expandafter\def\csname pn@etpr1.r2r.block_tryout.vsguard.collision_event_count.d\endcsname{$-$4.02}
\expandafter\def\csname pn@etpr1.r2r.block_tryout.vsguard.collision_event_count.hi\endcsname{$-$3.56}
\expandafter\def\csname pn@etpr1.r2r.block_tryout.vsguard.collision_event_count.lo\endcsname{$-$4.49}
\expandafter\def\csname pn@etpr1.r2r.block_tryout.vsguard.collision_event_count.off\endcsname{6.67}
\expandafter\def\csname pn@etpr1.r2r.block_tryout.vsguard.collision_event_count.on\endcsname{2.65}
\expandafter\def\csname pn@etpr1.r2r.block_tryout.vsguard.collision_event_count.rel\endcsname{$-$60}
\expandafter\def\csname pn@etpr1.r2r.block_tryout.vsguard.dirty_success_flag.d\endcsname{$-$16.0}
\expandafter\def\csname pn@etpr1.r2r.block_tryout.vsguard.dirty_success_flag.hi\endcsname{$-$14.2}
\expandafter\def\csname pn@etpr1.r2r.block_tryout.vsguard.dirty_success_flag.lo\endcsname{$-$17.8}
\expandafter\def\csname pn@etpr1.r2r.block_tryout.vsguard.dirty_success_flag.off\endcsname{34.3}
\expandafter\def\csname pn@etpr1.r2r.block_tryout.vsguard.dirty_success_flag.on\endcsname{18.2}
\expandafter\def\csname pn@etpr1.r2r.block_tryout.vsguard.dirty_success_flag.rel\endcsname{$-$47}
\expandafter\def\csname pn@etpr1.r2r.block_tryout.vsguard.gained\endcsname{53}
\expandafter\def\csname pn@etpr1.r2r.block_tryout.vsguard.lost\endcsname{163}
\expandafter\def\csname pn@etpr1.r2r.block_tryout.vsguard.navigation_error_m.d\endcsname{+0.3}
\expandafter\def\csname pn@etpr1.r2r.block_tryout.vsguard.navigation_error_m.hi\endcsname{+0.4}
\expandafter\def\csname pn@etpr1.r2r.block_tryout.vsguard.navigation_error_m.lo\endcsname{+0.2}
\expandafter\def\csname pn@etpr1.r2r.block_tryout.vsguard.navigation_error_m.off\endcsname{4.1}
\expandafter\def\csname pn@etpr1.r2r.block_tryout.vsguard.navigation_error_m.on\endcsname{4.4}
\expandafter\def\csname pn@etpr1.r2r.block_tryout.vsguard.navigation_error_m.rel\endcsname{+8}
\expandafter\def\csname pn@etpr1.r2r.block_tryout.vsguard.nc_rate.d\endcsname{$-$1.2}
\expandafter\def\csname pn@etpr1.r2r.block_tryout.vsguard.nc_rate.hi\endcsname{$-$1.0}
\expandafter\def\csname pn@etpr1.r2r.block_tryout.vsguard.nc_rate.lo\endcsname{$-$1.3}
\expandafter\def\csname pn@etpr1.r2r.block_tryout.vsguard.nc_rate.off\endcsname{3.6}
\expandafter\def\csname pn@etpr1.r2r.block_tryout.vsguard.nc_rate.on\endcsname{2.5}
\expandafter\def\csname pn@etpr1.r2r.block_tryout.vsguard.nc_rate.rel\endcsname{$-$32}
\expandafter\def\csname pn@etpr1.r2r.block_tryout.vsguard.ndtw.d\endcsname{+0.0}
\expandafter\def\csname pn@etpr1.r2r.block_tryout.vsguard.ndtw.hi\endcsname{+0.7}
\expandafter\def\csname pn@etpr1.r2r.block_tryout.vsguard.ndtw.lo\endcsname{$-$0.7}
\expandafter\def\csname pn@etpr1.r2r.block_tryout.vsguard.ndtw.off\endcsname{64.3}
\expandafter\def\csname pn@etpr1.r2r.block_tryout.vsguard.ndtw.on\endcsname{64.3}
\expandafter\def\csname pn@etpr1.r2r.block_tryout.vsguard.ndtw.rel\endcsname{+0}
\expandafter\def\csname pn@etpr1.r2r.block_tryout.vsguard.spl.d\endcsname{$-$3.6}
\expandafter\def\csname pn@etpr1.r2r.block_tryout.vsguard.spl.hi\endcsname{$-$2.4}
\expandafter\def\csname pn@etpr1.r2r.block_tryout.vsguard.spl.lo\endcsname{$-$4.8}
\expandafter\def\csname pn@etpr1.r2r.block_tryout.vsguard.spl.off\endcsname{53.3}
\expandafter\def\csname pn@etpr1.r2r.block_tryout.vsguard.spl.on\endcsname{49.7}
\expandafter\def\csname pn@etpr1.r2r.block_tryout.vsguard.spl.rel\endcsname{$-$7}
\expandafter\def\csname pn@etpr1.r2r.block_tryout.vsguard.step_count.d\endcsname{$-$25.9}
\expandafter\def\csname pn@etpr1.r2r.block_tryout.vsguard.step_count.hi\endcsname{$-$22.4}
\expandafter\def\csname pn@etpr1.r2r.block_tryout.vsguard.step_count.lo\endcsname{$-$29.5}
\expandafter\def\csname pn@etpr1.r2r.block_tryout.vsguard.step_count.off\endcsname{128.1}
\expandafter\def\csname pn@etpr1.r2r.block_tryout.vsguard.step_count.on\endcsname{102.2}
\expandafter\def\csname pn@etpr1.r2r.block_tryout.vsguard.step_count.rel\endcsname{$-$20}
\expandafter\def\csname pn@etpr1.r2r.block_tryout.vsguard.stuck_event_count.d\endcsname{+0.0}
\expandafter\def\csname pn@etpr1.r2r.block_tryout.vsguard.stuck_event_count.hi\endcsname{+0.0}
\expandafter\def\csname pn@etpr1.r2r.block_tryout.vsguard.stuck_event_count.lo\endcsname{+0.0}
\expandafter\def\csname pn@etpr1.r2r.block_tryout.vsguard.stuck_event_count.off\endcsname{0.0}
\expandafter\def\csname pn@etpr1.r2r.block_tryout.vsguard.stuck_event_count.on\endcsname{0.0}
\expandafter\def\csname pn@etpr1.r2r.block_tryout.vsguard.stuck_flag.d\endcsname{+0.0}
\expandafter\def\csname pn@etpr1.r2r.block_tryout.vsguard.stuck_flag.hi\endcsname{+0.0}
\expandafter\def\csname pn@etpr1.r2r.block_tryout.vsguard.stuck_flag.lo\endcsname{+0.0}
\expandafter\def\csname pn@etpr1.r2r.block_tryout.vsguard.stuck_flag.off\endcsname{0.0}
\expandafter\def\csname pn@etpr1.r2r.block_tryout.vsguard.stuck_flag.on\endcsname{0.0}
\expandafter\def\csname pn@etpr1.r2r.block_tryout.vsguard.success.d\endcsname{$-$6.0}
\expandafter\def\csname pn@etpr1.r2r.block_tryout.vsguard.success.hi\endcsname{$-$4.5}
\expandafter\def\csname pn@etpr1.r2r.block_tryout.vsguard.success.lo\endcsname{$-$7.6}
\expandafter\def\csname pn@etpr1.r2r.block_tryout.vsguard.success.off\endcsname{63.4}
\expandafter\def\csname pn@etpr1.r2r.block_tryout.vsguard.success.on\endcsname{57.4}
\expandafter\def\csname pn@etpr1.r2r.block_tryout.vsguard.success.rel\endcsname{$-$9}
\expandafter\def\csname pn@etpr1.r2r.block_tryout.vsoff.clean_success.d\endcsname{+0.0}
\expandafter\def\csname pn@etpr1.r2r.block_tryout.vsoff.clean_success.hi\endcsname{+0.0}
\expandafter\def\csname pn@etpr1.r2r.block_tryout.vsoff.clean_success.lo\endcsname{+0.0}
\expandafter\def\csname pn@etpr1.r2r.block_tryout.vsoff.clean_success.off\endcsname{29.1}
\expandafter\def\csname pn@etpr1.r2r.block_tryout.vsoff.clean_success.on\endcsname{29.1}
\expandafter\def\csname pn@etpr1.r2r.block_tryout.vsoff.clean_success.rel\endcsname{+0}
\expandafter\def\csname pn@etpr1.r2r.block_tryout.vsoff.collision_count.d\endcsname{$-$11.3}
\expandafter\def\csname pn@etpr1.r2r.block_tryout.vsoff.collision_count.hi\endcsname{$-$10.3}
\expandafter\def\csname pn@etpr1.r2r.block_tryout.vsoff.collision_count.lo\endcsname{$-$12.2}
\expandafter\def\csname pn@etpr1.r2r.block_tryout.vsoff.collision_count.off\endcsname{18.1}
\expandafter\def\csname pn@etpr1.r2r.block_tryout.vsoff.collision_count.on\endcsname{6.9}
\expandafter\def\csname pn@etpr1.r2r.block_tryout.vsoff.collision_count.rel\endcsname{$-$62}
\expandafter\def\csname pn@etpr1.r2r.block_tryout.vsoff.collision_event_count.d\endcsname{+4.07}
\expandafter\def\csname pn@etpr1.r2r.block_tryout.vsoff.collision_event_count.hi\endcsname{+4.56}
\expandafter\def\csname pn@etpr1.r2r.block_tryout.vsoff.collision_event_count.lo\endcsname{+3.60}
\expandafter\def\csname pn@etpr1.r2r.block_tryout.vsoff.collision_event_count.off\endcsname{2.60}
\expandafter\def\csname pn@etpr1.r2r.block_tryout.vsoff.collision_event_count.on\endcsname{6.67}
\expandafter\def\csname pn@etpr1.r2r.block_tryout.vsoff.collision_event_count.rel\endcsname{+157}
\expandafter\def\csname pn@etpr1.r2r.block_tryout.vsoff.dirty_success_flag.d\endcsname{+12.9}
\expandafter\def\csname pn@etpr1.r2r.block_tryout.vsoff.dirty_success_flag.hi\endcsname{+14.8}
\expandafter\def\csname pn@etpr1.r2r.block_tryout.vsoff.dirty_success_flag.lo\endcsname{+11.1}
\expandafter\def\csname pn@etpr1.r2r.block_tryout.vsoff.dirty_success_flag.off\endcsname{21.4}
\expandafter\def\csname pn@etpr1.r2r.block_tryout.vsoff.dirty_success_flag.on\endcsname{34.3}
\expandafter\def\csname pn@etpr1.r2r.block_tryout.vsoff.dirty_success_flag.rel\endcsname{+60}
\expandafter\def\csname pn@etpr1.r2r.block_tryout.vsoff.gained\endcsname{276}
\expandafter\def\csname pn@etpr1.r2r.block_tryout.vsoff.lost\endcsname{39}
\expandafter\def\csname pn@etpr1.r2r.block_tryout.vsoff.navigation_error_m.d\endcsname{$-$0.8}
\expandafter\def\csname pn@etpr1.r2r.block_tryout.vsoff.navigation_error_m.hi\endcsname{$-$0.6}
\expandafter\def\csname pn@etpr1.r2r.block_tryout.vsoff.navigation_error_m.lo\endcsname{$-$0.9}
\expandafter\def\csname pn@etpr1.r2r.block_tryout.vsoff.navigation_error_m.off\endcsname{4.8}
\expandafter\def\csname pn@etpr1.r2r.block_tryout.vsoff.navigation_error_m.on\endcsname{4.1}
\expandafter\def\csname pn@etpr1.r2r.block_tryout.vsoff.navigation_error_m.rel\endcsname{$-$16}
\expandafter\def\csname pn@etpr1.r2r.block_tryout.vsoff.nc_rate.d\endcsname{$-$10.3}
\expandafter\def\csname pn@etpr1.r2r.block_tryout.vsoff.nc_rate.hi\endcsname{$-$9.5}
\expandafter\def\csname pn@etpr1.r2r.block_tryout.vsoff.nc_rate.lo\endcsname{$-$11.1}
\expandafter\def\csname pn@etpr1.r2r.block_tryout.vsoff.nc_rate.off\endcsname{13.9}
\expandafter\def\csname pn@etpr1.r2r.block_tryout.vsoff.nc_rate.on\endcsname{3.6}
\expandafter\def\csname pn@etpr1.r2r.block_tryout.vsoff.nc_rate.rel\endcsname{$-$74}
\expandafter\def\csname pn@etpr1.r2r.block_tryout.vsoff.ndtw.d\endcsname{+3.4}
\expandafter\def\csname pn@etpr1.r2r.block_tryout.vsoff.ndtw.hi\endcsname{+4.2}
\expandafter\def\csname pn@etpr1.r2r.block_tryout.vsoff.ndtw.lo\endcsname{+2.5}
\expandafter\def\csname pn@etpr1.r2r.block_tryout.vsoff.ndtw.off\endcsname{60.9}
\expandafter\def\csname pn@etpr1.r2r.block_tryout.vsoff.ndtw.on\endcsname{64.3}
\expandafter\def\csname pn@etpr1.r2r.block_tryout.vsoff.ndtw.rel\endcsname{+6}
\expandafter\def\csname pn@etpr1.r2r.block_tryout.vsoff.spl.d\endcsname{+9.9}
\expandafter\def\csname pn@etpr1.r2r.block_tryout.vsoff.spl.hi\endcsname{+11.4}
\expandafter\def\csname pn@etpr1.r2r.block_tryout.vsoff.spl.lo\endcsname{+8.5}
\expandafter\def\csname pn@etpr1.r2r.block_tryout.vsoff.spl.off\endcsname{43.3}
\expandafter\def\csname pn@etpr1.r2r.block_tryout.vsoff.spl.on\endcsname{53.3}
\expandafter\def\csname pn@etpr1.r2r.block_tryout.vsoff.spl.rel\endcsname{+23}
\expandafter\def\csname pn@etpr1.r2r.block_tryout.vsoff.step_count.d\endcsname{+16.6}
\expandafter\def\csname pn@etpr1.r2r.block_tryout.vsoff.step_count.hi\endcsname{+20.6}
\expandafter\def\csname pn@etpr1.r2r.block_tryout.vsoff.step_count.lo\endcsname{+12.8}
\expandafter\def\csname pn@etpr1.r2r.block_tryout.vsoff.step_count.off\endcsname{111.6}
\expandafter\def\csname pn@etpr1.r2r.block_tryout.vsoff.step_count.on\endcsname{128.1}
\expandafter\def\csname pn@etpr1.r2r.block_tryout.vsoff.step_count.rel\endcsname{+15}
\expandafter\def\csname pn@etpr1.r2r.block_tryout.vsoff.stuck_event_count.d\endcsname{$-$1.2}
\expandafter\def\csname pn@etpr1.r2r.block_tryout.vsoff.stuck_event_count.hi\endcsname{$-$1.1}
\expandafter\def\csname pn@etpr1.r2r.block_tryout.vsoff.stuck_event_count.lo\endcsname{$-$1.3}
\expandafter\def\csname pn@etpr1.r2r.block_tryout.vsoff.stuck_event_count.off\endcsname{1.2}
\expandafter\def\csname pn@etpr1.r2r.block_tryout.vsoff.stuck_event_count.on\endcsname{0.0}
\expandafter\def\csname pn@etpr1.r2r.block_tryout.vsoff.stuck_event_count.rel\endcsname{$-$100}
\expandafter\def\csname pn@etpr1.r2r.block_tryout.vsoff.stuck_flag.d\endcsname{$-$44.4}
\expandafter\def\csname pn@etpr1.r2r.block_tryout.vsoff.stuck_flag.hi\endcsname{$-$42.1}
\expandafter\def\csname pn@etpr1.r2r.block_tryout.vsoff.stuck_flag.lo\endcsname{$-$46.7}
\expandafter\def\csname pn@etpr1.r2r.block_tryout.vsoff.stuck_flag.off\endcsname{44.4}
\expandafter\def\csname pn@etpr1.r2r.block_tryout.vsoff.stuck_flag.on\endcsname{0.0}
\expandafter\def\csname pn@etpr1.r2r.block_tryout.vsoff.stuck_flag.rel\endcsname{$-$100}
\expandafter\def\csname pn@etpr1.r2r.block_tryout.vsoff.success.d\endcsname{+12.9}
\expandafter\def\csname pn@etpr1.r2r.block_tryout.vsoff.success.hi\endcsname{+14.7}
\expandafter\def\csname pn@etpr1.r2r.block_tryout.vsoff.success.lo\endcsname{+11.1}
\expandafter\def\csname pn@etpr1.r2r.block_tryout.vsoff.success.off\endcsname{50.5}
\expandafter\def\csname pn@etpr1.r2r.block_tryout.vsoff.success.on\endcsname{63.4}
\expandafter\def\csname pn@etpr1.r2r.block_tryout.vsoff.success.rel\endcsname{+26}
\expandafter\def\csname pn@etpr1.r2r.ddsr_pp\endcsname{$-$0.6}
\expandafter\def\csname pn@etpr1.r2r.dsr_pp\endcsname{+14.1}
\expandafter\def\csname pn@etpr1.r2r.hist.flagged\endcsname{28{,}428}
\expandafter\def\csname pn@etpr1.r2r.hist.recall\endcsname{85.3}
\expandafter\def\csname pn@etpr1.r2r.hist.tp\endcsname{28{,}428}
\expandafter\def\csname pn@etpr1.r2r.kmiss\endcsname{77.7}
\expandafter\def\csname pn@etpr1.r2r.slide.cleansuccess\endcsname{37.7}
\expandafter\def\csname pn@etpr1.r2r.slide.cpds\endcsname{5.0}
\expandafter\def\csname pn@etpr1.r2r.slide.cpdsmed\endcsname{3}
\expandafter\def\csname pn@etpr1.r2r.slide.dirtysuccess\endcsname{26.9}
\expandafter\def\csname pn@etpr1.r2r.slide.dsr\endcsname{41.7}
\expandafter\def\csname pn@etpr1.r2r.slide.dsrhi\endcsname{44.5}
\expandafter\def\csname pn@etpr1.r2r.slide.dsrlo\endcsname{38.9}
\expandafter\def\csname pn@etpr1.r2r.slide.nc\endcsname{4.0}
\expandafter\def\csname pn@etpr1.r2r.slide.ncfail\endcsname{6.0}
\expandafter\def\csname pn@etpr1.r2r.slide.ncsucc\endcsname{2.9}
\expandafter\def\csname pn@etpr1.r2r.slide.proxydsr\endcsname{6.4}
\expandafter\def\csname pn@etpr1.r2r.slide.shareinsucc\endcsname{33.9}
\expandafter\def\csname pn@etpr1.r2r.slide.sr\endcsname{64.6}
\expandafter\def\csname pn@etpr1.r2r.slide.stall-mem1.abortep\endcsname{43}
\expandafter\def\csname pn@etpr1.r2r.slide.stall-mem1.aborts\endcsname{0.9}
\expandafter\def\csname pn@etpr1.r2r.slide.stall-mem1.caphit.off\endcsname{14.5}
\expandafter\def\csname pn@etpr1.r2r.slide.stall-mem1.caphit.on\endcsname{15.9}
\expandafter\def\csname pn@etpr1.r2r.slide.stall-mem1.clean_success.d\endcsname{+12.5}
\expandafter\def\csname pn@etpr1.r2r.slide.stall-mem1.clean_success.hi\endcsname{+14.0}
\expandafter\def\csname pn@etpr1.r2r.slide.stall-mem1.clean_success.lo\endcsname{+10.9}
\expandafter\def\csname pn@etpr1.r2r.slide.stall-mem1.clean_success.off\endcsname{37.7}
\expandafter\def\csname pn@etpr1.r2r.slide.stall-mem1.clean_success.on\endcsname{50.1}
\expandafter\def\csname pn@etpr1.r2r.slide.stall-mem1.clean_success.rel\endcsname{+33}
\expandafter\def\csname pn@etpr1.r2r.slide.stall-mem1.collision_count.d\endcsname{$-$2.5}
\expandafter\def\csname pn@etpr1.r2r.slide.stall-mem1.collision_count.hi\endcsname{$-$2.2}
\expandafter\def\csname pn@etpr1.r2r.slide.stall-mem1.collision_count.lo\endcsname{$-$2.9}
\expandafter\def\csname pn@etpr1.r2r.slide.stall-mem1.collision_count.off\endcsname{3.9}
\expandafter\def\csname pn@etpr1.r2r.slide.stall-mem1.collision_count.on\endcsname{1.4}
\expandafter\def\csname pn@etpr1.r2r.slide.stall-mem1.collision_count.rel\endcsname{$-$64}
\expandafter\def\csname pn@etpr1.r2r.slide.stall-mem1.collision_event_count.d\endcsname{$-$0.23}
\expandafter\def\csname pn@etpr1.r2r.slide.stall-mem1.collision_event_count.hi\endcsname{$-$0.13}
\expandafter\def\csname pn@etpr1.r2r.slide.stall-mem1.collision_event_count.lo\endcsname{$-$0.33}
\expandafter\def\csname pn@etpr1.r2r.slide.stall-mem1.collision_event_count.off\endcsname{1.13}
\expandafter\def\csname pn@etpr1.r2r.slide.stall-mem1.collision_event_count.on\endcsname{0.90}
\expandafter\def\csname pn@etpr1.r2r.slide.stall-mem1.collision_event_count.rel\endcsname{$-$20}
\expandafter\def\csname pn@etpr1.r2r.slide.stall-mem1.dirty_success_flag.d\endcsname{$-$13.2}
\expandafter\def\csname pn@etpr1.r2r.slide.stall-mem1.dirty_success_flag.hi\endcsname{$-$11.6}
\expandafter\def\csname pn@etpr1.r2r.slide.stall-mem1.dirty_success_flag.lo\endcsname{$-$14.8}
\expandafter\def\csname pn@etpr1.r2r.slide.stall-mem1.dirty_success_flag.off\endcsname{26.9}
\expandafter\def\csname pn@etpr1.r2r.slide.stall-mem1.dirty_success_flag.on\endcsname{13.7}
\expandafter\def\csname pn@etpr1.r2r.slide.stall-mem1.dirty_success_flag.rel\endcsname{$-$49}
\expandafter\def\csname pn@etpr1.r2r.slide.stall-mem1.gained\endcsname{45}
\expandafter\def\csname pn@etpr1.r2r.slide.stall-mem1.lost\endcsname{59}
\expandafter\def\csname pn@etpr1.r2r.slide.stall-mem1.navigation_error_m.d\endcsname{+0.0}
\expandafter\def\csname pn@etpr1.r2r.slide.stall-mem1.navigation_error_m.hi\endcsname{+0.1}
\expandafter\def\csname pn@etpr1.r2r.slide.stall-mem1.navigation_error_m.lo\endcsname{$-$0.0}
\expandafter\def\csname pn@etpr1.r2r.slide.stall-mem1.navigation_error_m.off\endcsname{4.0}
\expandafter\def\csname pn@etpr1.r2r.slide.stall-mem1.navigation_error_m.on\endcsname{4.0}
\expandafter\def\csname pn@etpr1.r2r.slide.stall-mem1.navigation_error_m.rel\endcsname{+1}
\expandafter\def\csname pn@etpr1.r2r.slide.stall-mem1.nc_rate.d\endcsname{$-$2.5}
\expandafter\def\csname pn@etpr1.r2r.slide.stall-mem1.nc_rate.hi\endcsname{$-$2.3}
\expandafter\def\csname pn@etpr1.r2r.slide.stall-mem1.nc_rate.lo\endcsname{$-$2.9}
\expandafter\def\csname pn@etpr1.r2r.slide.stall-mem1.nc_rate.off\endcsname{4.0}
\expandafter\def\csname pn@etpr1.r2r.slide.stall-mem1.nc_rate.on\endcsname{1.4}
\expandafter\def\csname pn@etpr1.r2r.slide.stall-mem1.nc_rate.rel\endcsname{$-$64}
\expandafter\def\csname pn@etpr1.r2r.slide.stall-mem1.ndtw.d\endcsname{+0.4}
\expandafter\def\csname pn@etpr1.r2r.slide.stall-mem1.ndtw.hi\endcsname{+0.8}
\expandafter\def\csname pn@etpr1.r2r.slide.stall-mem1.ndtw.lo\endcsname{$-$0.0}
\expandafter\def\csname pn@etpr1.r2r.slide.stall-mem1.ndtw.off\endcsname{65.8}
\expandafter\def\csname pn@etpr1.r2r.slide.stall-mem1.ndtw.on\endcsname{66.2}
\expandafter\def\csname pn@etpr1.r2r.slide.stall-mem1.ndtw.rel\endcsname{+1}
\expandafter\def\csname pn@etpr1.r2r.slide.stall-mem1.overcap\endcsname{0}
\expandafter\def\csname pn@etpr1.r2r.slide.stall-mem1.owned\endcsname{4.0}
\expandafter\def\csname pn@etpr1.r2r.slide.stall-mem1.resid.maneuver\endcsname{23}
\expandafter\def\csname pn@etpr1.r2r.slide.stall-mem1.resid.pass\endcsname{71}
\expandafter\def\csname pn@etpr1.r2r.slide.stall-mem1.spl.d\endcsname{$-$0.4}
\expandafter\def\csname pn@etpr1.r2r.slide.stall-mem1.spl.hi\endcsname{+0.5}
\expandafter\def\csname pn@etpr1.r2r.slide.stall-mem1.spl.lo\endcsname{$-$1.2}
\expandafter\def\csname pn@etpr1.r2r.slide.stall-mem1.spl.off\endcsname{55.5}
\expandafter\def\csname pn@etpr1.r2r.slide.stall-mem1.spl.on\endcsname{55.2}
\expandafter\def\csname pn@etpr1.r2r.slide.stall-mem1.spl.rel\endcsname{$-$1}
\expandafter\def\csname pn@etpr1.r2r.slide.stall-mem1.stalls\endcsname{0.3}
\expandafter\def\csname pn@etpr1.r2r.slide.stall-mem1.step_count.d\endcsname{$-$1.2}
\expandafter\def\csname pn@etpr1.r2r.slide.stall-mem1.step_count.hi\endcsname{+0.5}
\expandafter\def\csname pn@etpr1.r2r.slide.stall-mem1.step_count.lo\endcsname{$-$2.9}
\expandafter\def\csname pn@etpr1.r2r.slide.stall-mem1.step_count.off\endcsname{91.7}
\expandafter\def\csname pn@etpr1.r2r.slide.stall-mem1.step_count.on\endcsname{90.5}
\expandafter\def\csname pn@etpr1.r2r.slide.stall-mem1.step_count.rel\endcsname{$-$1}
\expandafter\def\csname pn@etpr1.r2r.slide.stall-mem1.stuck_event_count.d\endcsname{$-$0.1}
\expandafter\def\csname pn@etpr1.r2r.slide.stall-mem1.stuck_event_count.hi\endcsname{$-$0.1}
\expandafter\def\csname pn@etpr1.r2r.slide.stall-mem1.stuck_event_count.lo\endcsname{$-$0.1}
\expandafter\def\csname pn@etpr1.r2r.slide.stall-mem1.stuck_event_count.off\endcsname{0.1}
\expandafter\def\csname pn@etpr1.r2r.slide.stall-mem1.stuck_event_count.on\endcsname{0.0}
\expandafter\def\csname pn@etpr1.r2r.slide.stall-mem1.stuck_event_count.rel\endcsname{$-$100}
\expandafter\def\csname pn@etpr1.r2r.slide.stall-mem1.stuck_flag.d\endcsname{$-$5.4}
\expandafter\def\csname pn@etpr1.r2r.slide.stall-mem1.stuck_flag.hi\endcsname{$-$4.4}
\expandafter\def\csname pn@etpr1.r2r.slide.stall-mem1.stuck_flag.lo\endcsname{$-$6.5}
\expandafter\def\csname pn@etpr1.r2r.slide.stall-mem1.stuck_flag.off\endcsname{5.4}
\expandafter\def\csname pn@etpr1.r2r.slide.stall-mem1.stuck_flag.on\endcsname{0.0}
\expandafter\def\csname pn@etpr1.r2r.slide.stall-mem1.stuck_flag.rel\endcsname{$-$100}
\expandafter\def\csname pn@etpr1.r2r.slide.stall-mem1.success.d\endcsname{$-$0.8}
\expandafter\def\csname pn@etpr1.r2r.slide.stall-mem1.success.hi\endcsname{+0.3}
\expandafter\def\csname pn@etpr1.r2r.slide.stall-mem1.success.lo\endcsname{$-$1.9}
\expandafter\def\csname pn@etpr1.r2r.slide.stall-mem1.success.off\endcsname{64.6}
\expandafter\def\csname pn@etpr1.r2r.slide.stall-mem1.success.on\endcsname{63.8}
\expandafter\def\csname pn@etpr1.r2r.slide.stall-mem1.success.rel\endcsname{$-$1}
\expandafter\def\csname pn@etpr1.r2r.slide.stall-mem1.yields\endcsname{0.10}
\expandafter\def\csname pn@etpr1.r2r.slide.stuck\endcsname{5.4}
\expandafter\def\csname pn@etpr1.r2r.slide.tir\endcsname{16.8}
\expandafter\def\csname pn@etpr1.r2r.swap10\endcsname{4.0}
\expandafter\def\csname pn@etpr1.r2r.swap50\endcsname{32.9}
\expandafter\def\csname pn@etpr1.rxr.auc.ent\endcsname{.647}
\expandafter\def\csname pn@etpr1.rxr.auc.entnorm\endcsname{.668}
\expandafter\def\csname pn@etpr1.rxr.auc.geom\endcsname{.825}
\expandafter\def\csname pn@etpr1.rxr.auc.hist\endcsname{.799}
\expandafter\def\csname pn@etpr1.rxr.auc.marg\endcsname{.617}
\expandafter\def\csname pn@etpr1.rxr.auc.stop\endcsname{.387}
\expandafter\def\csname pn@etpr1.rxr.avail10\endcsname{39.0}
\expandafter\def\csname pn@etpr1.rxr.avail50\endcsname{79.7}
\expandafter\def\csname pn@etpr1.rxr.block.cleansuccess\endcsname{8.7}
\expandafter\def\csname pn@etpr1.rxr.block.cpds\endcsname{50.4}
\expandafter\def\csname pn@etpr1.rxr.block.cpdsmed\endcsname{48}
\expandafter\def\csname pn@etpr1.rxr.block.dirtysuccess\endcsname{16.2}
\expandafter\def\csname pn@etpr1.rxr.block.dsr\endcsname{65.1}
\expandafter\def\csname pn@etpr1.rxr.block.dsrhi\endcsname{70.3}
\expandafter\def\csname pn@etpr1.rxr.block.dsrlo\endcsname{59.6}
\expandafter\def\csname pn@etpr1.rxr.block.nc\endcsname{56.5}
\expandafter\def\csname pn@etpr1.rxr.block.ncfail\endcsname{66.2}
\expandafter\def\csname pn@etpr1.rxr.block.ncsucc\endcsname{27.4}
\expandafter\def\csname pn@etpr1.rxr.block.proxydsr\endcsname{57.9}
\expandafter\def\csname pn@etpr1.rxr.block.shareinsucc\endcsname{10.1}
\expandafter\def\csname pn@etpr1.rxr.block.sr\endcsname{24.9}
\expandafter\def\csname pn@etpr1.rxr.block.stuck\endcsname{81.5}
\expandafter\def\csname pn@etpr1.rxr.block.tir\endcsname{46.6}
\expandafter\def\csname pn@etpr1.rxr.ddsr_pp\endcsname{+14.0}
\expandafter\def\csname pn@etpr1.rxr.dsr_pp\endcsname{+39.5}
\expandafter\def\csname pn@etpr1.rxr.hist.flagged\endcsname{94{,}378}
\expandafter\def\csname pn@etpr1.rxr.hist.recall\endcsname{95.3}
\expandafter\def\csname pn@etpr1.rxr.hist.tp\endcsname{94{,}378}
\expandafter\def\csname pn@etpr1.rxr.kmiss\endcsname{85.4}
\expandafter\def\csname pn@etpr1.rxr.slide.cleansuccess\endcsname{13.5}
\expandafter\def\csname pn@etpr1.rxr.slide.cpds\endcsname{6.3}
\expandafter\def\csname pn@etpr1.rxr.slide.cpdsmed\endcsname{4}
\expandafter\def\csname pn@etpr1.rxr.slide.dirtysuccess\endcsname{50.9}
\expandafter\def\csname pn@etpr1.rxr.slide.dsr\endcsname{79.1}
\expandafter\def\csname pn@etpr1.rxr.slide.dsrhi\endcsname{81.8}
\expandafter\def\csname pn@etpr1.rxr.slide.dsrlo\endcsname{76.1}
\expandafter\def\csname pn@etpr1.rxr.slide.nc\endcsname{10.0}
\expandafter\def\csname pn@etpr1.rxr.slide.ncfail\endcsname{15.8}
\expandafter\def\csname pn@etpr1.rxr.slide.ncsucc\endcsname{6.8}
\expandafter\def\csname pn@etpr1.rxr.slide.proxydsr\endcsname{3.3}
\expandafter\def\csname pn@etpr1.rxr.slide.shareinsucc\endcsname{26.6}
\expandafter\def\csname pn@etpr1.rxr.slide.sr\endcsname{64.4}
\expandafter\def\csname pn@etpr1.rxr.slide.stuck\endcsname{8.7}
\expandafter\def\csname pn@etpr1.rxr.slide.tir\endcsname{16.2}
\expandafter\def\csname pn@etpr1.rxr.swap10\endcsname{8.3}
\expandafter\def\csname pn@etpr1.rxr.swap50\endcsname{33.7}
\expandafter\def\csname pn@range.auc.policymax\endcsname{.673}
\expandafter\def\csname pn@range.auc.policymin\endcsname{.301}
\expandafter\def\csname pn@range.auc.r2r.geom\endcsname{.759--.832}
\expandafter\def\csname pn@range.auc.r2r.hist\endcsname{.891--.981}
\expandafter\def\csname pn@range.dsr.block\endcsname{17.6--42.3}
\expandafter\def\csname pn@range.dsr.slide\endcsname{38.5--41.7}
\expandafter\def\csname pn@range.dsr_pp\endcsname{11.7--28.3}
\expandafter\def\csname pn@range.guard.coll.rel\endcsname{83--98}
\expandafter\def\csname pn@range.guard.n\endcsname{4}
\expandafter\def\csname pn@range.guard.owned\endcsname{8.4--18.4}
\expandafter\def\csname pn@range.guard.recovered\endcsname{49--79}
\expandafter\def\csname pn@range.guard.spl.d\endcsname{6.3--18.7}
\expandafter\def\csname pn@range.guard.stuck.off\endcsname{43.4--68.1}
\expandafter\def\csname pn@range.guard.stuck.on\endcsname{0.0}
\expandafter\def\csname pn@range.guard.success.d\endcsname{6.9--22.4}
\expandafter\def\csname pn@range.hist.recall\endcsname{85.3--99.3}
\expandafter\def\csname pn@range.hist.recall.r2r\endcsname{85.3--99.3}
\expandafter\def\csname pn@range.kmiss\endcsname{77.7--86.0}
\expandafter\def\csname pn@range.nagents\endcsname{5}
\expandafter\def\csname pn@range.nc.block\endcsname{13.9--62.4}
\expandafter\def\csname pn@range.nc.slide\endcsname{4.0--8.0}
\expandafter\def\csname pn@range.stuck.block\endcsname{43.4--68.1}
\expandafter\def\csname pn@range.stuck.slide\endcsname{4.2--6.3}
\expandafter\def\csname pn@range.swap10.r2r\endcsname{0.4--4.0}
\expandafter\def\csname pn@seq2seq.r2r.auc.ent\endcsname{.412}
\expandafter\def\csname pn@seq2seq.r2r.auc.geom\endcsname{.779}
\expandafter\def\csname pn@seq2seq.r2r.auc.hist\endcsname{.931}
\expandafter\def\csname pn@seq2seq.r2r.auc.marg\endcsname{.414}
\expandafter\def\csname pn@seq2seq.r2r.auc.stop\endcsname{.415}
\expandafter\def\csname pn@seq2seq.r2r.avail10\endcsname{3.0}
\expandafter\def\csname pn@seq2seq.r2r.avail50\endcsname{18.1}
\expandafter\def\csname pn@seq2seq.r2r.block.cleansuccess\endcsname{8.2}
\expandafter\def\csname pn@seq2seq.r2r.block.cpds\endcsname{78.8}
\expandafter\def\csname pn@seq2seq.r2r.block.cpdsmed\endcsname{19}
\expandafter\def\csname pn@seq2seq.r2r.block.dirtysuccess\endcsname{2.2}
\expandafter\def\csname pn@seq2seq.r2r.block.dsr\endcsname{20.9}
\expandafter\def\csname pn@seq2seq.r2r.block.dsrhi\endcsname{27.3}
\expandafter\def\csname pn@seq2seq.r2r.block.dsrlo\endcsname{15.8}
\expandafter\def\csname pn@seq2seq.r2r.block.nc\endcsname{53.3}
\expandafter\def\csname pn@seq2seq.r2r.block.ncfail\endcsname{58.7}
\expandafter\def\csname pn@seq2seq.r2r.block.ncsucc\endcsname{6.8}
\expandafter\def\csname pn@seq2seq.r2r.block.proxydsr\endcsname{19.4}
\expandafter\def\csname pn@seq2seq.r2r.block.shareinsucc\endcsname{0.7}
\expandafter\def\csname pn@seq2seq.r2r.block.sr\endcsname{10.4}
\expandafter\def\csname pn@seq2seq.r2r.block.stuck\endcsname{63.4}
\expandafter\def\csname pn@seq2seq.r2r.block.tir\endcsname{32.1}
\expandafter\def\csname pn@seq2seq.r2r.ddsr_pp\endcsname{+20.6}
\expandafter\def\csname pn@seq2seq.r2r.dsr_pp\endcsname{+11.7}
\expandafter\def\csname pn@seq2seq.r2r.hist.flagged\endcsname{445{,}658}
\expandafter\def\csname pn@seq2seq.r2r.hist.recall\endcsname{99.1}
\expandafter\def\csname pn@seq2seq.r2r.hist.tp\endcsname{445{,}657}
\expandafter\def\csname pn@seq2seq.r2r.kmiss\endcsname{85.4}
\expandafter\def\csname pn@seq2seq.r2r.slide.cleansuccess\endcsname{12.9}
\expandafter\def\csname pn@seq2seq.r2r.slide.cpds\endcsname{2.6}
\expandafter\def\csname pn@seq2seq.r2r.slide.cpdsmed\endcsname{1}
\expandafter\def\csname pn@seq2seq.r2r.slide.dirtysuccess\endcsname{9.2}
\expandafter\def\csname pn@seq2seq.r2r.slide.dsr\endcsname{41.5}
\expandafter\def\csname pn@seq2seq.r2r.slide.dsrhi\endcsname{46.4}
\expandafter\def\csname pn@seq2seq.r2r.slide.dsrlo\endcsname{36.8}
\expandafter\def\csname pn@seq2seq.r2r.slide.nc\endcsname{6.1}
\expandafter\def\csname pn@seq2seq.r2r.slide.ncfail\endcsname{7.4}
\expandafter\def\csname pn@seq2seq.r2r.slide.ncsucc\endcsname{1.6}
\expandafter\def\csname pn@seq2seq.r2r.slide.proxydsr\endcsname{1.2}
\expandafter\def\csname pn@seq2seq.r2r.slide.shareinsucc\endcsname{1.2}
\expandafter\def\csname pn@seq2seq.r2r.slide.sr\endcsname{22.1}
\expandafter\def\csname pn@seq2seq.r2r.slide.stuck\endcsname{5.1}
\expandafter\def\csname pn@seq2seq.r2r.slide.tir\endcsname{10.3}
\expandafter\def\csname pn@seq2seq.r2r.swap10\endcsname{3.0}
\expandafter\def\csname pn@seq2seq.r2r.swap50\endcsname{18.1}
\expandafter\def\csname pn@streamvln.mem.M1.false\endcsname{0.14}
\expandafter\def\csname pn@streamvln.mem.M1.prec\endcsname{99.2}
\expandafter\def\csname pn@streamvln.mem.M1.recall\endcsname{42.7}
\expandafter\def\csname pn@streamvln.mem.M8.false\endcsname{9.74}
\expandafter\def\csname pn@streamvln.mem.M8.prec\endcsname{65.0}
\expandafter\def\csname pn@streamvln.mem.M8.recall\endcsname{44.1}
\expandafter\def\csname pn@streamvln.mem.episodes\endcsname{154}
\expandafter\def\csname pn@streamvln.r2r.auc.ent\endcsname{.539}
\expandafter\def\csname pn@streamvln.r2r.auc.geom\endcsname{.832}
\expandafter\def\csname pn@streamvln.r2r.auc.hist\endcsname{.981}
\expandafter\def\csname pn@streamvln.r2r.auc.marg\endcsname{.518}
\expandafter\def\csname pn@streamvln.r2r.auc.stop\endcsname{.358}
\expandafter\def\csname pn@streamvln.r2r.avail10\endcsname{0.4}
\expandafter\def\csname pn@streamvln.r2r.avail50\endcsname{7.7}
\expandafter\def\csname pn@streamvln.r2r.block.cleansuccess\endcsname{22.8}
\expandafter\def\csname pn@streamvln.r2r.block.cpds\endcsname{9.5}
\expandafter\def\csname pn@streamvln.r2r.block.cpdsmed\endcsname{4}
\expandafter\def\csname pn@streamvln.r2r.block.dirtysuccess\endcsname{4.9}
\expandafter\def\csname pn@streamvln.r2r.block.dsr\endcsname{17.6}
\expandafter\def\csname pn@streamvln.r2r.block.dsrhi\endcsname{21.2}
\expandafter\def\csname pn@streamvln.r2r.block.dsrlo\endcsname{14.6}
\expandafter\def\csname pn@streamvln.r2r.block.nc\endcsname{41.4}
\expandafter\def\csname pn@streamvln.r2r.block.ncfail\endcsname{56.7}
\expandafter\def\csname pn@streamvln.r2r.block.ncsucc\endcsname{1.7}
\expandafter\def\csname pn@streamvln.r2r.block.proxydsr\endcsname{15.1}
\expandafter\def\csname pn@streamvln.r2r.block.shareinsucc\endcsname{0.2}
\expandafter\def\csname pn@streamvln.r2r.block.sr\endcsname{27.7}
\expandafter\def\csname pn@streamvln.r2r.block.stall-mem1.abortep\endcsname{47}
\expandafter\def\csname pn@streamvln.r2r.block.stall-mem1.aborts\endcsname{1.1}
\expandafter\def\csname pn@streamvln.r2r.block.stall-mem1.clean_success.d\endcsname{+6.9}
\expandafter\def\csname pn@streamvln.r2r.block.stall-mem1.clean_success.hi\endcsname{+8.2}
\expandafter\def\csname pn@streamvln.r2r.block.stall-mem1.clean_success.lo\endcsname{+5.8}
\expandafter\def\csname pn@streamvln.r2r.block.stall-mem1.clean_success.off\endcsname{22.8}
\expandafter\def\csname pn@streamvln.r2r.block.stall-mem1.clean_success.on\endcsname{29.7}
\expandafter\def\csname pn@streamvln.r2r.block.stall-mem1.clean_success.rel\endcsname{+30}
\expandafter\def\csname pn@streamvln.r2r.block.stall-mem1.collision_count.d\endcsname{$-$194.7}
\expandafter\def\csname pn@streamvln.r2r.block.stall-mem1.collision_count.hi\endcsname{$-$185.4}
\expandafter\def\csname pn@streamvln.r2r.block.stall-mem1.collision_count.lo\endcsname{$-$203.8}
\expandafter\def\csname pn@streamvln.r2r.block.stall-mem1.collision_count.off\endcsname{201.5}
\expandafter\def\csname pn@streamvln.r2r.block.stall-mem1.collision_count.on\endcsname{6.8}
\expandafter\def\csname pn@streamvln.r2r.block.stall-mem1.collision_count.rel\endcsname{$-$97}
\expandafter\def\csname pn@streamvln.r2r.block.stall-mem1.collision_event_count.d\endcsname{$-$40.49}
\expandafter\def\csname pn@streamvln.r2r.block.stall-mem1.collision_event_count.hi\endcsname{$-$36.79}
\expandafter\def\csname pn@streamvln.r2r.block.stall-mem1.collision_event_count.lo\endcsname{$-$44.25}
\expandafter\def\csname pn@streamvln.r2r.block.stall-mem1.collision_event_count.off\endcsname{46.78}
\expandafter\def\csname pn@streamvln.r2r.block.stall-mem1.collision_event_count.on\endcsname{6.29}
\expandafter\def\csname pn@streamvln.r2r.block.stall-mem1.collision_event_count.rel\endcsname{$-$87}
\expandafter\def\csname pn@streamvln.r2r.block.stall-mem1.dirty_success_flag.d\endcsname{+15.5}
\expandafter\def\csname pn@streamvln.r2r.block.stall-mem1.dirty_success_flag.hi\endcsname{+17.5}
\expandafter\def\csname pn@streamvln.r2r.block.stall-mem1.dirty_success_flag.lo\endcsname{+13.5}
\expandafter\def\csname pn@streamvln.r2r.block.stall-mem1.dirty_success_flag.off\endcsname{4.9}
\expandafter\def\csname pn@streamvln.r2r.block.stall-mem1.dirty_success_flag.on\endcsname{20.4}
\expandafter\def\csname pn@streamvln.r2r.block.stall-mem1.dirty_success_flag.rel\endcsname{+317}
\expandafter\def\csname pn@streamvln.r2r.block.stall-mem1.gained\endcsname{430}
\expandafter\def\csname pn@streamvln.r2r.block.stall-mem1.lost\endcsname{18}
\expandafter\def\csname pn@streamvln.r2r.block.stall-mem1.navigation_error_m.d\endcsname{$-$0.8}
\expandafter\def\csname pn@streamvln.r2r.block.stall-mem1.navigation_error_m.hi\endcsname{$-$0.6}
\expandafter\def\csname pn@streamvln.r2r.block.stall-mem1.navigation_error_m.lo\endcsname{$-$0.9}
\expandafter\def\csname pn@streamvln.r2r.block.stall-mem1.navigation_error_m.off\endcsname{6.2}
\expandafter\def\csname pn@streamvln.r2r.block.stall-mem1.navigation_error_m.on\endcsname{5.4}
\expandafter\def\csname pn@streamvln.r2r.block.stall-mem1.navigation_error_m.rel\endcsname{$-$12}
\expandafter\def\csname pn@streamvln.r2r.block.stall-mem1.nc_rate.d\endcsname{$-$36.9}
\expandafter\def\csname pn@streamvln.r2r.block.stall-mem1.nc_rate.hi\endcsname{$-$35.2}
\expandafter\def\csname pn@streamvln.r2r.block.stall-mem1.nc_rate.lo\endcsname{$-$38.4}
\expandafter\def\csname pn@streamvln.r2r.block.stall-mem1.nc_rate.off\endcsname{41.4}
\expandafter\def\csname pn@streamvln.r2r.block.stall-mem1.nc_rate.on\endcsname{4.5}
\expandafter\def\csname pn@streamvln.r2r.block.stall-mem1.nc_rate.rel\endcsname{$-$89}
\expandafter\def\csname pn@streamvln.r2r.block.stall-mem1.ndtw.d\endcsname{+3.8}
\expandafter\def\csname pn@streamvln.r2r.block.stall-mem1.ndtw.hi\endcsname{+4.9}
\expandafter\def\csname pn@streamvln.r2r.block.stall-mem1.ndtw.lo\endcsname{+2.7}
\expandafter\def\csname pn@streamvln.r2r.block.stall-mem1.ndtw.off\endcsname{56.6}
\expandafter\def\csname pn@streamvln.r2r.block.stall-mem1.ndtw.on\endcsname{60.4}
\expandafter\def\csname pn@streamvln.r2r.block.stall-mem1.ndtw.rel\endcsname{+7}
\expandafter\def\csname pn@streamvln.r2r.block.stall-mem1.owned\endcsname{18.4}
\expandafter\def\csname pn@streamvln.r2r.block.stall-mem1.resid.maneuver\endcsname{56}
\expandafter\def\csname pn@streamvln.r2r.block.stall-mem1.resid.pass\endcsname{43}
\expandafter\def\csname pn@streamvln.r2r.block.stall-mem1.spl.d\endcsname{+18.7}
\expandafter\def\csname pn@streamvln.r2r.block.stall-mem1.spl.hi\endcsname{+20.5}
\expandafter\def\csname pn@streamvln.r2r.block.stall-mem1.spl.lo\endcsname{+17.1}
\expandafter\def\csname pn@streamvln.r2r.block.stall-mem1.spl.off\endcsname{25.2}
\expandafter\def\csname pn@streamvln.r2r.block.stall-mem1.spl.on\endcsname{44.0}
\expandafter\def\csname pn@streamvln.r2r.block.stall-mem1.spl.rel\endcsname{+74}
\expandafter\def\csname pn@streamvln.r2r.block.stall-mem1.stalls\endcsname{7.8}
\expandafter\def\csname pn@streamvln.r2r.block.stall-mem1.step_count.d\endcsname{$-$196.6}
\expandafter\def\csname pn@streamvln.r2r.block.stall-mem1.step_count.hi\endcsname{$-$187.9}
\expandafter\def\csname pn@streamvln.r2r.block.stall-mem1.step_count.lo\endcsname{$-$205.7}
\expandafter\def\csname pn@streamvln.r2r.block.stall-mem1.step_count.off\endcsname{294.3}
\expandafter\def\csname pn@streamvln.r2r.block.stall-mem1.step_count.on\endcsname{97.7}
\expandafter\def\csname pn@streamvln.r2r.block.stall-mem1.step_count.rel\endcsname{$-$67}
\expandafter\def\csname pn@streamvln.r2r.block.stall-mem1.stuck_event_count.d\endcsname{$-$2.8}
\expandafter\def\csname pn@streamvln.r2r.block.stall-mem1.stuck_event_count.hi\endcsname{$-$2.4}
\expandafter\def\csname pn@streamvln.r2r.block.stall-mem1.stuck_event_count.lo\endcsname{$-$3.2}
\expandafter\def\csname pn@streamvln.r2r.block.stall-mem1.stuck_event_count.off\endcsname{2.8}
\expandafter\def\csname pn@streamvln.r2r.block.stall-mem1.stuck_event_count.on\endcsname{0.0}
\expandafter\def\csname pn@streamvln.r2r.block.stall-mem1.stuck_event_count.rel\endcsname{$-$100}
\expandafter\def\csname pn@streamvln.r2r.block.stall-mem1.stuck_flag.d\endcsname{$-$43.4}
\expandafter\def\csname pn@streamvln.r2r.block.stall-mem1.stuck_flag.hi\endcsname{$-$41.2}
\expandafter\def\csname pn@streamvln.r2r.block.stall-mem1.stuck_flag.lo\endcsname{$-$45.6}
\expandafter\def\csname pn@streamvln.r2r.block.stall-mem1.stuck_flag.off\endcsname{43.4}
\expandafter\def\csname pn@streamvln.r2r.block.stall-mem1.stuck_flag.on\endcsname{0.0}
\expandafter\def\csname pn@streamvln.r2r.block.stall-mem1.stuck_flag.rel\endcsname{$-$100}
\expandafter\def\csname pn@streamvln.r2r.block.stall-mem1.success.d\endcsname{+22.4}
\expandafter\def\csname pn@streamvln.r2r.block.stall-mem1.success.hi\endcsname{+24.5}
\expandafter\def\csname pn@streamvln.r2r.block.stall-mem1.success.lo\endcsname{+20.3}
\expandafter\def\csname pn@streamvln.r2r.block.stall-mem1.success.off\endcsname{27.7}
\expandafter\def\csname pn@streamvln.r2r.block.stall-mem1.success.on\endcsname{50.1}
\expandafter\def\csname pn@streamvln.r2r.block.stall-mem1.success.rel\endcsname{+81}
\expandafter\def\csname pn@streamvln.r2r.block.stall-mem1.yields\endcsname{0.06}
\expandafter\def\csname pn@streamvln.r2r.block.stuck\endcsname{43.4}
\expandafter\def\csname pn@streamvln.r2r.block.tir\endcsname{23.0}
\expandafter\def\csname pn@streamvln.r2r.ddsr_pp\endcsname{+20.9}
\expandafter\def\csname pn@streamvln.r2r.dsr_pp\endcsname{+28.3}
\expandafter\def\csname pn@streamvln.r2r.hist.flagged\endcsname{365{,}449}
\expandafter\def\csname pn@streamvln.r2r.hist.recall\endcsname{98.6}
\expandafter\def\csname pn@streamvln.r2r.hist.tp\endcsname{365{,}449}
\expandafter\def\csname pn@streamvln.r2r.kmiss\endcsname{80.3}
\expandafter\def\csname pn@streamvln.r2r.slide.cleansuccess\endcsname{34.4}
\expandafter\def\csname pn@streamvln.r2r.slide.cpds\endcsname{4.0}
\expandafter\def\csname pn@streamvln.r2r.slide.cpdsmed\endcsname{2}
\expandafter\def\csname pn@streamvln.r2r.slide.dirtysuccess\endcsname{21.5}
\expandafter\def\csname pn@streamvln.r2r.slide.dsr\endcsname{38.5}
\expandafter\def\csname pn@streamvln.r2r.slide.dsrhi\endcsname{41.5}
\expandafter\def\csname pn@streamvln.r2r.slide.dsrlo\endcsname{35.6}
\expandafter\def\csname pn@streamvln.r2r.slide.nc\endcsname{6.8}
\expandafter\def\csname pn@streamvln.r2r.slide.ncfail\endcsname{13.2}
\expandafter\def\csname pn@streamvln.r2r.slide.ncsucc\endcsname{1.6}
\expandafter\def\csname pn@streamvln.r2r.slide.proxydsr\endcsname{2.6}
\expandafter\def\csname pn@streamvln.r2r.slide.shareinsucc\endcsname{3.3}
\expandafter\def\csname pn@streamvln.r2r.slide.sr\endcsname{56.0}
\expandafter\def\csname pn@streamvln.r2r.slide.stuck\endcsname{5.7}
\expandafter\def\csname pn@streamvln.r2r.slide.tir\endcsname{4.9}
\expandafter\def\csname pn@streamvln.r2r.swap10\endcsname{0.4}
\expandafter\def\csname pn@streamvln.r2r.swap50\endcsname{7.7}

\newcommand{\pn}[1]{\ifcsname pn@#1\endcsname\csname pn@#1\endcsname\else\TBD\fi}

\newcommand{\TBD}{\textcolor{red}{\textsc{tbd}}}

\newif\ifdraft\drafttrue
\ifdraft
  \newcommand{\todo}[1]{\textcolor{red}{\small[\textsc{todo}: #1]}}
\else
  \newcommand{\todo}[1]{}
\fi

\ifdraft
  \newcommand\hieu[1]{\noindent{\color{blue} {\bf \fbox{Hieu}} {\it#1}}}
  \newcommand\lqan[1]{\noindent{\color{red} {\bf \fbox{An}} {\it#1}}}
  \newcommand\drdung[1]{\noindent{\color{green} {\bf \fbox{DrDung}} {\it#1}}}
\else
  \newcommand\hieu[1]{}
  \newcommand\lqan[1]{}
  \newcommand\drdung[1]{}
\fi

\title[Successful but Unsafe]{Successful but Unsafe: An Analytical Benchmark for Revealing Collisions in Vision-and-Language Navigation}

\ifprismAnonymousModeSelected
  \author{Anonymous Author(s)}
  \affiliation{%
    \institution{Anonymous Institution}
    \city{Anonymous City}
    \country{Anonymous Country}}
  \email{anonymous@example.org}
\else
  \author{Hieu D. Pham}
  \affiliation{%
    \institution{Center for AI Research, VinUniversity}
    \city{Hanoi}
    \country{Vietnam}}
  \email{24hieu.pd@vinuni.edu.vn}
  \author{Quoc-An Luong}
  \affiliation{%
    \institution{Center for AI Research, VinUniversity}
    \city{Hanoi}
    \country{Vietnam}}
  \email{an.lq@vinuni.edu.vn}
  \author{Dung D. Le}
  \affiliation{%
    \institution{Center for AI Research, VinUniversity}
    \city{Hanoi}
    \country{Vietnam}}
  \email{dung.ld@vinuni.edu.vn}
\fi

\begin{abstract}
Vision-and-language navigation (VLN) agents are typically evaluated by whether they reach a target and how efficiently they follow an instructed route, yet these metrics provide little information about collision behavior. Moreover, continuous VLN evaluation commonly uses a sliding regime, where an agent can slide along obstacles after a collision, potentially masking the effect of unsafe actions on navigation performance. We introduce \benchname{}, a benchmark for analyzing collision safety under two complementary evaluation regimes: sliding, reflecting common VLN evaluation practice, and blocking, where collision-inducing actions prevent forward motion and more directly expose their consequences. Across six published agents on R2R-CE, 38.5-56.8\% of successful episodes contain contact under sliding, while switching to blocking decreases success rate by 11.7-36.1 percentage points. We further analyze signals for anticipating imminent collisions and find that depth geometry and recent movement history are more informative than policy-derived signals, while safe alternatives near the policy's preferred action are rarely available. Motivated by these findings, we introduce a training-free collision guard that vetoes risky forward actions, executes an escape maneuver, and returns control to the original policy. Under blocking, the guard reduces collision steps by 62-98\% and improves success rate by 6.9-29.3 percentage points across five recurrent, topological, and vision-language-action agents. Under sliding, it leaves success rate unchanged while raising the share of collision-free successes by 4.1-12.4 points. These results reveal a substantial gap between task success and collision-safe navigation and show that lightweight collision-aware intervention can improve both safety and navigation performance. \footnote{Code is publicly available at \url{https://github.com/dinhieufam/vln-safe}}
\end{abstract}


\begin{document}
\maketitle

%======================================================================
\section{Introduction}
\label{sec:intro}

\begin{figure}[!t]
\centering
\includegraphics[width=\columnwidth]{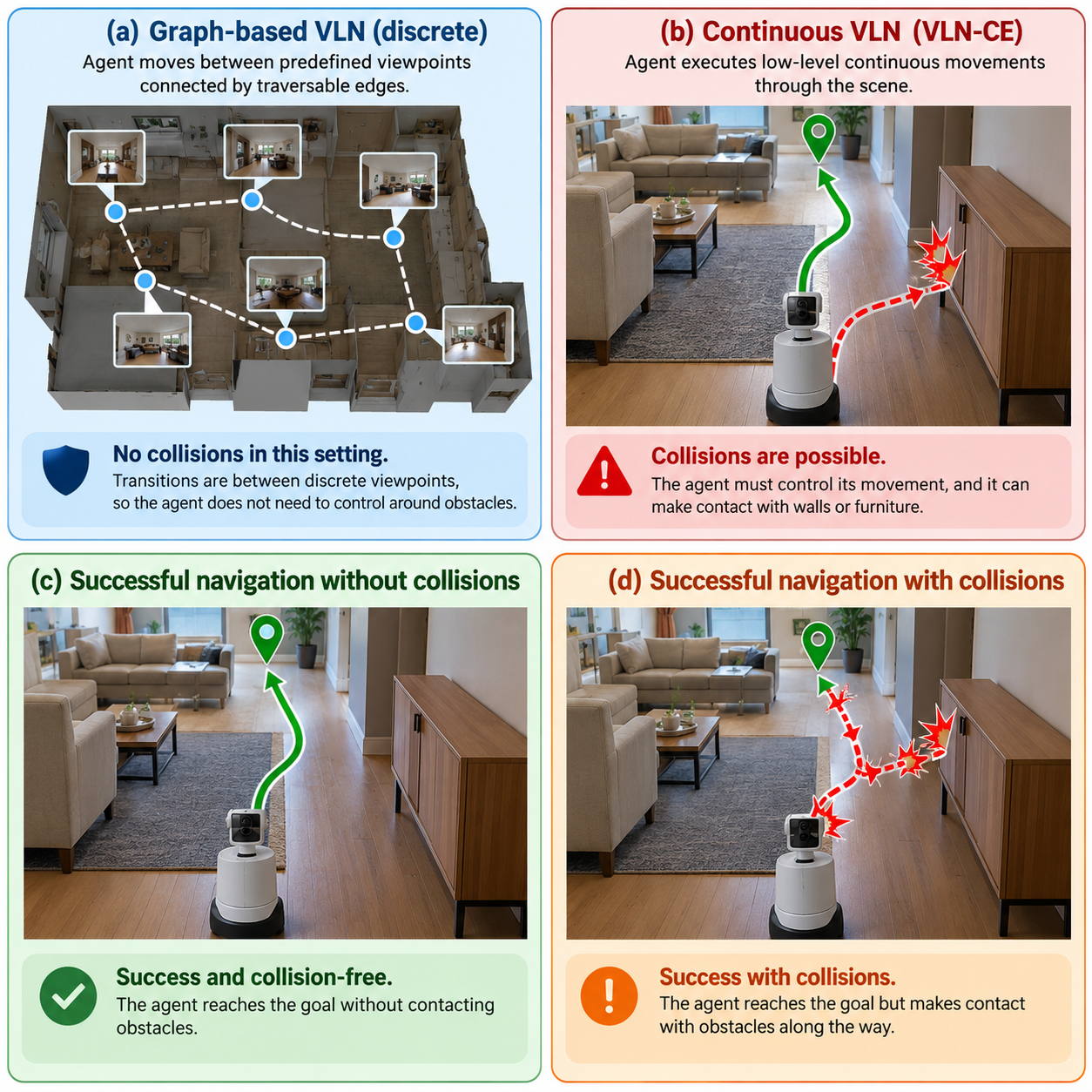}
\caption{Overview of the VLN settings and contact outcomes. (a) Graph-based VLN moves between predefined viewpoints. (b) Continuous VLN-CE executes low-level movements that can contact obstacles. (c,d) Successful navigation can be collision-free or include contact along the route.}
\label{fig:setting-visualization}
\Description{Four panels compare navigation settings and outcomes. (a) A route follows edges between fixed viewpoints. (b) A continuous route contacts furniture. (c) A successful route reaches the goal without obstacle contact. (d) A successful route reaches the goal after several contacts.}
\end{figure}

Vision-and-language navigation (VLN) requires an agent to follow natural-language instructions to reach a goal in an environment~\cite{anderson2018r2r}. Its applications to household and service robots have drawn steadily growing attention~\cite{gu2022vlnsurvey,zhang2024vlnsurvey}. VLN agents are commonly evaluated by success rate (SR), success weighted by path length (SPL), and normalized dynamic time warping (nDTW)~\cite{anderson2018spl,ilharco2019ndtw}. These metrics assess goal completion, path efficiency, and similarity to a reference route, but they do not measure collisions with the environment. An agent may therefore complete a navigation task while making contact with walls or furniture along the way, which would undermine its usefulness on a real robot. Evaluating such contact is necessary to know whether successful navigation is also physically safe.

This gap in evaluation reflects how VLN developed. Early approaches followed the graph-based setting, in which agents moved between predefined viewpoints connected by traversable edges~\cite{anderson2018r2r}. These transitions did not require control around obstacles, so collisions were not part of the task. Subsequent work moved to continuous environments (VLN-CE), where agents execute low-level movement actions through the scene and contact with obstacles becomes possible~\cite{krantz2020vlnce}. However, the standard navigation metrics still focus on reaching the goal and following an appropriate route. Collision measurements are therefore needed alongside these metrics to assess the physical behavior of an agent. Figure~\ref{fig:setting-visualization} contrasts these settings and shows how a successful continuous episode may still contain contact.

To address this gap, we introduce \benchname{}, a benchmark for evaluating contact-related safety in continuous vision-and-language navigation. Rather than reporting a single aggregate collision rate, \benchname{} organizes evaluation around two complementary questions that characterize contact during navigation and establish how evaluation choices affect the reported measurements:

\begin{itemize}
  \item \textbf{P1: Do successful episodes also avoid collisions?} We measure how often contact occurs in successful and unsuccessful episodes, distinguishing task completion from collision-free navigation.
  \item \textbf{P2: How do evaluation choices affect the measurements?} We compare simulator collision-handling settings and methods for detecting collisions to assess their influence on reported performance.
\end{itemize}

Measuring contact reveals when navigation succeeds without avoiding collisions, but it does not tell us which information could help prevent them. We therefore conduct a separate analysis of collision-prediction signals, comparing information from the navigation policy with two sources outside it: depth geometry and the agent's own movement history. The analysis finds that contact is predictable from outside the policy but not from within it, and that reranking the policy's candidate choices rarely has a safe alternative to choose. It motivates a training-free collision guard that vetoes forward steps predicted to collide, executes a collision-free escape, and returns control to the unchanged policy. Where the simulator does not let an agent slide along obstacles, contact stalls the agent, and removing it recovers success rather than trading it away.

In summary, our contributions are:

\begin{enumerate}
  \item \textbf{A benchmark for contact-related safety in VLN.} We introduce \benchname{} to measure contact in successful and unsuccessful episodes and to examine how simulator settings and collision definitions affect the measurements (Sections~\ref{sec:bench} and~\ref{sec:bench-results}).
  \item \textbf{An analysis of collision-prediction signals.} We compare the policy's own candidate scores, a depth-based geometric predictor, and movement history, and measure how often a safe alternative candidate exists at a risky decision. This analysis identifies which information sources predict contact and informs the design of our method (Section~\ref{sec:signal-analysis} and~\ref{sec:predict}).
  \item \textbf{A training-free collision guard.} We propose a runtime layer that uses geometry and movement history to veto forward steps, without retraining the policy or reading its scores. On recurrent, topological, and vision-language-action agents it removes most collision steps, raises success where sliding is disabled, and yields cleaner successes at no success cost where it is enabled (Sections~\ref{sec:guard} and~\ref{sec:guard-results}).
\end{enumerate}

%======================================================================
\section{Related Work}
\label{sec:related}

\textbf{Vision-and-language navigation.}
Vision-and-language navigation (VLN) is a multimodal task in which an agent follows a natural-language instruction (e.g., ``go downstairs, turn left, and stop by the kitchen table'') using visual observations to navigate a real or simulated environment. Early work formulated VLN as instruction following over discrete navigation graphs, where agents moved among connected viewpoints. Room-to-Room (R2R) paired instructions with the Matterport3D navigation graph~\cite{anderson2018r2r,chang2017matterport3d}. Agents on these graphs improved through pretrained transformers, topological memory, and large-scale data augmentation~\cite{hao2020prevalent,chen2021hamt,chen2022duet,wang2023scalevln}. VLN-CE then transferred the task to continuous Habitat environments, where agents execute low-level movements through the scene rather than moving only between predefined viewpoints~\cite{krantz2020vlnce,savva2019habitat}. The benchmark landscape also broadened to include longer routes, multilingual instructions, and object-oriented goals~\cite{jain2019r4r,ku2020rxr,qi2020reverie}.

Continuous VLN agents use several decision interfaces. Some predict and select candidate waypoints~\cite{krantz2021waypoint,hong2022cwp,krantz2022sim2sim}, whereas others construct metric, volumetric, or topological representations~\cite{an2023bevbert,wang2024hnr,wang2025g3dlf,an2025etpnav,ye2025etpr1,ye2026toporl}. Navigation vision-language-action models instead generate discretized motion actions from visual observations~\cite{zhang2024navid,zhang2025uninavid,cheng2025navila,wei2025streamvln,zeng2026janusvln,zhu2026navida,wang2026livevln,peng2026solnav}. Other recent systems combine a vision-language planner with a trajectory controller or use a multimodal large language model (MLLM) to select predicted waypoints~\cite{wei2025dualvln,bao2026goal2pixel,liu2025navforesee,dang2026rynnbrain,zhang2026qwenrobotnav,zhou2024navgpt,chen2024mapgpt,long2024instructnav,qiao2025opennav,canav,li2025obstaclemapvln,li2026oneagent,liu2026hamvln,li2026agenticnav,deng2026spacevln,pan2026o2cnav}. Despite these architectural differences, evaluation remains centered on SR, SPL, and nDTW~\cite{anderson2018spl,ilharco2019ndtw}. These metrics measure task completion, efficiency, and route fidelity, but do not indicate whether navigation involved physical contact. Our work complements these metrics by evaluating contact-related behavior in frozen VLN-CE agents.

\textbf{Safety in embodied navigation.}
Safe reinforcement learning and runtime safety filters seek to keep an agent within an acceptable set of states or actions~\cite{garcia2015safe}. Relevant approaches include control barrier functions and their learned or latent variants~\cite{ames2019cbf,latentcbf,anysafe}, model-free shields~\cite{mpshielding}, conformal wrappers that account for perception uncertainty~\cite{zhou2024adaptiveconformalcbf,feldman2025sparsehumanconformal,doula2025safepath}, VLM-regulated filters~\cite{chen2026alphaadj}, and methods for verifying foundation-model controllers~\cite{corsi2026verifiable}. Many of these methods require some combination of an explicit safety specification, state estimate, dynamics model, or calibrated uncertainty. Such information may not be directly available from a language-conditioned navigation policy operating in an unseen environment.

Related navigation methods include training-free repulsive control for visual navigation~\cite{kim2025care}, attention-guided action filtering, and post-training safety alignment for vision-language-action policies~\cite{attnvlafilter,actonwhatyousee}. These methods demonstrate several ways of modifying an agent's behavior, but they do not establish how contact should be measured alongside task success on standard VLN-CE episodes. We instead begin with an evaluation protocol and then study whether a lightweight intervention can use signals available from an existing frozen policy.

\textbf{Collision-aware VLN.}
Several works address obstacles or contact within VLN. Safe-VLN trains a LiDAR-masked waypoint predictor and applies candidate reselection after a collision~\cite{yue2024safevln}; ObVLN introduces obstacle-aware training~\cite{hong2024obvln}; and SafeVLA learns a constrained navigation policy~\cite{zhang2025safevla}. VLN-CM uses a depth-derived occupancy map to select collision-free step lengths without task-specific training~\cite{jeong2024vlncm}. These methods primarily improve collision avoidance through retraining, constrained action generation, or reactive correction. ETPNav's controller also includes a Tryout heuristic that retries a blocked step along alternative headings~\cite{an2025etpnav}. It is specific to controllers that execute multi-step waypoint plans, and we disable it in all runs (Section~\ref{sec:bench}). Our guard instead leaves the frozen policy's decisions intact. It vetoes only forward primitives identified as blocked by geometry or movement history, then returns control to the policy after a collision-free escape.

Other work studies specific forms of contact or evaluates navigation in newly constructed environments. HA-VLN and AdaVLN introduce moving humans and report human-contact measures~\cite{li2025havln,loh2024adavln}. Physics-enabled simulators provide collision measurements for new scenes or embodiments~\cite{wang2025vlnpe,lin2025vlnverse,pham2026humanoidvln, wu2026navverse}, while TouchSafeBench evaluates whether vision-language models can anticipate contact from egocentric observations~\cite{wang2026touchsafebench}. These efforts broaden the study of navigation safety. Our benchmark addresses a complementary question: how much contact occurs when published agents are evaluated on existing VLN-CE tasks, and how strongly that conclusion depends on the simulator configuration and collision definition.

\textbf{Diagnostic analysis of VLN agents.}
Prior analyses examine the capabilities that limit navigation performance~\cite{wang2026whatlimits,failureattribution}. Other studies evaluate robustness to viewpoint changes, infeasible instructions, dynamic scenes, long navigation horizons, and test-time distribution shift~\cite{sun2025vil,vlnnf,sun2025dynamicvln,song2025lhvln,gao2024fstta,atena,uncertaintygaussian,groundcontrol}. These studies primarily explain changes in task completion or navigation accuracy. We analyze a different dimension of behavior: when and how physical contact occurs in both successful and unsuccessful episodes. We also test whether signals available during the policy's forward pass can predict upcoming contact, and compare them with external geometric information.

%======================================================================
\section{Method}
\label{sec:method}

\begin{figure*}[h!]
\centering
\includegraphics[width=\textwidth]{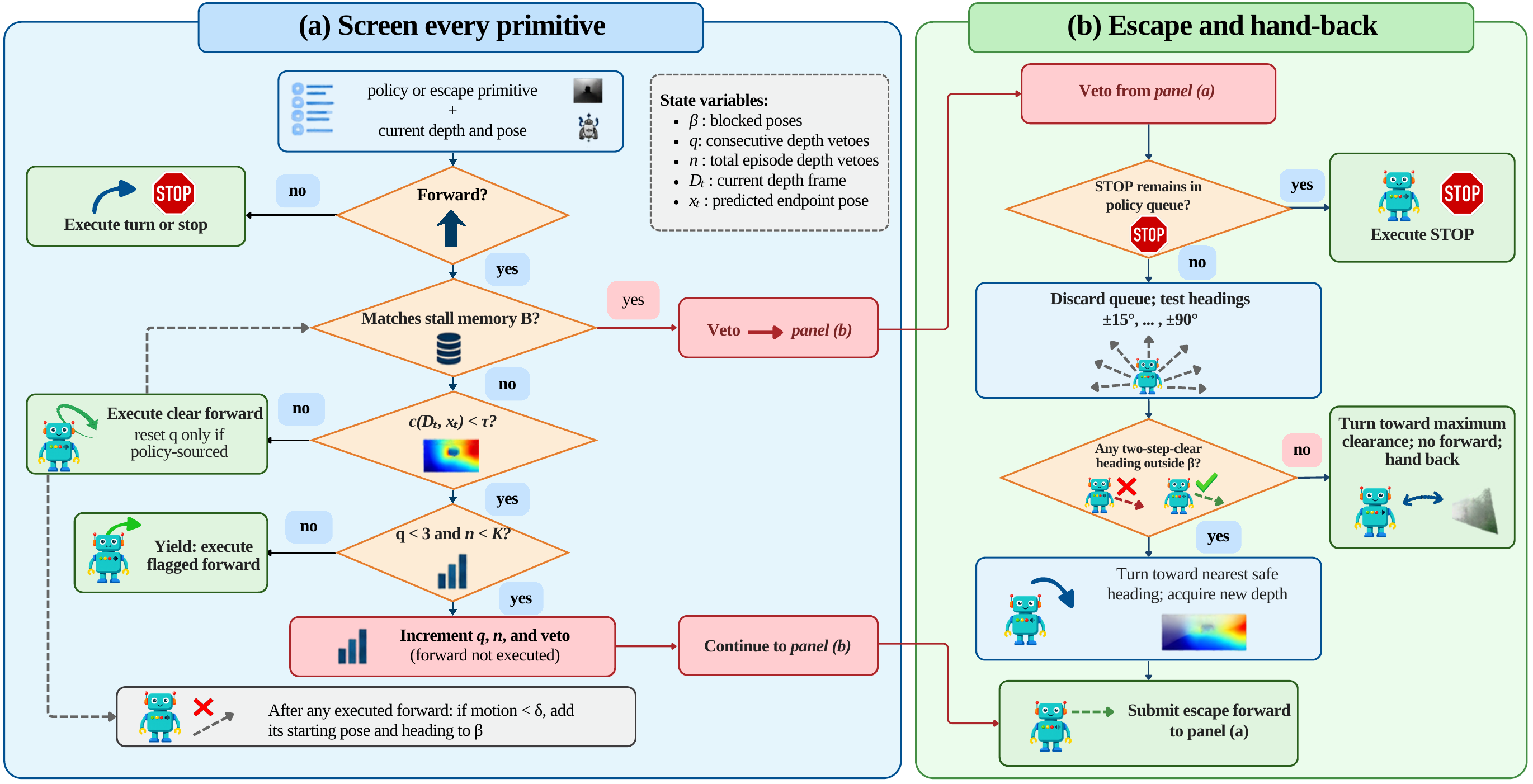}
\caption{The collision guard. (a) Every forward primitive, from the policy or from an escape, is screened by the stall memory $\mathcal B$ and the depth clearance test $c(D_t,x_t)<\tau$; turns and stops pass unchanged, and two counters ($q$ consecutive and $n$ per-episode depth vetoes) let a flagged step through rather than trap the agent. (b) A veto executes a pending \textsc{stop}, or turns to the nearest heading whose two-step path is clear and outside $\mathcal B$, submits one escape forward back to (a), and returns control to the policy.}
\label{fig:guard-pipeline}
\Description{A two-panel flowchart. The first panel screens policy and escape primitives using stall memory, depth clearance, and two veto counters. A veto enters the second panel, which handles STOP, selects an escape heading, and returns an escape forward to the first panel for another check.}
\end{figure*}

\subsection{The \benchname{} Benchmark}
\label{sec:bench}

\benchname{} extends standard VLN-CE evaluation with measurements of physical contact. It does not replace SR, SPL, or nDTW; it determines whether behavior these metrics classify as successful is also collision-free. Frozen policies are evaluated through two probes: P1 relates contact to episode outcome, and P2 measures how sensitive these findings are to the simulator's collision-handling regime and to the predicate used to identify contact.

\subsubsection{P1: Contact-Aware Episode Evaluation}
\label{sec:p1} 

An episode succeeds when the agent emits \textsc{stop} within 3\,m geodesic distance of the goal; it ends at \textsc{stop} or after 500 primitive actions, and a timeout is a failure. Habitat flags a commanded translation as a collision when its navigation-mesh filter shortens the requested movement. With $p_t$, $p_t^{\star}$, and $p_t'$ the starting, commanded, and filtered positions, we use the simulator flag
\[
\kappa_{\rho,t}=
\mathbf{1}\!\left[
\lVert p_t'-p_t\rVert^2+10^{-5}
<
\lVert p_t^{\star}-p_t\rVert^2
\right],
\]
where $\rho$ is the collision-handling regime. $\kappa_{\rho,t}=1$ means that the navigation mesh shortened the commanded translation, not that rigid bodies intersected; only \textsc{forward} actions can set it.

A primitive with $\kappa_{\rho,t}=1$ is a \emph{collision step}, and a maximal run of consecutive collision steps is a \emph{collision event}, which separates isolated contact from repeated attempts to move through the same boundary. An episode is a \emph{dirty success} if it succeeds and contains at least one collision step, and a \emph{clean success} otherwise. The central P1 statistic is the dirty-success rate
\[
\mathrm{DSR}=\Pr[\text{contact}\mid\text{success}]
=\frac{\text{dirty successes}}{\text{successful episodes}},
\]
reported with a Wilson 95\% interval alongside the joint distribution of outcome and contact status. We also report collision steps and events per episode, the share of collision steps that occur in successes, and the per-episode collision-step rate (collision steps divided by executed primitives, averaged over episodes). A depth-based clearance profile (minimum clearance, time near surfaces, near misses) complements these flags.

\subsubsection{P2: Sensitivity to Evaluation Choices}
\label{sec:p2}

The first evaluation choice is Habitat's collision-handling regime. Under $\rho=\slideon{}$, a target outside the navigation mesh is projected back onto it, so the agent moves tangentially along the boundary. Under $\rho=\slideoff{}$, the same forward command is rejected and the agent stays at or near its starting position. Turns are unaffected in both. We run the same policy on the same episodes, in the same order and with the same seed, under both regimes, and test the paired differences with McNemar's exact test for success and a paired bootstrap for DSR.

The second choice is the collision predicate. Because the simulator flag is not always exposed, some VLN implementations infer contact from a failure to move. We reproduce this displacement proxy as
$\hat{\kappa}_t=\mathbf{1}[a_t=\textsc{forward}\wedge\lVert p_{t+1}-p_t\rVert<10^{-4}\,\mathrm{m}]$.
It misses a shortened translation whenever sliding still moves the agent, so we report its \emph{miss rate}: the share of episodes with engine contact that $\hat{\kappa}$ classifies as clean, measured under \slideon{}. A \emph{stuck event} begins after five consecutive forward actions with $\hat{\kappa}_t=1$; because sliding preserves displacement, we interpret it only within a regime.

\subsection{Collision-Prediction Signal Analysis}
\label{sec:signal-analysis}

\subsubsection{Decision Interfaces}
\label{sec:interfaces}

Agents expose different types of scored choices. A primitive-action interface (\Tone{}) assigns scores to immediate actions: move forward, turn, or stop. A spatial-candidate interface (\Ttwo{}) instead scores a variable-size set of waypoints or navigation subgoals, and a controller translates the selected candidate into a sequence of primitive actions. Under the evaluated action space and contact predicate, only a forward primitive can set the collision flag; a turn or a stop is therefore contact-free. This shared primitive interface is where the method intervenes.

\subsubsection{Prediction Task and Signals}

At each logged decision point, we label whether an engine collision occurs within the next three executed primitives under \slideoff{}. We use the same fixed horizon for both interfaces and evaluate signals from three sources. \emph{Policy outputs:} normalized candidate entropy, top-two margin, and stop probability from the logged score distribution. For \Tone{}, whose candidate set always has four actions, we use the raw categorical entropy. \emph{Geometry:} predicted clearance for the pending forward step, computed from the depth point cloud of Section~\ref{sec:setup}. \emph{Movement history:} whether the agent is within 1\,cm and half of one turn increment of a pose from which an earlier forward step in the episode moved less than 1\,cm. This signal requires only odometry. These are the outputs the released checkpoints expose; the analysis asks whether risk is accessible from the deployed score interface, not whether hidden representations encode it.

\subsection{A Training-Free Collision Guard}
\label{sec:guard}

The guard is a runtime layer between a frozen policy and the simulator. It sees every primitive the policy proposes, the current depth frame, and odometry; it never reads the policy's scores or changes its weights, and it intervenes only on forward primitives (Figure~\ref{fig:guard-pipeline}). Forward steps generated by the guard itself pass through the same checks, using the depth frame acquired after the escape turn.

\textbf{Stall veto.} After every executed forward step the guard compares the new pose with the old one. If the agent moved less than $\delta = 1$\,cm, the starting pose and heading join the episode's blocked set $\mathcal{B}$. A later forward step is vetoed when its position is within $\delta$ of a stored pose and its heading within half a turn increment. As Section~\ref{sec:predict} shows, nearly every forward step this rule flags would have collided. The rule removes repeated presses, but the first blocked step at a new pose is still needed to create the memory entry.

\textbf{Depth veto.} Before a forward step, the guard projects the current depth frame to an occupancy point cloud and computes the clearance $c(D_t, x_t)$ between the agent's body and the nearest point at the pose the step would end in. It vetoes the step when $c < \tau = 0.05$\,m. The guard uses only the current frame because accumulated frames can retain obstacle points after they leave the camera's view. The eight-frame guard also lowers success (Section~\ref{sec:ablation}).

Two counters keep the depth veto from trapping the agent. The consecutive counter $q$ increases after each depth veto and resets only when a forward step proposed by the policy is found clear; once $q=3$, depth-flagged forward steps are allowed through (a \emph{yield}) until that reset. After $K=12$ depth vetoes in an episode, all later depth-flagged forwards yield. Stall vetoes neither increment nor obey these caps. We fixed $K=12$ before evaluation as a fallback for policies without a finite decision budget, above the maximum of 9 depth vetoes per ETP-R1 episode in a held-out R2R-CE validation-seen run (95th percentile 3). It rarely binds: in our evaluation runs it is never reached by ETP-R1 or ETPNav under \slideoff{} on R2R-CE, and reached in at most 1.5\% of R2R-CE and 3.0\% of RxR-CE episodes for any agent.

\textbf{Escape and hand-back.} If the policy's remaining queue already contains \textsc{stop}, the guard executes it. Otherwise a veto discards the queue and examines headings at $15^\circ$ increments up to $90^\circ$ on either side using the pre-turn depth frame. A heading qualifies when the projected two-step path stays at least $\tau$ from observed obstacles and the predicted pose does not match $\mathcal B$; the guard takes the smallest qualifying turn. If none qualifies, it turns toward the greatest observed clearance, takes no forward step, and returns control. After a qualifying turn, the escape forward is screened like any other, so a veto can trigger another escape. Control then returns to the policy, which re-decides from the reached pose: a \Ttwo{} agent re-plans on its topological graph, and a \Tone{} agent receives the next observation. For ETP-R1, re-decisions count against its released 15-decision budget.

\textbf{What the guard needs.} Panoramic depth, odometry, access to the controller's queued primitives, and the ability to intercept an action before execution. It needs no candidate set, scores, weights, or training data, so the same logic and the same metric thresholds ($\tau$, $\delta$, the consecutive limit, and $K$) apply to every agent and both decision interfaces. Because depth can miss navigation-mesh boundaries and the caps deliberately let flagged steps through, this is a heuristic runtime guard, not a formal collision-avoidance guarantee.
%======================================================================
\section{Experimental Setup}
\label{sec:setup}

\textbf{Simulator and benchmarks.} We use Habitat~\cite{savva2019habitat} with Matterport3D~\cite{chang2017matterport3d} to simulate the indoor environment. Habitat does not simulate rigid-body dynamics for the agent: motion is constrained to a navigation mesh baked for a cylindrical body, and a forward step whose endpoint leaves the mesh is either projected back onto it (\slideon{}) or rejected (\slideoff{}). We use R2R-CE~\cite{krantz2020vlnce,anderson2018r2r} and RxR-CE~\cite{ku2020rxr} as sources of navigation instructions and ground-truth trajectories. R2R-CE is the primary benchmark, while RxR-CE, whose camera rig, turn angle, and body dimensions differ, serves as a generality check. We evaluate on all 1{,}839 episodes of the unseen validation split of R2R-CE and on 1{,}223 episodes of that of RxR-CE, one per distinct trajectory and language group to avoid RxR's near-duplicate rows.

\textbf{Agents.} We evaluate six agents from three architectural families: the VLN-CE recurrent baselines Seq2Seq and CMA~\cite{krantz2020vlnce}; the navigation VLAs StreamVLN~\cite{wei2025streamvln} and OmniNav~\cite{xue2025omninav}; and the topological planners ETPNav~\cite{an2025etpnav} and ETP-R1~\cite{ye2025etpr1}. We use the authors' released checkpoints where available and otherwise reproduce the agents from their papers. All policies remain frozen and run through a common harness.

\textbf{Instrumentation.} At each primitive we log $\kappa_\rho$, $\hat{\kappa}$, and depth from a twelve-camera panoramic ring, projected into a metric occupancy point cloud with a per-pixel floor test so that low obstacles are not discarded with the floor. Camera intrinsics and mounting height are read from each benchmark's configuration. At every decision we also log the policy's full score distribution, so the signal analysis of Section~\ref{sec:signal-analysis} is computed offline from the same runs.

\textbf{Statistics.} Benchmark rates carry Wilson 95\% intervals. Guard results are paired on episode: guarded and unguarded runs share episodes, order, and seed, and every difference carries a 95\% bootstrap interval over episodes (2{,}000 resamples). All numbers are regenerated from per-episode and per-step logs by the released analysis scripts.

\textbf{Protocol and scope.} Policies are evaluated without retraining: checkpoint weights, configuration, tokenizer, instructions, candidate or waypoint generator, prompts, and stopping rule are unchanged. The harness adapts only the execution layer, so every decision is executed as primitive simulator transitions: the topological agents' released control-based mode, and quantized primitives instead of a waypoint teleport for OmniNav. Contact in \benchname{} is deliberately narrow: the navigation mesh shortened a commanded translation. It carries no force, speed, damage, or obstacle semantics, does not model dynamic agents or objects absent from the mesh, and therefore measures contact-related behavior rather than certifying physical safety.
%======================================================================
\section{Results}
\label{sec:results}

\subsection{Benchmark Results}
\label{sec:bench-results}

% Machine-written by analysis/make_tables.py from results/summary/*.json -- do not edit by hand.
% Machine-written by analysis/make_tables.py -- do not edit by hand.
\begin{table*}[h!]
\centering\small
\setlength{\tabcolsep}{3.5pt}
\caption{Benchmark results under \slideon{} and \slideoff{} (\%). Small colored
values beside \slideoff{} report \slideoff{}$-$\slideon{} changes:
\textcolor{worse}{red} is worse under \slideoff{} and
\textcolor{better}{green} is better, following the arrows. Contact
uses $\kappa_\rho$; DSR is
$\Pr[\text{contact}\mid\text{success}]$, stuck denotes any proxy stuck event,
and proxy miss is measured under \slideon{}. Bold marks the primary P1 and
predicate statistics.}
\label{tab:profile}
\begin{tabular}{@{}llccccccccc@{}}
\toprule
& & \multicolumn{2}{c}{Success rate $\uparrow$}
& \multicolumn{2}{c}{\NC{} rate $\downarrow$}
& \multicolumn{2}{c}{Dirty success $\downarrow$}
& \multicolumn{2}{c}{Stuck $\downarrow$}
& \multicolumn{1}{c}{Proxy miss $\downarrow$} \\
\cmidrule(lr){3-4}\cmidrule(lr){5-6}\cmidrule(lr){7-8}
\cmidrule(lr){9-10}\cmidrule(l){11-11}
Agent & Tier & \slideon{} & \slideoff{} & \slideon{} & \slideoff{}
& \textbf{\slideon{}} & \slideoff{} & \slideon{} & \slideoff{}
& \textbf{\slideon{}} \\
\midrule
\multicolumn{11}{@{}l}{\textit{R2R-CE, 1{,}839 episodes}}\\
ETP-R1 & \Ttwo{} & 64.6 & 50.5\badchange{$-14.1$} & 4.0 & 13.9\badchange{$+9.9$} & \textbf{41.7} & 42.3\badchange{$+0.6$} & 5.4 & 44.4\badchange{$+39.0$} & \textbf{77.7} \\
ETPNav & \Ttwo{} & 56.9 & 41.5\badchange{$-15.4$} & 4.9 & 21.0\badchange{$+16.1$} & \textbf{40.2} & 31.8\goodchange{$-8.4$} & 4.2 & 43.8\badchange{$+39.6$} & \textbf{85.0} \\
OmniNav & \Tone{} & 67.5 & 31.4\badchange{$-36.1$} & 4.9 & 30.5\badchange{$+25.6$} & \textbf{56.8} & 48.4\goodchange{$-8.4$} & 5.3 & 70.5\badchange{$+65.2$} & \textbf{87.6} \\
StreamVLN & \Tone{} & 56.0 & 27.7\badchange{$-28.3$} & 6.8 & 41.4\badchange{$+34.6$} & \textbf{38.5} & 17.6\goodchange{$-20.9$} & 5.7 & 43.4\badchange{$+37.7$} & \textbf{80.3} \\
CMA & \Tone{} & 30.2 & 12.1\badchange{$-18.1$} & 8.0 & 62.4\badchange{$+54.4$} & \textbf{40.9} & 23.3\goodchange{$-17.6$} & 6.3 & 68.1\badchange{$+61.8$} & \textbf{83.9} \\
Seq2Seq & \Tone{} & 22.1 & 10.4\badchange{$-11.7$} & 6.1 & 53.3\badchange{$+47.2$} & \textbf{41.5} & 20.9\goodchange{$-20.6$} & 5.1 & 63.4\badchange{$+58.3$} & \textbf{85.4} \\
\midrule
\multicolumn{11}{@{}l}{\textit{RxR-CE, 1{,}223 episodes}}\\
ETP-R1 & \Ttwo{} & 64.4 & 24.9\badchange{$-39.5$} & 10.0 & 56.5\badchange{$+46.5$} & \textbf{79.1} & 65.1\goodchange{$-14.0$} & 8.7 & 81.5\badchange{$+72.8$} & \textbf{85.4} \\
ETPNav & \Ttwo{} & 58.0 & 23.4\badchange{$-34.6$} & 10.0 & 57.3\badchange{$+47.3$} & \textbf{78.3} & 62.9\goodchange{$-15.4$} & 8.1 & 81.0\badchange{$+72.9$} & \textbf{86.0} \\
\bottomrule
\end{tabular}
\end{table*}

\subsubsection{P1: Contact in Successful and Unsuccessful Episodes}
\label{sec:profiles}

\begin{figure}[t]
\centering
\includegraphics[width=\columnwidth]{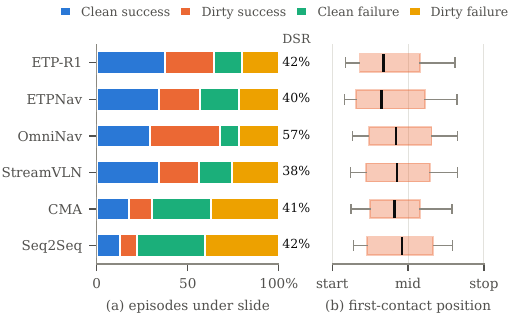}
\caption{R2R-CE under \slideon{}. (a) Episode outcome and contact status; the
dirty-success rate (DSR) is printed at right. (b) Position of the first
collision step within each dirty success, as a fraction of the episode
(boxes: quartiles; whiskers: 10th-90th percentiles; bar: median).}
\label{fig:contact-timing}
\Description{Two panels sharing one row per agent. Left: horizontal stacked bars of
clean success, dirty success, clean failure and dirty failure shares for six agents,
with dirty-success rates between 38 and 57 percent. Right: box plots showing that
first contact in a dirty success typically happens between one quarter and two thirds
of the way through the episode, with medians between 0.32 and 0.46.}
\end{figure}

\textbf{The agent with the highest success rate also has the highest dirty-success rate.} Under \slideon{}, the standard regime, 38.5-56.8\% of successful R2R-CE episodes contain at least one collision step (Table~\ref{tab:profile}, Figure~\ref{fig:contact-timing}a). From the 2020 recurrent baselines to recent VLAs and topological planners, better navigation has not produced cleaner successful trajectories. On RxR-CE's longer routes the share rises to 79.1\% for ETP-R1 (Wilson interval 76.1-81.8\%) and 78.3\% for ETPNav.

\textbf{Contact in a success is repeated and occurs early.} A dirty success by ETP-R1 contains 5.0 collision steps on average (median 3). Its first contact is not an artifact of the final approach to the goal: across agents, the median first contact occurs 32-46\% of the way through the episode (Figure~\ref{fig:contact-timing}b). Successful episodes nevertheless hold only 1.2-42.1\% of all collision steps; the rest occur in failures. Success rate alone therefore says little about whether a route was driven cleanly, while an aggregate collision rate hides whether contact occurred in a success or a failure.

\subsubsection{P2: Effects of Simulator Settings and Collision Definitions}
\label{sec:regime}

\textbf{The simulator regime changes success rate substantially.} Letting agents slide raises success rate by 11.7-36.1 percentage points on R2R-CE (McNemar $p<10^{-50}$ for every agent) and by 34.6 and 39.5 percentage points for ETPNav and ETP-R1 on RxR-CE. OmniNav drops from 67.5\% to 31.4\% without sliding. Yet of the papers we surveyed that report standard VLN-CE results, only \AuditRegimeStated{} state the regime, and none conditions safety on outcome. Published success rates thus partly reflect an undocumented simulator setting, which limits both cross-paper comparison and what they imply for deployment.

\textbf{Without sliding, contact turns successes into failures.} Under \slideoff{}, the per-episode collision-step rate rises from 4.0-8.0\% to 13.9-62.4\%, the stuck-episode rate from 4.2-6.3\% to 43.4-70.5\%, and the dirty-success rate falls for every agent except ETP-R1 (whose change, $+0.6$ percentage points, has a paired interval containing zero) to 17.6-48.4\%. The fall is not an artifact of shorter episodes: it persists after stratifying by episode length, and successes under \slideoff{} are no shorter than under \slideon{} except for StreamVLN. Instead, 44-91\% of the dirty successes under \slideon{} fail under \slideoff{}, against 11-42\% of the
clean ones: without tangential motion, the contact that a sliding agent glides
through stops it.

\textbf{The engine flag and displacement proxy capture different behavior.} The engine flag identifies commands shortened by the navigation mesh, including contacts that still allow sliding. The displacement proxy instead identifies near-zero motion and is useful for detecting stalls, but it classifies 77.7-87.6\% of episodes with engine-flagged contact as clean under \slideon{}, where sliding preserves some displacement. For example, it yields a 6.4\% dirty-success rate for ETP-R1, versus 41.7\% with the engine flag. The measures are therefore complementary, not interchangeable: we use the engine flag to measure contact and the proxy to identify stalled motion.

\textbf{The method is developed under \slideoff{}.} Because \slideon{} converts
blocked motion into tangential progress, \slideoff{} is the conservative regime
for an agent meant for robotic deployment, and the gap between regimes measures
how much reported success depends on sliding. The signal analysis and the guard
therefore use \slideoff{}, and the guard is also evaluated under \slideon{}.

\subsection{Collision-Prediction Signal Results}
\label{sec:predict}

% Machine-written by analysis/make_tables.py from results/summary/*.json -- do not edit by hand.
% Machine-written by analysis/make_tables.py -- do not edit by hand.
\begin{table}[t]
\centering\small
\setlength{\tabcolsep}{2.5pt}
\caption{Collision-prediction signals under \slideoff{}. Columns report AUROC,
movement-history recall, and the share of risky decisions with a geometrically
safe alternative within 10 score points (Swap); the latter two are percentages.
The largest AUROC per row is bold; $^{\ast}$ marks \Tone{} agents, whose
alternatives are primitives. \nmark{}: not exposed (OmniNav emits one waypoint,
with no scored alternatives).}
\label{tab:signals}
\begin{tabular}{@{}lccccccc@{}}
\toprule
& \multicolumn{2}{c}{External $\uparrow$}
& \multicolumn{3}{c}{Policy outputs $\uparrow$}
& \multicolumn{2}{c}{Coverage $\uparrow$} \\
\cmidrule(lr){2-3}\cmidrule(lr){4-6}\cmidrule(l){7-8}
Agent & Depth & Hist. & Ent. & Stop & Marg. & Hist.\,rec. & Swap \\
\midrule
\multicolumn{8}{@{}l}{\textit{R2R-CE}}\\
ETP-R1 & .811 & \textbf{.891} & .673 & .416 & .637 & 85.3 & 4.0 \\
ETPNav & .759 & \textbf{.916} & .564 & .524 & .492 & 91.1 & 0.7 \\
OmniNav$^{\ast}$ & .826 & \textbf{.862} & \nmark & .493 & \nmark & 82.3 & \nmark \\
StreamVLN$^{\ast}$ & .832 & \textbf{.981} & .539 & .358 & .518 & 98.6 & 0.4 \\
CMA$^{\ast}$ & .826 & \textbf{.974} & .523 & .301 & .519 & 99.3 & 2.5 \\
Seq2Seq$^{\ast}$ & .779 & \textbf{.931} & .412 & .415 & .414 & 99.1 & 3.0 \\
\midrule
\multicolumn{8}{@{}l}{\textit{RxR-CE}}\\
ETP-R1 & \textbf{.825} & .799 & .668 & .387 & .617 & 95.3 & 8.3 \\
ETPNav & \textbf{.826} & .814 & .560 & .523 & .505 & 95.7 & 2.6 \\
\bottomrule
\end{tabular}
\end{table}

\textbf{Policy outputs do not reliably expose collision risk.} In their native directions, the policy-derived signals range from area under the receiver operating characteristic curve (AUROC) 0.301-0.673 (Table~\ref{tab:signals}), several below 0.5. A below-chance AUROC indicates an inverse association rather than no information, but direction and strength vary across agents and signals, and even the strongest policy signal trails both external sources for its agent. A score-only guard would need agent-specific calibration.

\textbf{Repeated contact is highly predictable; first contact requires geometry.} Depth geometry reaches AUROC 0.759-0.832 on R2R-CE and movement history 0.862-0.981. History flags 82.3-99.3\% of collided forward steps, and all but one flagged step across all runs also collided (for ETP-R1, 28{,}428 of 28{,}428): under the pose tolerance above, a previously blocked forward command fails again almost deterministically. This captures the repeated presses that form stuck episodes, but by construction it cannot warn about the first contact at a new pose. Restricted to decisions at poses with no earlier stall, depth still reaches 0.714-0.851, whereas the policy signals fall to 0.478-0.645. Conversely, depth misses navigation-mesh boundaries without a rendered surface, which history identifies after the first blocked step.

\textbf{Score-constrained candidate swaps cover few risky decisions.} Among risky decisions, only 0.4-4.0\% have a geometrically safe alternative within 10 score points of the chosen action (Swap, Table~\ref{tab:signals}). For \Ttwo{} agents a suitable spatial candidate is often absent; for \Tone{} agents a contact-free primitive exists but usually lies far below the chosen action in score.

\textbf{Design consequences.} The method therefore reads risk from geometry and movement history, which behave consistently across agents, rather than from calibrated policy scores. It acts at the primitive-execution layer, where a forward command can be vetoed and replaced by a contact-free turn before the policy replans, rather than relying on an alternative in the candidate set. Both sources need only depth and odometry, so the method applies to either decision interface without access to policy internals.

% Machine-written by analysis/make_tables.py from results/summary/*.json -- do not edit by hand.
% Hand-styled from results/summary/guard_effect.json and guard_mechanism.json (2026-10-07).
\begin{table*}[h!]
\centering\small
\setlength{\tabcolsep}{2.6pt}
\caption{The collision guard, paired by episode with the unguarded run of the same agent
(R2R-CE: 1{,}839 episodes; RxR-CE: 1{,}223). Each metric reports No guard (left) and
Guard (right); small values give Guard$-$No guard, \textcolor{better}{green} better and
\textcolor{worse}{red} worse. Clean: success without any collision step. Immobile:
episodes with a 40-primitive window of net displacement below 0.5\,m, a stall measure
the stall veto cannot satisfy by construction. Own: share of executed primitives chosen
by the guard. $^{\ast}$: paired 95\% bootstrap interval excludes zero. SR, SPL, nDTW,
Clean, Immobile and Own in \%; steps and events per episode.}
\label{tab:guard}
\begin{tabular}{@{}lcccccccccccccccc@{}}
\toprule
& \multicolumn{2}{c}{SR $\uparrow$} & \multicolumn{2}{c}{SPL $\uparrow$}
& \multicolumn{2}{c}{nDTW $\uparrow$} & \multicolumn{2}{c}{Clean $\uparrow$}
& \multicolumn{2}{c}{Coll.\ steps $\downarrow$} & \multicolumn{2}{c}{Events $\downarrow$}
& \multicolumn{2}{c}{Immobile $\downarrow$} & Own \\
\cmidrule(lr){2-3}\cmidrule(lr){4-5}\cmidrule(lr){6-7}\cmidrule(lr){8-9}
\cmidrule(lr){10-11}\cmidrule(lr){12-13}\cmidrule(lr){14-15}
Agent & No & \textbf{Guard} & No & Guard & No & Guard & No & \textbf{Guard}
& No & \textbf{Guard} & No & Guard & No & Guard & \\
\midrule
\multicolumn{16}{@{}l}{\textit{R2R-CE, \slideoff{}}}\\
ETP-R1 & 50.5 & \textbf{57.4}\goodchange{$+6.9^{\ast}$} & 43.3 & 49.7$^{\ast}$ & 60.9 & 64.3$^{\ast}$ & 29.1 & \textbf{39.2}\goodchange{$+10.1^{\ast}$} & 18.1 & \textbf{3.2}\goodchange{$-83\%$} & 2.60 & 2.65 & 29.0 & 15.3 & 8.4 \\
ETPNav & 41.5 & \textbf{52.2}\goodchange{$+10.7^{\ast}$} & 36.4 & 45.7$^{\ast}$ & 57.9 & 62.0$^{\ast}$ & 28.3 & \textbf{37.9}\goodchange{$+9.6^{\ast}$} & 21.1 & \textbf{3.0}\goodchange{$-86\%$} & 2.15 & 2.51 & 33.0 & 16.3 & 9.8 \\
OmniNav & 31.4 & \textbf{60.7}\goodchange{$+29.3^{\ast}$} & 30.0 & 56.2$^{\ast}$ & 58.7 & 69.2$^{\ast}$ & 16.2 & \textbf{27.0}\goodchange{$+10.8^{\ast}$} & 18.7 & \textbf{7.1}\goodchange{$-62\%$} & 3.36 & 6.27 & 13.2 & 12.3 & 20.1 \\
StreamVLN & 27.7 & \textbf{50.1}\goodchange{$+22.4^{\ast}$} & 25.2 & 44.0$^{\ast}$ & 56.6 & 60.4$^{\ast}$ & 22.8 & \textbf{29.7}\goodchange{$+6.9^{\ast}$} & 201.5 & \textbf{6.8}\goodchange{$-97\%$} & 46.78 & 6.29 & 54.6 & 10.1 & 18.4 \\
CMA & 12.1 & \textbf{24.7}\goodchange{$+12.6^{\ast}$} & 11.5 & 22.4$^{\ast}$ & 46.1 & 49.1$^{\ast}$ & 9.3 & \textbf{12.7}\goodchange{$+3.4^{\ast}$} & 299.8 & \textbf{6.3}\goodchange{$-98\%$} & 3.63 & 5.78 & 67.8 & 22.5 & 11.7 \\
\midrule
\multicolumn{16}{@{}l}{\textit{RxR-CE, \slideoff{}}}\\
ETP-R1 & 24.9 & \textbf{52.9}\goodchange{$+28.0^{\ast}$} & 22.0 & 46.0$^{\ast}$ & 46.1 & 64.2$^{\ast}$ & 8.7 & \textbf{20.5}\goodchange{$+11.8^{\ast}$} & 81.0 & \textbf{4.7}\goodchange{$-94\%$} & 3.60 & 3.54 & 77.6 & 16.5 & 16.5 \\
ETPNav & 23.4 & \textbf{49.7}\goodchange{$+26.3^{\ast}$} & 20.7 & 43.0$^{\ast}$ & 44.4 & 61.8$^{\ast}$ & 8.7 & \textbf{19.8}\goodchange{$+11.1^{\ast}$} & 82.7 & \textbf{4.6}\goodchange{$-94\%$} & 3.22 & 3.50 & 77.6 & 14.6 & 15.6 \\
\midrule
\multicolumn{16}{@{}l}{\textit{R2R-CE, \slideon{}}}\\
ETP-R1 & 64.6 & 63.8\badchange{$-0.8$} & 55.5 & 55.2 & 65.8 & 66.2 & 37.7 & \textbf{50.1}\goodchange{$+12.4^{\ast}$} & 3.9 & \textbf{1.4}\goodchange{$-64\%$} & 1.13 & 0.90 & 3.1 & 3.8 & 4.0 \\
ETPNav & 56.9 & 56.3\badchange{$-0.6$} & 49.1 & 48.9 & 62.4 & 62.9$^{\ast}$ & 34.0 & \textbf{45.4}\goodchange{$+11.4^{\ast}$} & 4.5 & \textbf{1.9}\goodchange{$-58\%$} & 1.15 & 0.98 & 4.0 & 4.0 & 4.2 \\
StreamVLN & 56.0 & 56.7\goodchange{$+0.7$} & 49.4 & 49.8 & 63.3 & 62.9 & 34.4 & \textbf{42.5}\goodchange{$+8.1^{\ast}$} & 26.0 & \textbf{9.4}\goodchange{$-64\%$} & 5.98 & 5.28 & 6.6 & 1.5 & 9.1 \\
CMA & 30.2 & 29.6\badchange{$-0.6$} & 28.3 & 27.2 & 51.2 & 50.5$^{\ast}$ & 17.8 & \textbf{21.9}\goodchange{$+4.1^{\ast}$} & 27.2 & \textbf{3.2}\goodchange{$-88\%$} & 1.54 & 2.20 & 9.4 & 6.5 & 6.2 \\
\bottomrule
\end{tabular}
\end{table*}

\begin{table}[t]
\centering\footnotesize
\setlength{\tabcolsep}{1.5pt}
\caption{Guard mechanisms under \slideoff{}. Flips: failures$\to$successes / successes$\to$failures. Stuck: share of net success gain from episodes stuck without the guard. Timeout (500 steps) and budget exhaustion show No guard / Guard rates (\%); guarded values are bold. Budget applies to ETP agents only.}
\label{tab:guard-mechanism}
\begin{tabular}{@{}lcccc@{}}
\toprule
Agent & \shortstack{Flips\\F$\to$S/S$\to$F}
& \shortstack{Stuck\\gain (\%)}
& \shortstack{Timeout\\No/Guard (\%)}
& \shortstack{Budget\\No/Guard (\%)} \\
\midrule
\multicolumn{5}{@{}l}{\textit{R2R-CE}}\\
ETP-R1 & 173/46 & 93 & 0.1/\textbf{0.1} & 38.4/\textbf{28.6} \\
ETPNav & 233/37 & 94 & 0.0/\textbf{0.0} & 41.2/\textbf{26.8} \\
OmniNav & 601/62 & 95 & 0.0/\textbf{0.0} & \nmark{}/\nmark{} \\
StreamVLN & 430/18 & 79 & 52.9/\textbf{1.1} & \nmark{}/\nmark{} \\
CMA & 251/20 & 100 & 62.8/\textbf{11.1} & \nmark{}/\nmark{} \\
\midrule
\multicolumn{5}{@{}l}{\textit{RxR-CE}}\\
ETP-R1 & 360/17 & 99 & 0.2/\textbf{0.1} & 80.9/\textbf{55.8} \\
ETPNav & 342/20 & 99 & 0.3/\textbf{0.2} & 80.5/\textbf{54.0} \\
\bottomrule
\end{tabular}
\end{table}

\subsection{Collision Guard Results}
\label{sec:guard-results}

We evaluate one guard configuration on five R2R-CE agents and two RxR-CE agents under \slideoff{}, and on four R2R-CE agents under \slideon{}. Each guarded run is paired by episode with the unguarded run of the same agent (Table~\ref{tab:guard}).

\subsubsection{Without Sliding: Fewer Collisions, More Successes}
\label{sec:guard-main}

\textbf{Collisions fall and success rises on every agent.} On R2R-CE the guard removes 62-98\% of collision steps while choosing only 8.4-20.1\% of executed primitives. Success rate rises by 6.9-29.3 percentage points, SPL by 6.4-26.2 points, nDTW by 3.0-10.5 points, and collision-free success rate by 3.4-10.8 percentage points; every paired interval excludes zero. The guard turns 173-601 failures into successes per agent and only 18-62 successes into failures (Table~\ref{tab:guard-mechanism}). It recovers 49-81\% of the success that each agent loses when sliding is disabled, without enabling sliding. On RxR-CE, whose camera rig, turn angle, and body differ, the same metric thresholds raise success by 28.0 and 26.3 percentage points and remove 94\% of collision steps.

\textbf{The gain comes mostly from ending stuck episodes.} Stuck events vanish under the guard, but that is true by construction of the stall veto, so we also measure immobility (a 40-primitive window with under 0.5\,m net displacement), which falls from 13.2-67.8\% to 10.1-22.5\% of episodes. Of the net success gain, 93-100\% comes from episodes that were stuck without the guard (79\% for StreamVLN; Table~\ref{tab:guard-mechanism}).

\textbf{The guard also reduces budget exhaustion and timeouts.} Escaping a repeated press returns control to the policy instead of spending its budget at one obstacle: ETP-R1 exhausts its 15-decision budget in 28.6\% of episodes instead of 38.4\%, and the timeout share falls from 52.9\% to 1.1\% for StreamVLN and from 62.8\% to 11.1\% for CMA (Table~\ref{tab:guard-mechanism}). The 19\% of StreamVLN's net success gain attributable to 500-step timeouts complements the gains from stuck-episode recovery. OmniNav also gains 5.2 percentage points (95\% interval 3.0-7.4) on episodes that were never stuck; its episodes grow longer (62.9 to 86.2 primitives) because, unguarded, it stops early after 25 decisions without progress.

\textbf{Collision events need not fall with collision steps.} Events fall by 87\% for StreamVLN, are unchanged for ETP-R1, and rise for ETPNav, CMA, and OmniNav. Intervening and replanning splits a long press into short contacts: 51-71\% of the remaining events lie within 12 primitives of another, and merging them gives 0.95-2.81 events per episode, below each agent's unguarded count. Of the remaining collision steps, 49-62\% occur on an escape forward that meets a boundary not visible to depth, 37-47\% on policy forwards that depth did not flag, and at most 3\% on yielded steps.

\subsubsection{With Sliding: Cleaner Successes at No Success Cost}
\label{sec:guard-slide}

Under \slideon{}, the regime of published results, blocked commands still produce tangential motion, so there is little stalling to undo. The guard changes success rate by $-0.8$ to $+0.7$ percentage points, with every interval containing zero, while raising collision-free success rate by 4.1-12.4 percentage points (every interval excluding zero) and removing 58-88\% of collision steps. It chooses 4.0-9.1\% of executed primitives. This result does not depend on un-sticking: in the published regime, the guard makes the same agents succeed as often but more cleanly.

\subsubsection{Ablations and Baselines}
\label{sec:ablation}

% Machine-written by analysis/make_tables.py from results/summary/*.json -- do not edit by hand.
% Hand-styled from results/summary/guard_effect.json and guard_mechanism.json (2026-10-07).
\begin{table}[t]
\centering\small
\setlength{\tabcolsep}{1.6pt}
\caption{Guard components on ETP-R1, R2R-CE, \slideoff{}. Small values compare Guard with No guard, and indented variants with Guard (\textcolor{better}{green} better, \textcolor{worse}{red} worse). Stall only and depth only remove the depth and the stall veto; 8-frame depth accumulates eight frames; the oracle replaces both predictors with the navigation mesh itself. SR, SPL, Clean in \%; steps and events per episode.}
\label{tab:ablation}
\begin{tabular}{@{}lccccc@{}}
\toprule
Condition & SR $\uparrow$ & SPL $\uparrow$ & Clean $\uparrow$ & Steps $\downarrow$
& Events $\downarrow$ \\
\midrule
No guard & 50.5 & 43.3 & 29.1 & 18.1 & 2.60 \\
\textbf{Guard (ours)} & \textbf{57.4}\goodchange{$+6.9$} & \textbf{49.7}\goodchange{$+6.4$} & \textbf{39.2}\goodchange{$+10.1$} & \textbf{3.2}\goodchange{$-14.9$} & 2.65\badchange{$+0.05$} \\
\quad stall only & 57.6\goodchange{$+0.2$} & 49.5\badchange{$-0.2$} & 29.1\badchange{$-10.1$} & 4.4\badchange{$+1.2$} & 3.51\badchange{$+0.86$} \\
\quad depth only & 52.9\badchange{$-4.5$} & 45.5\badchange{$-4.2$} & 39.2 & 13.2\badchange{$+10.0$} & 2.05\goodchange{$-0.60$} \\
\quad 8-frame depth & 55.2\badchange{$-2.2$} & 47.1\badchange{$-2.6$} & 44.1\goodchange{$+4.9$} & 2.1\goodchange{$-1.1$} & 1.81\goodchange{$-0.84$} \\
\quad $\tau=0.10$\,m & 59.1\goodchange{$+1.7$} & 51.1\goodchange{$+1.4$} & 47.0\goodchange{$+7.8$} & 2.3\goodchange{$-0.9$} & 1.92\goodchange{$-0.73$} \\
\midrule
\quad oracle predictor & 62.6\goodchange{$+5.2$} & 53.9\goodchange{$+4.2$} & 61.3\goodchange{$+22.1$} & 0.2\goodchange{$-3.0$} & 0.15\goodchange{$-2.50$} \\
\bottomrule
\end{tabular}
\end{table}

Table~\ref{tab:ablation} varies the guard on ETP-R1 under \slideoff{}.

\textbf{The two vetoes do different jobs.} With the stall veto alone, which is also the simplest reactive baseline (turn away after a blocked step), success matches the full guard (57.6\% versus 57.4\%, a 0.2-percentage-point difference) but collision-free success is unchanged from no guard (29.1\%). With the depth veto alone, success rises only to 52.9\%, but collision-free success reaches 39.2\%, exactly the full guard's. Movement history thus supplies the success gain and depth supplies the clean successes, and the two add up on collision steps.

\textbf{Oracle bound.} Replacing both predictors with the navigation mesh itself, with the same escape and caps, gives 62.6\% success and 61.3\% collision-free success with 0.24 collision steps per episode. The guard thus attains 57\% of the oracle's success gain, 84\% of its collision-step reduction, and 31\% of its collision-free-success gain (on StreamVLN, 76\% of the success gain). The remaining gap lies in perception, since the escape machinery is shared; most headroom for a better predictor is in clean successes.

\textbf{Sensitivity.} The primary depth threshold, $\tau=0.05$\,m, was fixed before the evaluation runs rather than tuned on these results. In this ablation, $\tau=0.10$\,m performs better on SR, SPL, and collision-free success. Across $\tau\in\{0.02, 0.05, 0.10, 0.20\}$\,m, success stays within 57.2-59.1\%, while collision-free success rises from 33.6\% to about 47\% at $\tau\geq0.10$\,m; every setting beats no guard on success, SPL, and clean success. Eight-frame depth memory removes more collision steps but lowers success to 55.2\%, consistent with its false vetoes.
%======================================================================
\section{Conclusion}
\label{sec:conclusion}

\subsection{Discussion and Limitations}
\label{sec:discussion}

\textbf{What VLN papers should report.} Three additions would expose contact with little overhead. First, state the collision-handling regime, which moves success rate by 11.7-39.5 points in our paired runs. Second, report the dirty-success rate using the simulator's engine flag: under \slideon{}, the displacement proxy classifies 77.7-87.6\% of engine-contact episodes as clean. Third, report collision steps and events together, since removing repeated presses can cut steps sharply while splitting the remaining contact into more events.

\textbf{Where collision information lives.} Collision risk is not reliably available through the evaluated policy-score interfaces, although hidden policy representations might still encode it. Depth geometry and movement history are defined consistently across decision interfaces, which is what lets one guard configuration wrap recurrent, topological, and VLA policies. The two signals are complementary: history removes repeated presses almost without false alarms but only after a first blocked step, while depth anticipates first contacts but misses mesh boundaries without a rendered surface. The oracle gap shows that a better first-contact predictor, rather than a different escape, is the main remaining lever.

\textbf{Measurement limitations.} Our contact flag marks a navigation-mesh event rather than a rigid-body impact, with no force, speed, damage, or obstacle semantics, and no dynamic agents or objects absent from the mesh. The benchmark measures contact occurrence, not physical harm.

\textbf{Method limitations.} The guard assumes panoramic depth, odometry, access to queued controller primitives, and action interception. Its constants were fixed before evaluation and shared across agents and benchmarks; we report sensitivity to $\tau$ and the binding rate of $K$, but not to $\delta$ or the consecutive limit. Its runtime cost is 8.5\,ms per primitive on a CPU, 14\% of a guarded ETP-R1 step. Odometry drift can weaken stall matching, depth cannot reveal every boundary, and the caps deliberately let some flagged steps through, so the guard is a heuristic rather than a formal guarantee. Under \slideoff{} most of its success gain comes from ending stuck episodes; under \slideon{} it improves cleanliness rather than success. It has not been validated on a physical robot.

Standard VLN metrics measure goal completion and route quality without reporting simulator contact. \benchname{} reports contact by episode outcome and quantifies its sensitivity to simulator settings and collision definitions. Under the standard \slideon{} regime, 38.5-56.8\% of successful R2R-CE episodes contain engine contact across six agents, and disabling sliding lowers success rate by 11.7-36.1 points. The agents' scores do not expose collision risk reliably, whereas depth and movement history provide signals shared across decision interfaces. A training-free guard built on these signals removes 62-98\% of collision steps and raises success by 6.9-29.3 points across five agents under \slideoff{}, and raises collision-free success by 4.1-12.4 points at unchanged success under \slideon{}, without retraining or replacing the navigation policy.

\balance
\bibliographystyle{ACM-Reference-Format}
\bibliography{references}

\end{document}